\documentclass[lettersize,journal]{IEEEtran}

\usepackage{amsmath,amsfonts}
\usepackage{array}
\usepackage{textcomp}
\usepackage{url}
\usepackage{verbatim}
\usepackage{graphicx}
\usepackage{cite}

\usepackage{adjustbox}
\usepackage{booktabs}
\usepackage{colortbl}
\usepackage{multirow}
\usepackage{subcaption}
\usepackage{xcolor}
\usepackage[normalem]{ulem}
\useunder{\uline}{\ul}{}

\usepackage{orcidlink}

\begin{document}

\title{
Unifying Physically-Informed Weather Priors in \\ A Single Model for Image Restoration \\ Across Multiple Adverse Weather Conditions}

\author{
Jiaqi Xu\orcidlink{0000-0003-1279-6782}, \textit{Student Member, IEEE},
Xiaowei Hu\orcidlink{0000-0002-5708-7018}, \textit{Member, IEEE},
Lei Zhu\orcidlink{0000-0003-3871-663X}, \textit{Member, IEEE}, \\
and Pheng-Ann Heng\orcidlink{0000-0003-3055-5034}, \textit{Senior Member, IEEE}
\thanks{J. Xu and P.-A. Heng are with the Department of Computer Science and Engineering, The Chinese University of Hong Kong, Hong Kong SAR, China.}
\thanks{X. Hu is with the Shanghai Artificial Intelligence Laboratory, Shanghai, China.}
\thanks{L. Zhu is with ROAS Thrust, the Hong Kong University of Science and Technology (Guangzhou), Guangzhou, China and
The Hong Kong University of Science and Technology, Department of Electronic and Computer Engineering, Hong Kong SAR, China.}
\thanks{Corresponding author: Xiaowei Hu (e-mail: huxiaowei@pjlab.org.cn).}
}

\markboth{IEEE Transactions on Circuits and Systems for Video Technology}{}%

\maketitle

\begin{abstract}
Image restoration under multiple adverse weather conditions aims to develop a single model to recover the underlying scene with high visibility.
Weather-related artifacts vary with the particle's distance to the camera according to the established scene visibility analysis, where close and faraway regions are more affected by falling drops and fog effects, respectively.
Existing methods fail to consider this weather-specific physical visual process; thus, the restoration performance is limited.
In this work, we analyze the common visual factors in adverse weather conditions and present a unified imaging model that considers the individually visible particles and fog-like aggregate scattering effects.
Further, we design a novel weather-prior-based network, which leverages the weather-related prior information to help recover the scene by enhancing the features using the estimated occlusion and transmission.
Experimental results in multiple adverse scenarios show the superiority of our method against state-of-the-art methods.
\end{abstract}


\section{Introduction}
\IEEEPARstart{A}{dverse} weather conditions, such as haze, rain, and snow, degenerate the scene visibility and contrast,
which unfavorably impair the visual appearance of the captured photo and the performance of downstream vision systems, \textit{e.g.}, autonomous driving and video surveillance.
There are many efforts to recover the clear scene and to make the vision systems more robust to various bad weather conditions,
such as dehazing~\cite{he2010single,cai2016dehazenet,li2018benchmarking,wu2021contrastive},
deraining~\cite{garg2007vision,fu2017removing,qian2018attentive,li2019heavy},
and desnowing~\cite{liu2018desnownet,chen2020jstasr,zhang2021deep}.
Nevertheless, these works mainly focus on designing specific methods for restoring the captured image under one typical weather condition,
limiting their capabilities to deal with complex artifacts under multiple weather scenarios.

Recently, All-in-One~\cite{li2020all} and its following works~\cite{valanarasu2022transweather,chen2022learning,ozdenizci2023restoring,zhu2023learning} use a single deep-network-based method to deal with image artifacts caused by many types of adverse weathers.
The all-in-one design benefits from a compact model using only one set of model parameters, which is able to generalize to different weather scenarios.
However, existing image restoration methods under multiple adverse weather conditions suffer from several limitations.
First, some approaches~\cite{valanarasu2022transweather,chen2022learning,ozdenizci2023restoring} fail to consider the specific imaging characteristics under weather conditions.
Ignoring physical phenomena underlying the visual process in adverse weather situations may limit the performance of the developed methods, and the physical prior plays an important role in understanding weather-related artifacts and scene recovery.
Second, All-in-One~\cite{li2020all} and WGWS-Net~\cite{zhu2023learning} use separate encoders or model parameters for different weather conditions to obtain the restored image.
Such designs need manual selection for the pathway and cannot tackle the ambiguity among different weather types.

\begin{figure}
    \centering
    \includegraphics[width=\hsize]{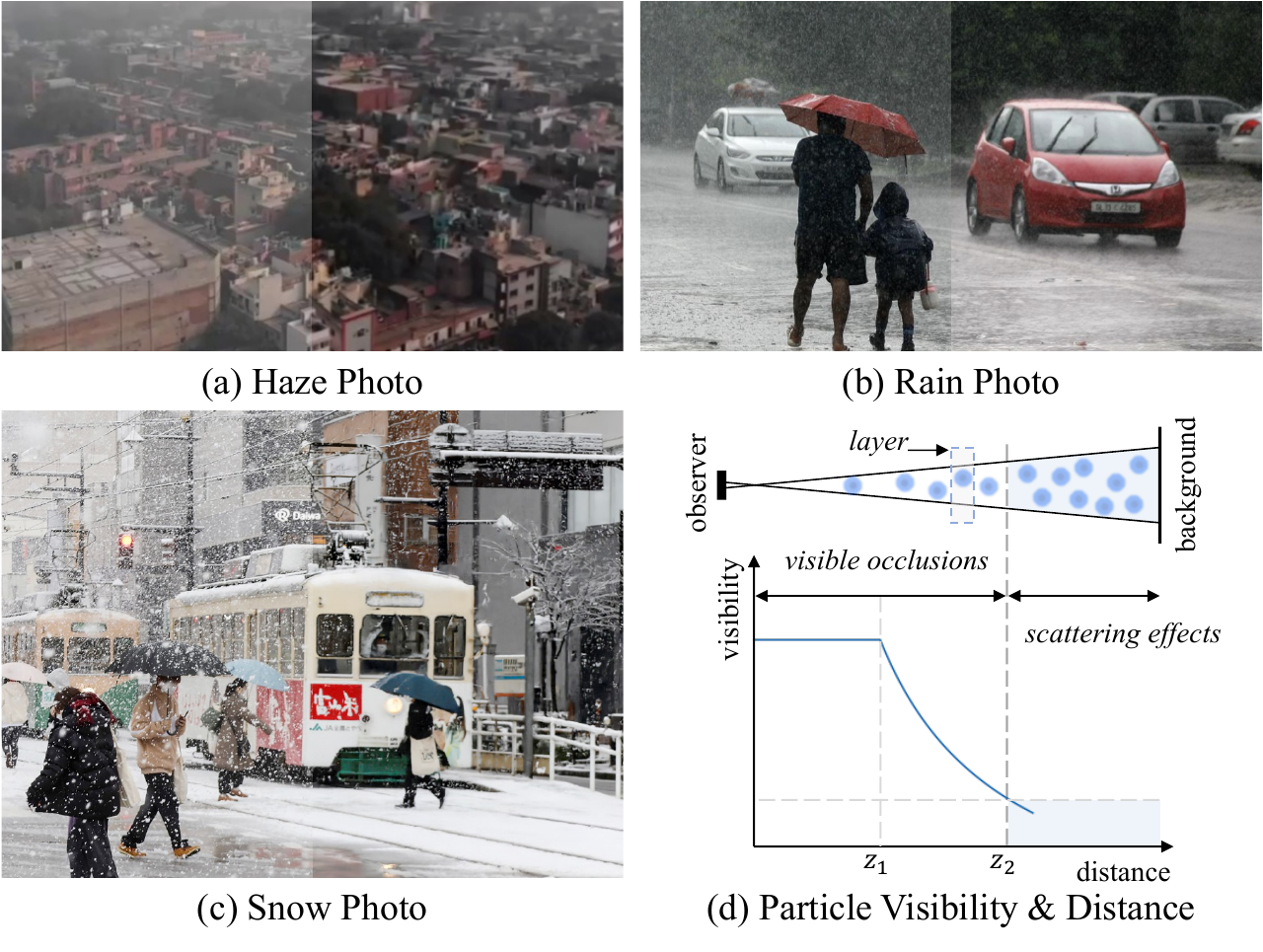}
    \caption{Examples of real photos for (a)~haze, (b)~rain, and (c)~snow show the presence of optionally visible particles and aggregate scattering effects with depth. Our restoration results for (a)~haze, (b)~rain, and (c)~snow are shown in the right part. (d) shows~the visibility analysis of particle's intensity change versus drop's distance~\cite{garg2007vision}.}
    \label{fig:visibility}
\end{figure}

Different weather conditions exhibit distinct physical properties and visual effects. For instance, haze droplets are too small to be individually detected by the camera, resulting in aggregate scattering effects. In contrast, rain and snow particles, when close to the camera, are visibly occlusive.
Additionally, various weather conditions share similar physical and visual processes. The visibility of scenes in rain and snow depends on the scene depth, with distant regions experiencing a higher volume of drops, resembling the effects of fog, especially in heavy rain or snow; see Fig.~\ref{fig:visibility}~(a), (b), and (c).
According to Garg and Kayar~\cite{garg2007vision}, the intensity change caused by falling particles is modeled as a function of the particle's distance from the camera, which can be divided into regions where the particles are either visible or not individually visible; see Fig.~\ref{fig:visibility}~(d).

In this work, we design a novel physically-based unified imaging model to characterize the images captured under common adverse weather conditions, including haze, rain, and snow, motivated by the physics-based vision research~\cite{garg2007vision,hu2019depth}.
Generally, the degenerated visual effects under common adverse weathers are affected by individually visible particle occlusions close to the camera, \textit{e.g.}, rain streaks and snowflakes, and fog-like aggregate scattering effects far away from the camera.
Hence, we present a simple but effective formulation to capture the essential degenerated factors in bad weather conditions by considering the visible rain and snow occlusion as well as the haze effects described by the atmospheric scattering model \cite{narasimhan2002vision,he2010single}.
In this way, we are able to estimate the corresponding occlusion, transmission, and atmospheric light and then leverage the estimated imaging information as weather priors to help scene restoration.

Further, we formulate a Weather-prior-based Network (WeatherNet) for image restoration across multiple adverse weather conditions.
Our network consists of two stages, \textit{i.e.}, the weather-related prior estimation stage and the weather prior-based scene refinement stage.
Specifically, the first stage is to estimate weather-related priors, including transmission, occlusion, and light, while the second stage is to leverage the weather prior-related information and features to recover the clear scene.
A weather-aware cross-attention module is designed for effectively injecting the weather priors into the scene recovery during the second stage.
We design the transmission-guided global attention to enhance the features from non-local regions with similar transmission values and occlusion-guided local attention to propagate features from neighboring non-occluded areas.
To this end, the scene refinement process establishes both global and local dependencies from weather-prior-related features in a transmission and occlusion-guided manner.
Our end-to-end network has the ability to tackle the degenerated image under any bad weather condition, including the mixed weather types, and generate the restored clean image.

Our main contributions are as follows:
\begin{itemize}
\item We formulate a unified imaging model for common adverse weather conditions (haze, rain, snow) based on the physical visual phenomena and synthesize a new dataset with mixed weather effects.
\item We develop a weather-prior-based network for image restoration, featuring a weather-aware cross-attention module that effectively incorporates transmission and occlusion information.
\item Experimental results show that our network outperforms state-of-the-art methods in both quantitative and qualitative evaluations across various weather scenarios. This work underscores the importance of imaging in designing and evaluating methods for adverse weather restoration.
\end{itemize}

\section{Related Work}
\label{sec:background}

\subsection{Specific Weather-Related Restoration}

Image restoration under adverse weather conditions has been widely studied, including
dehazing~\cite{narasimhan2002vision,he2010single,cai2016dehazenet,li2018benchmarking,zhang2018densely,qin2020ffa,dong2020multi,song2023vision,wang2017single,xie2021variational},
deraining~\cite{garg2007vision,fu2017removing,yang2017deep,li2018recurrent,wang2019spatial,hu2019depth,li2019heavy,jiang2020multi,yang2020single,you2015adherent,qian2018attentive,hu2021single,jiang2020decomposition,wu2020subjective},
and desnowing~\cite{barnum2010analysis,liu2018desnownet,chen2020jstasr,chen2021all,zhang2021deep,quan2023image}.
These works mainly focus on designing algorithms for restoring specific weather-related artifacts.

\paragraph{Haze Image Restoration}
The atmospheric scattering model is widely used to describe the formation of hazy images~\cite{narasimhan2002vision,he2010single}.
Early image dehazing works relied on the haze imaging model, physical priors, and heuristics~\cite{fattal2008single,he2010single,berman2016non}.
Later, deep learning-based methods showed improved performance,
which leveraged the haze imaging model and predicted the physical model-related components~\cite{cai2016dehazenet,ren2016single,li2017aod,zhang2018densely}
or recovered the haze-free scene images end-to-end~\cite{li2018single,qu2019enhanced}.
More recent studies proposed advanced and prior-based architecture and module designs~\cite{ren2018gated,deng2019deep,liu2019griddehazenet,chen2019gated,qin2020ffa,dong2020multi,guo2022image,ye2022perceiving,song2023vision},
such as attention~\cite{qin2020ffa}, feature fusion~\cite{dong2020multi}, and transformer~\cite{guo2022image,song2023vision}.
Besides, some works considered training strategies, generalization ability, and other properties~\cite{sakaridis2018semantic,hong2020distilling,shao2020domain,wu2021contrastive,zheng2021ultra,chen2021psd,liu2021synthetic,yang2022self},
including contrastive learning~\cite{wu2021contrastive}, domain generalization~\cite{shao2020domain,liu2021synthetic}, and semantic understanding~\cite{sakaridis2018semantic}.

\paragraph{Rain Image Restoration}
Early works detected and removed rain by analyzing the characteristics of rain streaks in image space or frequency space~\cite{garg2007vision,santhaseelan2015utilizing}.
Deep models were then used for rain removal with promising results~\cite{fu2017removing,yang2017deep}.
The following works presented more advanced architectures and techniques, including recurrent network~\cite{li2018recurrent}, conditional GAN~\cite{zhang2019image}, attention mechanism~\cite{wang2019spatial}, multi-scale feature fusion~\cite{jiang2020multi}, and transformer~\cite{xiao2022image}.
Other studies discussed the rain appearance modeling besides rain streaks~\cite{hu2019depth,hu2021single,wang2020rethinking,yang2020single}.
Besides, rain generation~\cite{wang2021rain}, contrastive learning~\cite{ye2022unsupervised}, semi-supervised learning~\cite{huang2021memory}, and adversarial attacks~\cite{yu2022towards} were also explored to better understand the rain properties in the real world and help downstream visions.
Additionally, some works dealed with the removal of raindrops~\cite{qian2018attentive,quan2019deep,quan2021removing} with similar techniques.

\paragraph{Snow Image Restoration}
The visible snow artifacts captured by the camera mainly differ from rain artifacts in that snow particles are with complex temporal trajectories and varying sizes and shapes.
Early works applied frequency-space analysis~\cite{barnum2010analysis} and histogram of orientations of streaks~\cite{bossu2011rain} to detect and remove snow and rain.
DesnowNet~\cite{liu2018desnownet} was among the first deep learning-based method for snow removal with a two-stage structure.
JSTASR~\cite{chen2020jstasr} presented a size and transparency-aware network for desnowing.
HDCWNet~\cite{chen2021all} showed that dual-tree complex wavelet representation could help extract snow patterns.
Zhang~\textit{et al.}~\cite{zhang2021deep} considered the semantic and depth priors for snow removal.

While these weather-specific approaches are effective at addressing individual weather conditions, they struggle to restore images affected by multiple or mixed weather scenarios.

\subsection{Multiple Weather-Related Restoration}
Several recent works focused on developing single network-based methods for image restoration under multiple adverse weather conditions~\cite{li2020all,valanarasu2022transweather,chen2022learning,ozdenizci2023restoring,zhu2023learning}.
This task is more challenging than image restoration for one type of specific weather, as the images suffer from various weather-related artifacts.
All-in-One~\cite{li2020all} comprised separate encoders for different weather types and a generic decoder, which introduced weather-specific and general operations.
TransWeather~\cite{valanarasu2022transweather} was a transformer-based network with learnable weather-type queries for dealing with weather-related image degradations.
Chen~\textit{et al.}~\cite{chen2022learning} proposed a two-stage teacher-student knowledge distillation training scheme and a contrastive learning regularization to enhance the model capability for image restoration under weather conditions.
{\"O}zdenizci and Legenstein~\cite{ozdenizci2023restoring} adapted the diffusion model for weather-related artifact removal and introduced a patch-based inference strategy for a stable denoising process.
Zhu~\textit{et al.}~\cite{zhu2023learning} learned weather-general and weather-specific features but carelessly overlooked the general visible occlusions in the imaging model.
AIRFormer \cite{gao2023frequency} tackled multi-weather image restoration using a frequency-guided transformer. GridFormer \cite{wang2024gridformer} proposed a grid-structured transformer with enhanced attention mechanisms. MWFormer \cite{zhu2024mwformer} extracted weather-specific features via a hyper-network but required multiple rounds of inference for images with mixed weather conditions.
Xu~\textit{et al.}~\cite{xu2024towards} developed a semi-supervised learning framework that adopts various large vision-language models to provide pesudo labels for real photos.
Nevertheless, these methods either ignore weather-related physical visual characteristics~\cite{valanarasu2022transweather,chen2022learning,ozdenizci2023restoring,wang2024gridformer}, leading to suboptimal restoration quality, especially in real-world applications, or require additional weather-type classification during inference~\cite{li2020all,zhu2023learning,zhu2024mwformer}, making them incapable of handling mixed weather conditions.
This work seeks to improve adverse-weather image restoration by explicitly incorporating the imaging priors associated with diverse weather conditions.

\subsection{Imaging Models Under Weather Conditions}
Weather conditions can be broadly classified into steady, \textit{e.g.}, haze, and dynamic, \textit{e.g.}, rain and snow, in terms of physical properties and visual effects~\cite{garg2007vision}.
Haze effects are widely described by the atmospheric scattering model \cite{narasimhan2002vision}, where individual droplets are too small to be observed through the camera.
On the contrary, the falling rain and snow particles are clearly visible in the photo.
Snowflakes have varying sizes and shapes and complex trajectories compared to raindrops.
The visual effects of raindrops and snowflakes depend on weather conditions as well as camera parameters.
Nevertheless, the visual appearances of steady and dynamic weather conditions share some common properties, as droplets in the faraway regions can only produce aggregate scattering effects, and no individual particles are visible~\cite{garg2007vision}.

Early weather-related image restoration works provided a unified appearance modeling of rain and snow in the image space~\cite{garg2007vision} or in the frequency space~\cite{barnum2010analysis}.
Later, deep learning-based methods separately studied the removal of visible rain streaks~\cite{fu2017removing} and snowflakes~\cite{liu2018desnownet}.
Some following works revisited the aggregate scattering effects in rain~\cite{hu2019depth,li2019heavy,hu2021single} and snow~\cite{chen2020jstasr}.
For image restoration under multiple adverse weather conditions, All-in-One~\cite{li2020all} used the rain~\cite{li2019heavy} and snow~\cite{liu2018desnownet} appearance modeling and argued that images in different weather conditions undergo different physical visual processes, which was then adopted by the following works~\cite{valanarasu2022transweather}.
Additionally, some combinations of weather effects are artificially created in image decomposition \cite{han2022blind}.
In contrast, we argue that the images captured in bad weather conditions can be viewed from a more unified perspective through the lens of visibility analysis.

\section{Imaging Model Formulation}
\label{sec:imaging}

\subsection{Scene Visibility Analysis}
\paragraph{Visibility and distance}
According to Garg and Kayar~\cite{garg2007vision}, the visibility of falling particle occlusions in rain and snow varies with the particle's distance $z$ from the camera and can be mainly divided into three situations as shown in Fig.~\ref{fig:visibility}~(d):

\begin{itemize}
  \item In region $0< z < z_1$, the particle visibility is affected by camera exposure time and does not depend on $z$, where $z_1 = 2fa$, $a$ is the drop radius, and $f$ is the focal length.
  \item In region $z_1 < z < z_2$, the particle visibility decreases with the distance as $1/z$, where $z_2 = R z_1$, and $R$ is a constant.
  \item In region $z > z_2$, the appearance of the individual particle is too small to be captured by the camera, and a volume of particles produces fog-like aggregate scattering effects.
\end{itemize}

Based on the aforementioned observations, we classify the visual phenomena caused by rain and snow as follows: in the region where $z < z_2$, we refer to them as \textit{visible particle occlusions}, while in the region where $z > z_2$, we denote them as \textit{haze scattering effects}.
Furthermore, within the region where $z < z_2$, it is possible for multiple visible drops at varying depths in the 3D space to project onto the same pixel location in the 2D image plane. Consequently, a pixel with greater scene depth is subject to the influence of a greater number of drops, each with different particle visibility, which is termed \textit{volumetric effects}.

\begin{figure}
    \centering
    \includegraphics[width=\hsize]{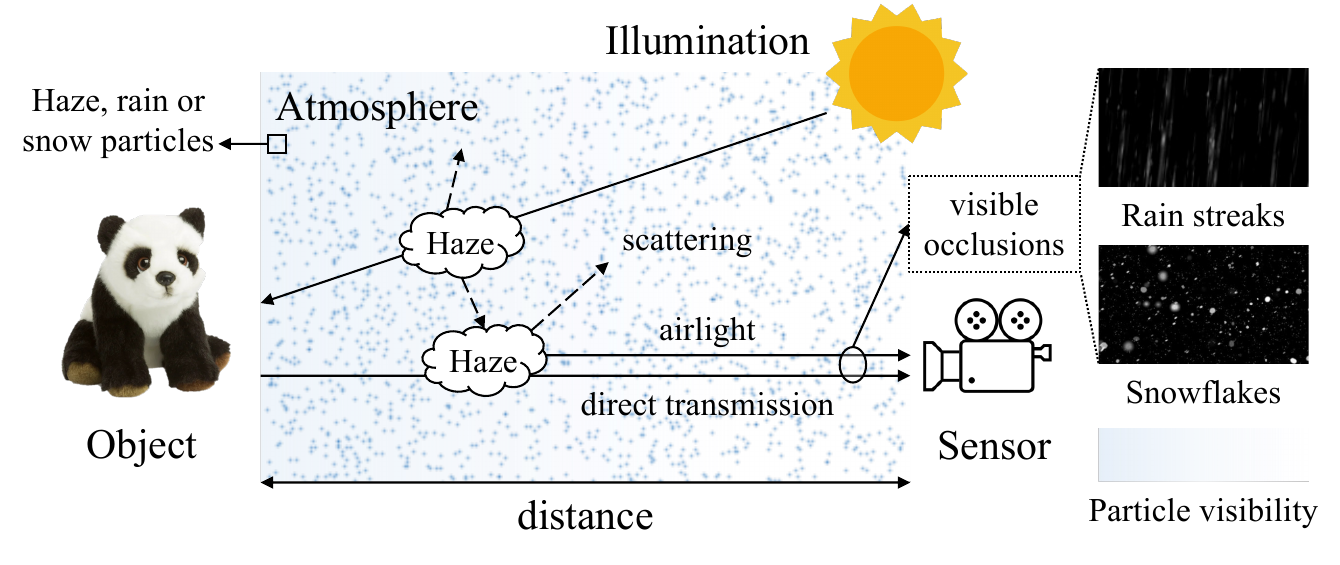}
    \caption{The visual process in adverse weather conditions.
    The weather-related artifacts are caused by haze scattering effects and visible particle occlusions.}
    \label{fig:optical}
\end{figure}

\begin{figure*}
    \centering
    \includegraphics[width=0.98\hsize]{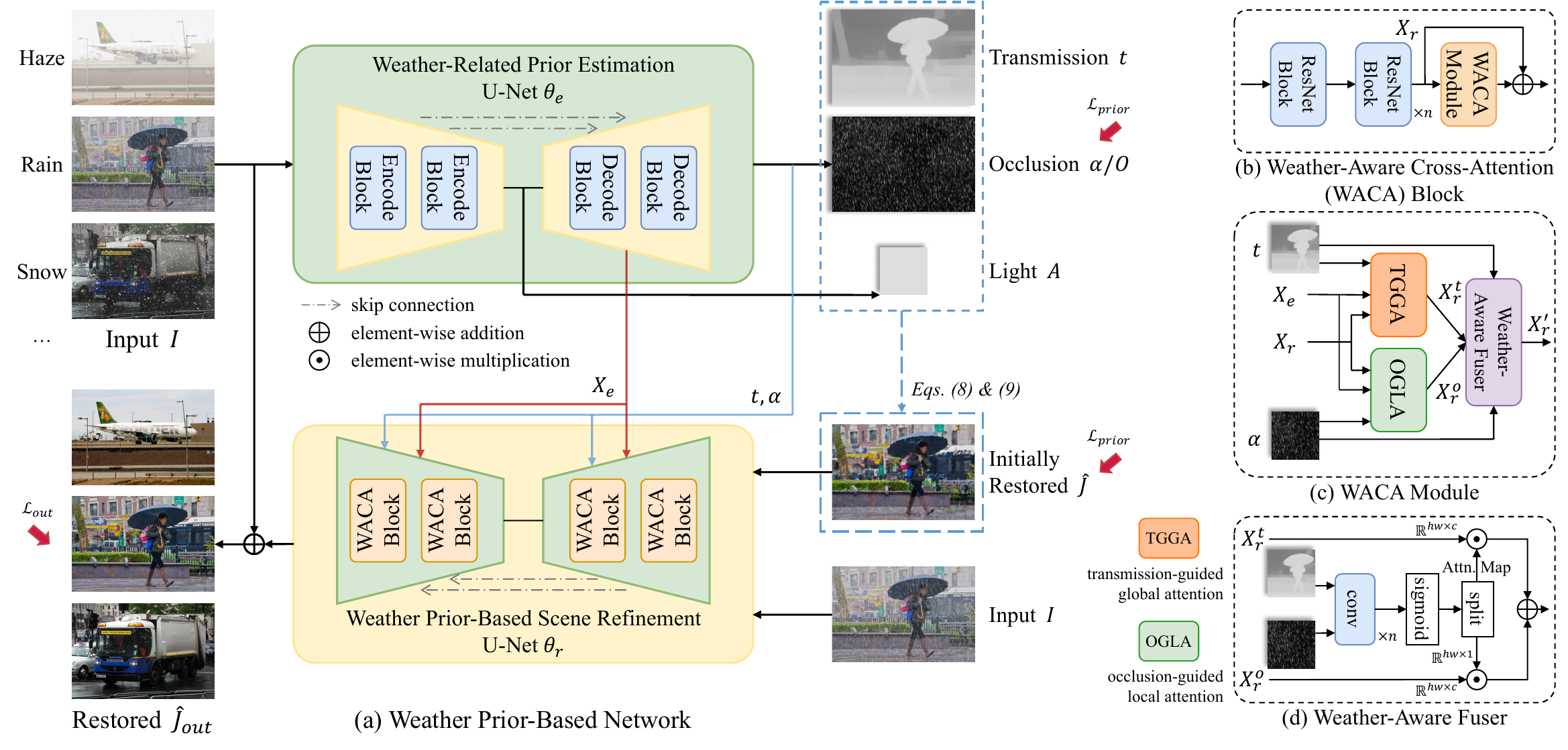}
    \caption{The overall architecture of our WeatherNet.
    (a) WeatherNet contains a weather-related prior estimation U-Net~$\theta_e$ and a weather prior-based scene refinement U-Net~$\theta_r$.
    (b) The weather-aware cross-attention (WACA) blocks are used in $\theta_r$ with ResNet blocks and a designed WACA module.
    (c) WACA module comprises TGGA, OGLA, and weather-aware fuser for prior-guided feature enhancement.
    (d) The components of weather-aware fuser.}
    \label{fig:weathernet}
\end{figure*}

\paragraph{Existing imaging models}
We briefly summarize the existing imaging models for haze, rain, and snow.
The formation of a haze image is described by the atmospheric scattering model~\cite{narasimhan2002vision,he2010single} as:
\begin{equation}
    \label{eq:scattering}
    I(x) = J(x) t(x) + A (1 - t(x)) \ ,
\end{equation}
where $I$ and $J$ are the observed haze image and clear scene radiance, $t$ and $A$ are the medium transmission and global atmospheric light, and $x$ is the pixel index.
The transmission $t(x) = e ^ {-\beta d(x)}$ describes the portion of the light scattering, which depends on the scene depth $d$ and the atmosphere scattering coefficient $\beta$.

The formation of a rain image can be simplified as an additive composite model~\cite{li2016rain} as:
\begin{equation}
    \label{eq:rain}
    \tilde{I}(x) = J(x) + O(x) \ ,
\end{equation}
where $\tilde{I}$ is the rain image degenerated by rain streaks, and $O$ is the color intensity map of sparse rain streaks.
Besides, $O$ can comprise one layer or multiple layers of rain streaks, \textit{i.e.}, $O=\Sigma_{l=1}^n O_l$ and $n$ is the number of layers.
When considering the scattering effects, the rain model~\cite{yang2017deep,li2019heavy} can be described as:
\begin{equation}
    \label{eq:rain-haze}
    I(x) = \tilde{I}(x) t(x) + A (1 - t(x)) \ ,
\end{equation}
where $I$ is the rain image degenerated by rain streaks and haze effects, $t$ and $A$ are transmission and light as in Eq.~(\ref{eq:scattering}), and $\tilde{I}$ is the scattering effect-free rain image as in Eq.~(\ref{eq:rain}).

Similarly, the formation of a snow image is described by an alpha matting model~\cite{liu2018desnownet} as:
\begin{equation}
    \label{eq:snow}
    \tilde{I}(x) = O(x) \alpha(x) + J(x) (1 - \alpha(x)) \ ,
\end{equation}
where $\tilde{I}$ is the snow image affected by snowflakes, $\alpha$ is the snow transparency, and $O$ is the snow color intensity.
When considering the scattering effects, the snow model~\cite{chen2020jstasr} is formulated as:
\begin{equation}
    \label{eq:snow-haze}
    I(x) = \tilde{I}(x) t(x) + A (1 - t(x)) \ ,
\end{equation}
where $I$ is the snow image affected by snowflakes and haze effects, $t$ and $A$ are transmission and light as in Eq.~(\ref{eq:scattering}), and $\tilde{I}$ is the scattering effect-free snow image as in Eq.~(\ref{eq:snow}).

Overall, the rain and snow imaging models follow a similar footprint in the recent deep learning-based studies, \textit{i.e.}, with only rain streaks or snowflakes in Eqs.~(\ref{eq:rain})~and~(\ref{eq:snow}), or with scattering effects in Eqs.~(\ref{eq:rain-haze})~and~(\ref{eq:snow-haze}), except for some detail deviations.

\begin{figure}
    \centering
    \includegraphics[width=\hsize]{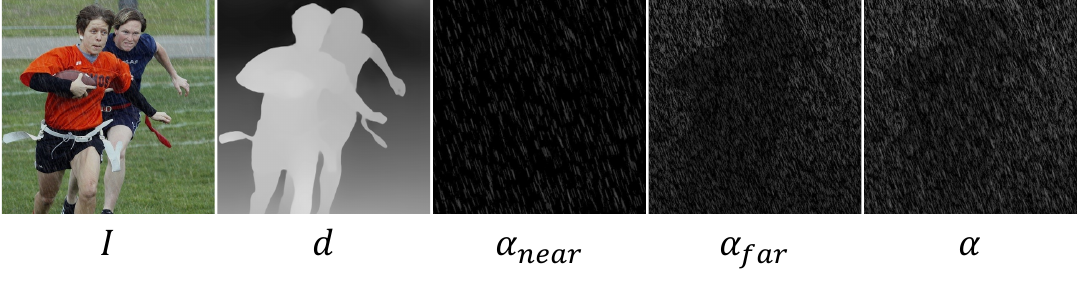}
    \caption{The volumetric effects $\alpha$ of rain and snow are considered for the near regions, $\alpha_{near}$, and far regions, $\alpha_{far}$, to synthesize the degraded image $I$, where $\alpha_{near}$ is computed with multiple layers based on depth $d$.}
    \label{fig:multiple_layers}
\end{figure}

\subsection{Our Formulation}
In this work, we regard an image taken under common adverse weather conditions, such as haze, rain, and snow, as a composition of haze scattering effects and (optionally) visible particle occlusions; see Fig.~\ref{fig:optical}.

First, we consider the haze scattering effects as follows:
\begin{equation}
    \label{eq:ours1}
    B(x) = J(x) t(x) + A (1 - t(x)) \ ,
\end{equation}
where $B$ is the background haze image affected by the aggregated scattering effects, similar to Eq.~\eqref{eq:scattering}.

Then, we consider the rain and snow visible occlusions:
\begin{equation}
    \label{eq:ours2}
    I(x) = O(x) \alpha(x) + B(x) (1 - \alpha(x)) \ ,
\end{equation}
where $I$ is the observed image under adverse weather conditions; $\alpha$ and $O$ denote the occlusion transparency and brightness, respectively.
Note that $O$ is usually more constant in a photo~\cite{liu2018desnownet}.

Unlike the approach presented in Hu \textit{et al.}'s work~\cite{hu2019depth}, which solely focuses on modeling a single rain layer, our method takes into account the volumetric effects of both visible rain and snow.
To capture these volumetric effects, we divide the volume containing visible particles in the far regions into multiple slender layers and combine their contributions, denoted as $\alpha_l$, to simulate the overall effect as follows:
\begin{equation}
\label{eq:ours3}
\begin{split}
\alpha(x) &= \alpha_{near}(x) + \alpha_{far}(x) \ , \\
\alpha_{far}(x) &= (1 - e^{-\beta d(x)}) \sum_{l=1}^{N} \alpha_{l} \ ,
\end{split}
\end{equation}
Here, $\alpha_{near}$ and $\alpha_{far}$ model the visible occlusions in the regions $0< z < z_1$ and $z_1 < z < z_2$, respectively, with $N$ representing the number of thin layers, as illustrated in Fig. \ref{fig:multiple_layers}.
Importantly, our approach results in more visible raindrops in distant regions characterized by larger scene depths, denoted as $d$. It is worth noting that $\alpha_{near}$ also consists of one or more layers, albeit without considering the variations in visibility with depth.

Our imaging model formulation includes some special cases:
(i)~When there are no visible rain and snow occlusions, \textit{i.e.}, $I(x) = J(x) t(x) + A (1 - t(x))$ with $\alpha=0$, which is the haze condition as in Eq.~\eqref{eq:scattering}.
(ii)~When the scattering effects are negligible, \textit{i.e.}, $I(x)=O(x) \alpha(x) + J(x) (1 - \alpha(x))$ with $t=1$, which corresponds to light rain and snow situations, as in Eqs.~\eqref{eq:rain}~and~\eqref{eq:snow}.
(iii)~In common rain and snow scenarios, particle occlusions and scattering effects are visible, and our overall formulation is similar to Eqs.~\eqref{eq:rain-haze}~and~\eqref{eq:snow-haze}.
In this way, our unified imaging model is able to cover most of the cases of haze, rain, and snow models as expressed in Eq.~\eqref{eq:scattering}~-~Eq.~\eqref{eq:snow-haze}.

\paragraph*{Discussion}
Our aim is not to present an accurate physics-based appearance modeling that applies to all adverse weather conditions.
In fact, the visual effects of bad weather conditions are complex, and the modeling is challenging.
The appearance of rain and snow is affected by additional factors, including rain and snow intensity, camera parameters (\textit{e.g.}, exposure time and depth of field), and scene brightness.
Instead, we aim to introduce an imaging model capturing the essential visual factors for common types of adverse weather from a more unified perspective, \textit{i.e.}, visible occlusions, and scattering effects.
These physical priors help to design image restoration methods considering both weather-general and weather-specific characteristics under the mixed weather scenarios, which is limited in the previous works~\cite{li2020all,valanarasu2022transweather,chen2022learning,ozdenizci2023restoring} and is more practical in the real world.

\section{Methodology}
\label{sec:network}

\begin{figure}
    \centering
    \includegraphics[width=0.95\hsize]{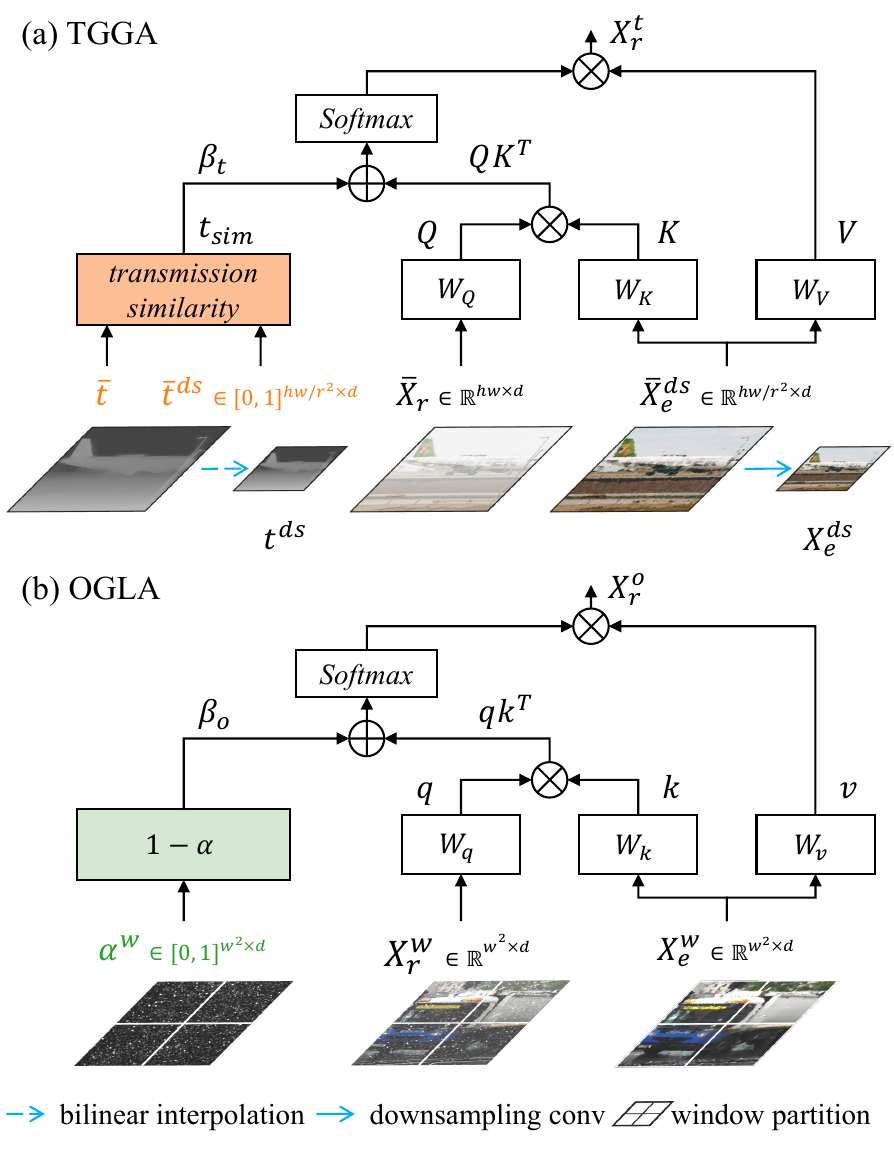}
    \caption{Transmission-guided global attention (TGGA) and occlusion-guided local attention (OGLA).
    (a)~TGGA performs global cross-attention with reduced key/value resolution and transmission guidance.
    (b)~OGLA adopts the window-based local cross-attention under the occlusion guidance.}
    \label{fig:tgaola}
\end{figure}

\subsection{Network Overview}
The overall architecture of our proposed weather prior-based network is illustrated in Fig.~\ref{fig:weathernet} (a), which is a two-stage framework with a weather-related prior estimation stage and a weather prior-based scene refinement stage.
Our end-to-end network receives the image under any type (including mixed types) of adverse weather conditions as the input and produces a clean image as the output.

Specifically, the degenerated image $I$ is first fed into the weather-related prior estimation stage, which aims to estimate the weather-related prior information, including transmission $t$, occlusion $\alpha$ and $O$, and atmospheric light $A$.
Therefore, the initially restored scene is obtained by solving $J$ from the Eqs.~(\ref{eq:ours1})~and~(\ref{eq:ours2}) as:
\begin{align}
    \hat{B}(x) &= (\ I(x) - O(x) \alpha(x)\ ) \ /\ (1 - \alpha(x)) \ ,\\
    \label{eq:ours4}
    \hat{J}(x) &= (\ \hat{B}(x) - A (1 - t(x))\ ) \ /\ t(x) \ ,
\end{align}
where $\hat{B}$ and $\hat{J}$ are the computed occlusion-free (haze) background and the underlying scene, respectively, based on the estimated information.
Then, the initially restored image $\hat{J}$ and the degenerated image $I$ together are fed into the weather prior-based scene refinement stage to produce the final artifact-free scene image.
Besides, the estimated prior information, \textit{i.e.}, $t$ and $\alpha$, and extracted features in the first estimation stage are also injected into the second stage to help scene recovery.
In detail, our designed Weather-Aware Cross-Attention (WACA) module enhances the refinement-stage features $X_r$ with the estimation-stage features $X_e$, which propagates the cross-stage features based on the cross-attention mechanism~\cite{vaswani2017attention} with the estimated transmission and occlusion priors.

The implementations of the two stages are built upon U-Net~\cite{ronneberger2015u,rombach2022high}, where the WACA blocks are used in the second stage; see Fig.~\ref{fig:weathernet}~(b) for details.
Moreover, in the first stage, we consider that the atmospheric light $A$ contains more global environment information, whereas the transmission $t$ and occlusion $\alpha$ estimation focus on pixel-wise information.
Hence, the prediction of $A$ is achieved by pooling the last feature map from the encoder of the estimation U-Net, while $t$ and $\alpha$ are obtained from the decoder, all with separate MLP-like prediction heads.

\subsection{Weather-Aware Cross-Attention}
\label{sec:waca}
The weather priors, \textit{e.g.}, transmission and occlusion, describe the weather-related degeneration as discussed earlier.
Hence, we design a weather-aware cross-attention (WACA) module to explore the weather prior-based information and features to enhance the scene recovery in Fig.~\ref{fig:weathernet} (c).

\subsubsection{Transmission-Guided Global Attention}
\label{sec:tgga}
The transmission $t$ describes the portion of the haze scattering effects and depends on the scene depth.
The transmission is global-wise information, where $t$ within a local patch is almost uniform~\cite{he2010single}, and some non-local patches in a scene have similar $t$ due to the similar depths.
Such areas suffer from similarly reduced visibility due to the haze effects.
Hence, the transmission-guided global attention (TGGA) enhances the refinement features at one location using the estimation stage's non-local features of similar transmission values; see Fig.~\ref{fig:tgaola}~(a).

Specifically, given the refinement feature map $X_r \in \mathbb{R}^{h \times w \times c}$ with spatial resolution $h \times w$ and feature channels $c$, the estimation feature map $X_e$, and the estimated transmission $t$, we first obtain the downsampled $X_e^{ds} \in \mathbb{R}^{h/r \times w/r \times c}$ and $t^{ds} \in \mathbb{R}^{h/r \times w/r \times 1}$ to a reduced spatial resolution, where $r$ is the downsampling ratio.
In this way, the more uniform transmission~\cite{he2010single} and its corresponding features are compressed into a patch-wise representation, which can save computation costs~\cite{chu2021twins}.
Then, $X_r$ is used as the query $Q = W_Q \bar{X}_r$, and $X_e^{ds}$ is used as the key and value $[K, V] = [W_K \bar{X}_e^{ds}, W_V \bar{X}_e^{ds}]$, where $Q \in \mathbb{R} ^ {hw \times d}$ and $K, V \in \mathbb{R} ^ {hw / r^2 \times d}$, $d$ is the dimension of query/key with multi-head attention used~\cite{vaswani2017attention}; $\bar{X}$~is the feature flattening operation.

To ensure the pixel benefits from the helpful features of similar transmissions, we first compute the pair-wise transmission similarity $t_{sim}(i, j) = 1 - {\lvert t_i - t_j \rvert} ^2$ between pixels $i \in t$ and $j \in t^{ds}$.
After that, the transmission-guided attention is computed as:
\begin{equation}
    Sim (Q, K) = Q K^T / \sqrt{d} + \beta_t t_{sim} \ ,
\end{equation}
where $\beta_t$ is a learnable scalar for transmission guidance.

\subsubsection{Occlusion-Guided Local Attention}
\label{sec:ogla}
The rain and snow present the particles that occlude local regions in the input image, and the affected textures in these local regions are highly correlated.
Thus, we design the occlusion-guided local attention (OGLA), which propagates the estimation features to refinement features in a localized manner by referring to the regions with less occlusion; see Fig.~\ref{fig:tgaola}~(b).

In detail, the refinement and estimation feature maps $X_r$ and $X_e$, and occlusion transparency $\alpha$ are first split into non-overlapping local windows~\cite{liu2021swin,liu2022swin} with window size $w$.
Then the efficient cross-attention is computed in each local window with the query $q = Wq X_r^w$ and the key and value $[k, v] = [W_k X_e^w, W_v X_e^w]$, where $q, k, v \in \mathbb{R}^{w^2 \times d}$, and $X^w$ denotes the window partition operation.

The occlusion transparency is leveraged to propagate helpful information from the visible regions by adjusting the aggregation weights in attention.
Accordingly, the occlusion-guided attention is computed as:
\begin{equation}
    Sim (q, k) = q k^T / \sqrt{d} + \beta_o (1 - \alpha ^ w) \ ,
\end{equation}
where $\beta_o$ is a learnable scalar for occlusion guidance.

\subsubsection{Weather-Aware Fuser}
The weather-aware fuser (WAF) is to integrate the enhanced features $X_r^t$ and $X_r^o$ from the TGGA and OGLA, respectively, with the guidance of transmission and occlusion.
Fig.~\ref{fig:weathernet}~(d) shows the detailed structure, where WAF takes the features $X_r^t$, $X_r^o$, estimated transmission $t$, and estimated occlusion~$\alpha$ as the inputs.
Then, the concatenated input is fed into a sequence of depthwise conv, 1~$\times$~1 convs, and a sigmoid activation to generate the fusion attention weights $a_t, a_o \in {[0, 1]}^{h \times w \times 1}$.
Finally, the fused feature $X_r'$ is obtained by aggregating the features with $X_r' = a_t \cdot X_t + a_o \cdot X_o$, where $\cdot$ is the element-wise multiplication.

\begin{figure}
    \centering
    \captionsetup[subfigure]{font=small,justification=centering}
    \begin{subfigure}{0.19\hsize}
        \includegraphics[width=\hsize]{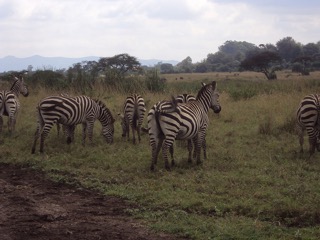}
    \end{subfigure}
    \begin{subfigure}{0.19\hsize}
        \includegraphics[width=\hsize]{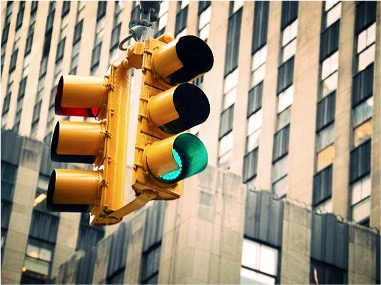}
    \end{subfigure}
    \begin{subfigure}{0.19\hsize}
        \includegraphics[width=\hsize]{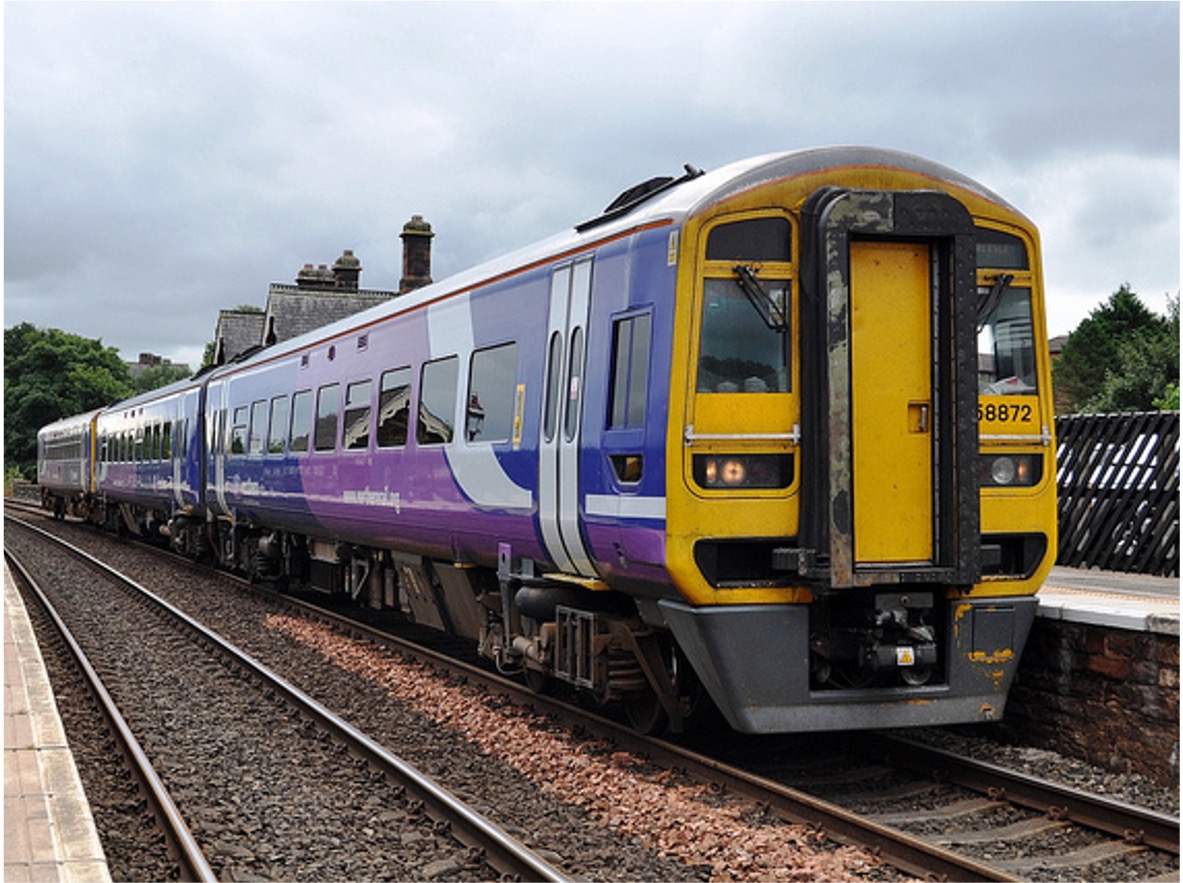}
    \end{subfigure}
    \begin{subfigure}{0.19\hsize}
        \includegraphics[width=\hsize]{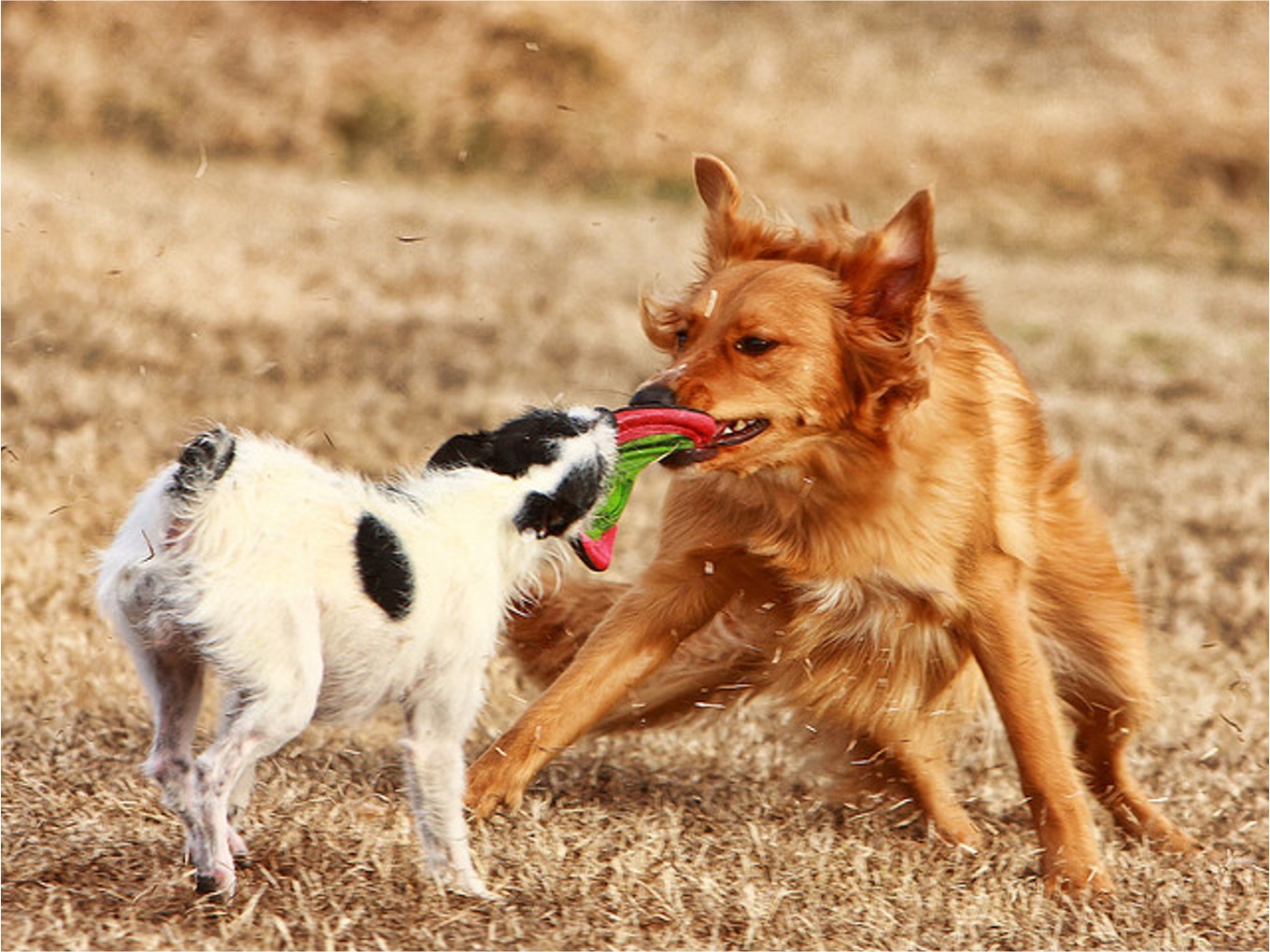}
    \end{subfigure}
    \begin{subfigure}{0.19\hsize}
        \includegraphics[width=\hsize]{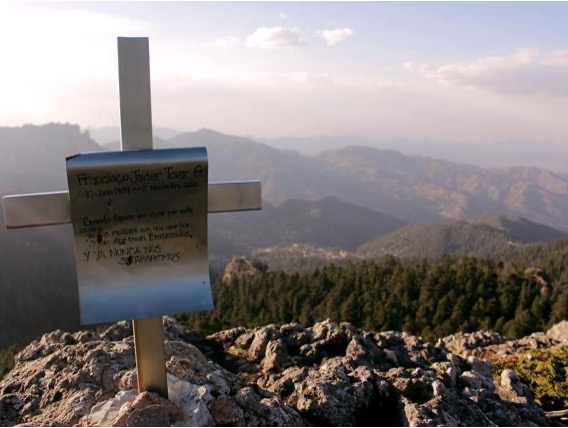}
    \end{subfigure}
    \\ \vspace{1pt}
    \begin{subfigure}{0.19\hsize}
        \includegraphics[width=\hsize]{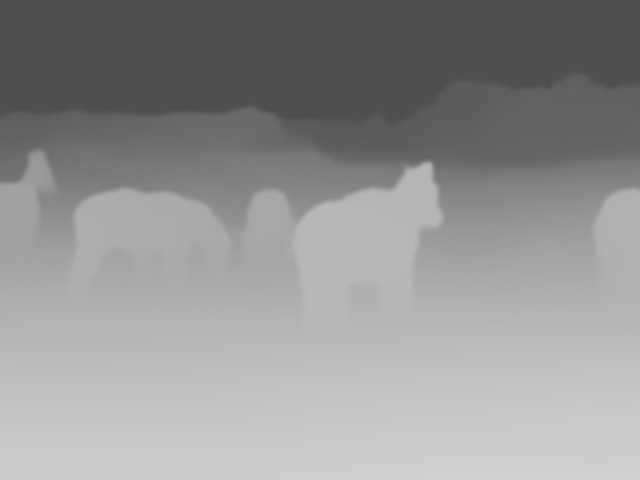}
    \end{subfigure}
    \begin{subfigure}{0.19\hsize}
        \includegraphics[width=\hsize]{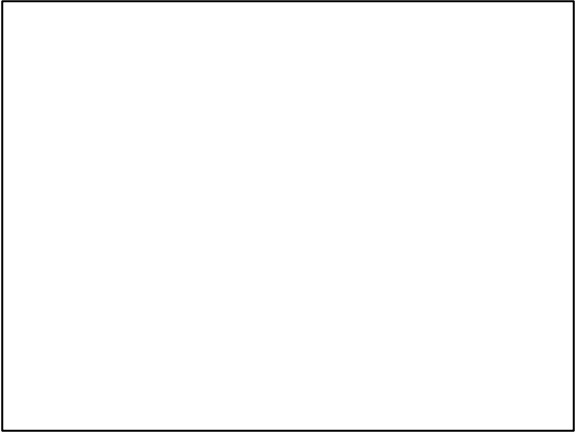}
    \end{subfigure}
    \begin{subfigure}{0.19\hsize}
        \includegraphics[width=\hsize]{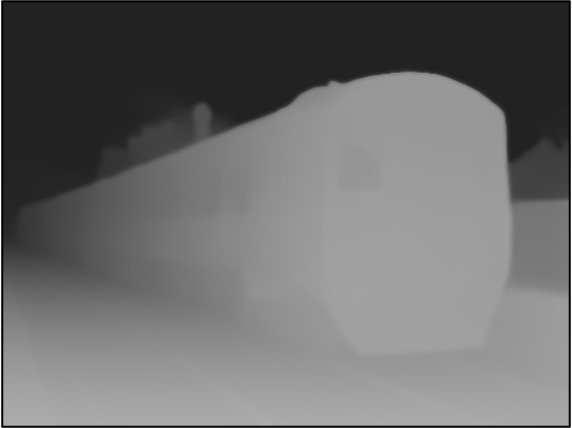}
    \end{subfigure}
    \begin{subfigure}{0.19\hsize}
        \includegraphics[width=\hsize]{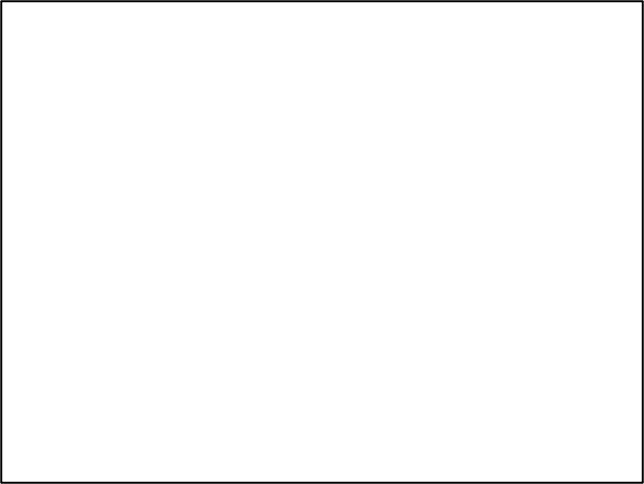}
    \end{subfigure}
    \begin{subfigure}{0.19\hsize}
        \includegraphics[width=\hsize]{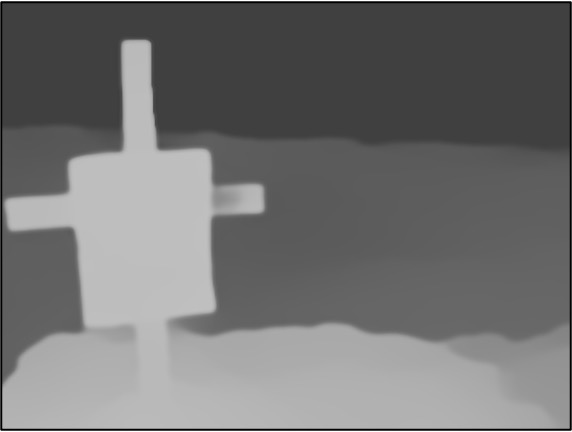}
    \end{subfigure}
    \\ \vspace{1pt}
    \begin{subfigure}{0.19\hsize}
        \includegraphics[width=\hsize]{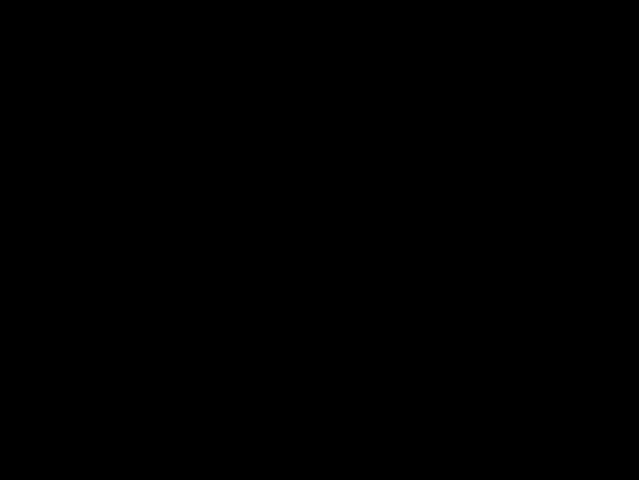}
    \end{subfigure}
    \begin{subfigure}{0.19\hsize}
        \includegraphics[width=\hsize]{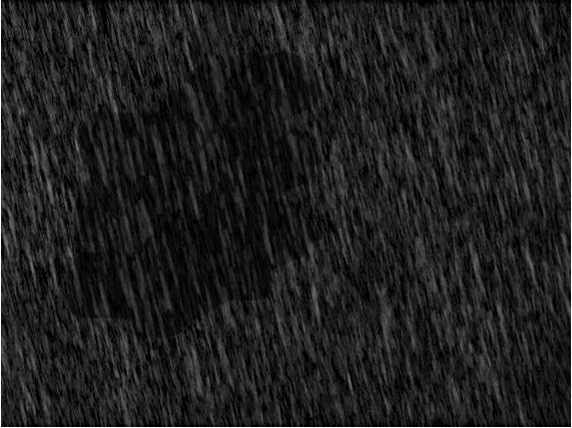}
    \end{subfigure}
    \begin{subfigure}{0.19\hsize}
        \includegraphics[width=\hsize]{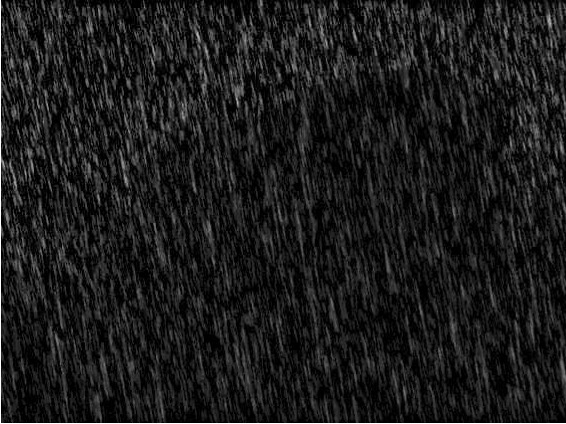}
    \end{subfigure}
    \begin{subfigure}{0.19\hsize}
        \includegraphics[width=\hsize]{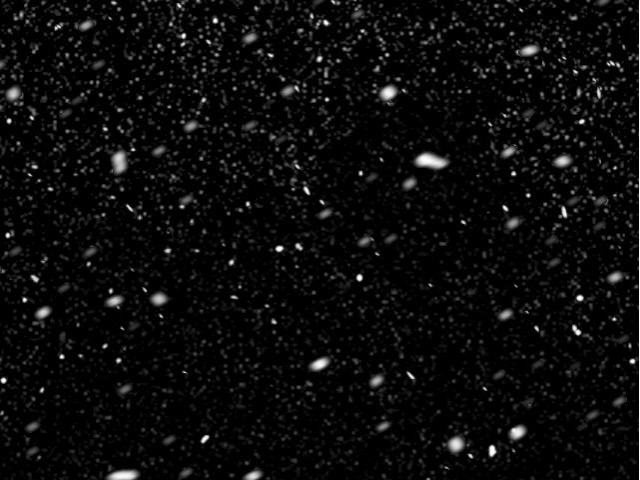}
    \end{subfigure}
    \begin{subfigure}{0.19\hsize}
        \includegraphics[width=\hsize]{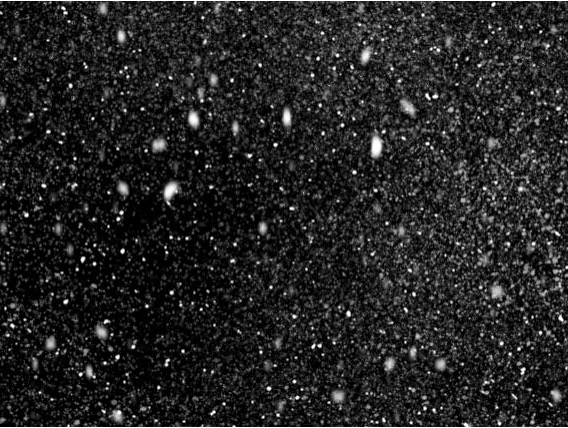}
    \end{subfigure}
    \\ \vspace{1pt}
    \begin{subfigure}{0.19\hsize}
        \includegraphics[width=\hsize]{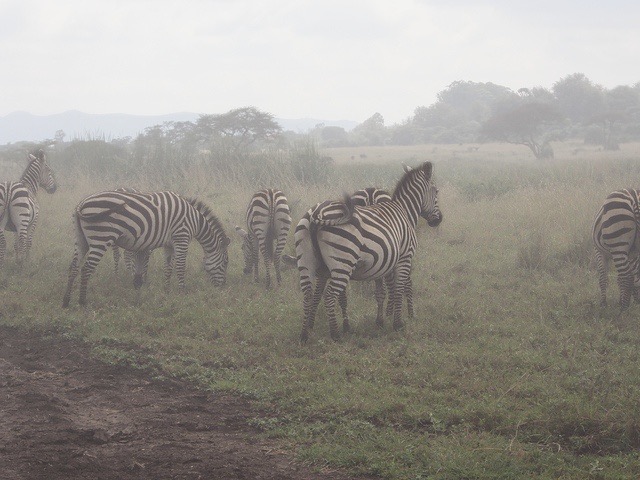}
        \caption{Haze \\ \hfill}
    \end{subfigure}
    \begin{subfigure}{0.19\hsize}
        \includegraphics[width=\hsize]{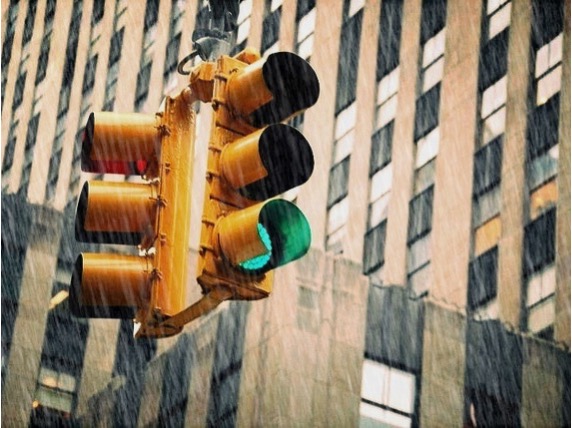}
        \caption{Rain \\ \hfill}
    \end{subfigure}
    \begin{subfigure}{0.19\hsize}
        \includegraphics[width=\hsize]{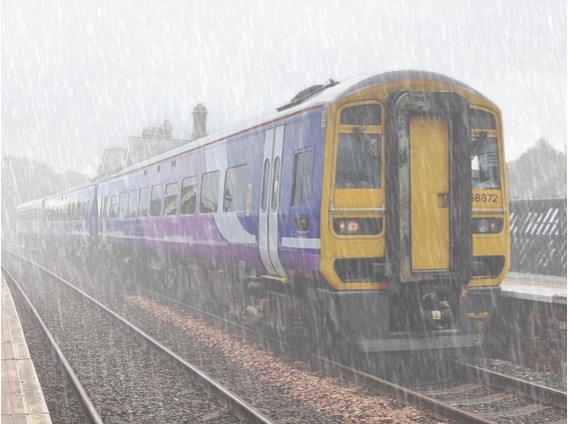}
        \caption{Rain w/ haze}
    \end{subfigure}
    \begin{subfigure}{0.19\hsize}
        \includegraphics[width=\hsize]{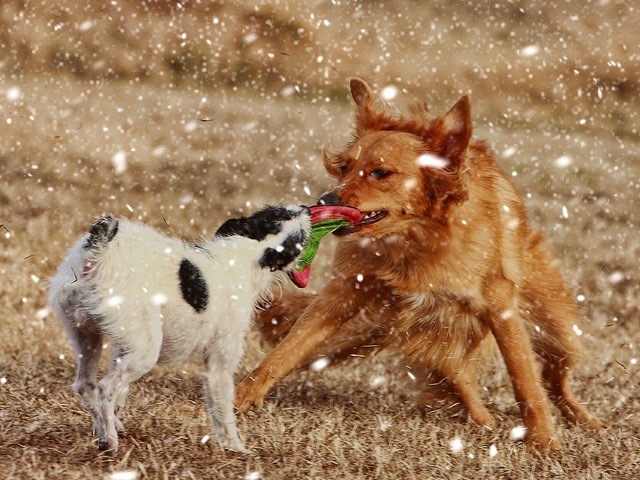}
        \caption{Snow \\ \hfill}
    \end{subfigure}
    \begin{subfigure}{0.19\hsize}
        \includegraphics[width=\hsize]{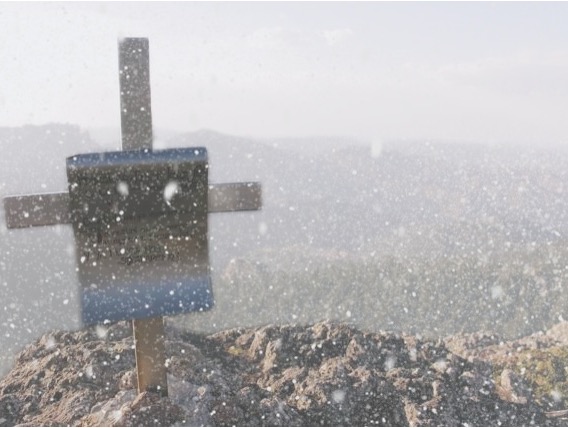}
        \caption{Snow w/ haze}
    \end{subfigure}
    \\
    \caption{Samples of our Weather30K dataset, shown with ground truth images, transmission maps, occlusion maps, and synthetic degenerated images from top to bottom.
    Note that there are no scattering effects with transmission maps $t=1$ for the simulated light (b) rain and (d) snow images, and there are no visible occlusions in (a) haze images.
    }
    \label{fig:vis_weather30k}
\end{figure}

\begin{table}[tp]
\centering
\caption{
Quantitative comparisons on our Weather30K (the first experimental setting).
The best and second-best results are marked using \textbf{bold} and {\ul underline}, respectively.
}
\label{tab:setting1}
\begin{adjustbox}{width=\hsize}
\begin{tabular}{c|c|ccc}
\toprule
\multirow{2}{*}{Type}                                                     & \multirow{2}{*}{Method}                        & \multicolumn{3}{c}{PSNR $\uparrow$ / SSIM $\uparrow$}                                            \\ \cline{3-5} 
                                                                          &                                                & Haze                           & Rain                           & Snow                           \\ \midrule
\multirow{2}{*}{\begin{tabular}[c]{@{}c@{}}Mixed\\ Training\end{tabular}} & Restormer~\cite{zamir2022restormer}            & 23.43/0.9055                   & 27.49/0.9049                   & 24.61/0.8343                   \\
                                                                          & NAFNet~\cite{chen2022simple}                   & 22.26/0.8946                   & {\ul 27.70}/0.9131             & 24.32/0.8401                   \\ \hline
\multirow{4}{*}{\begin{tabular}[c]{@{}c@{}}Multi-\\ Task\end{tabular}}    & TransWeather~\cite{valanarasu2022transweather} & 25.11/0.9152                   & 27.19/0.9051                   & 24.37/0.8258                   \\
                                                                          & Two-Stage~\cite{chen2022learning}              & 25.51/{\ul 0.9282}             & 27.55/0.9141                   & 24.86/{\ul 0.8476}             \\
                                                                          & WeatherDiff~\cite{ozdenizci2023restoring}      & 23.59/0.8962                   & 26.40/0.9047                   & 23.95/0.8350                   \\
                                                                          & WGWS-Net~\cite{zhu2023learning}                & 25.42/0.9240                   & {\ul27.60}/{\ul 0.9159}             & {\ul 25.24}/0.8452             \\
                                                                          & MWFormer~\cite{zhu2024mwformer}                & {\ul 25.59}/0.9172             & 27.56/0.9068                   & 25.00/0.8424                    \\
                                                                          & Our Method                                     & \textbf{27.31}/\textbf{0.9342} & \textbf{28.84}/\textbf{0.9252} & \textbf{26.34}/\textbf{0.8653} \\ \bottomrule
\end{tabular}
\end{adjustbox}
\end{table}

\begin{figure*}
    \centering
    \captionsetup[subfigure]{labelformat=empty,justification=centering}
    \begin{subfigure}{0.137\textwidth}
        \includegraphics[width=\textwidth]{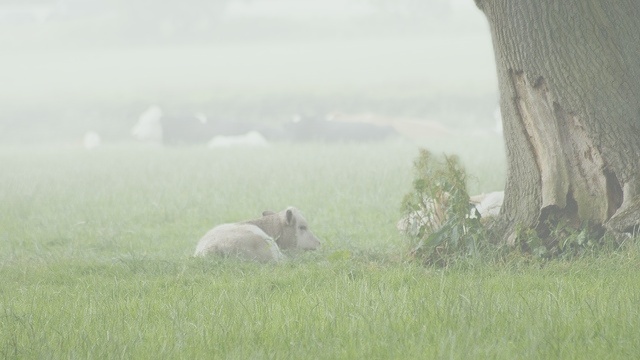}
    \end{subfigure}
    \begin{subfigure}{0.137\textwidth}
        \includegraphics[width=\textwidth]{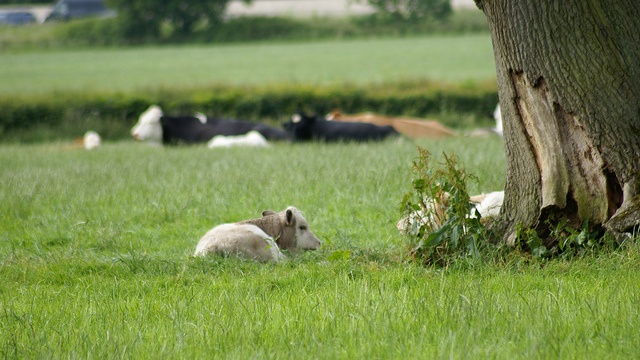}
    \end{subfigure}
    \begin{subfigure}{0.137\textwidth}
        \includegraphics[width=\textwidth]{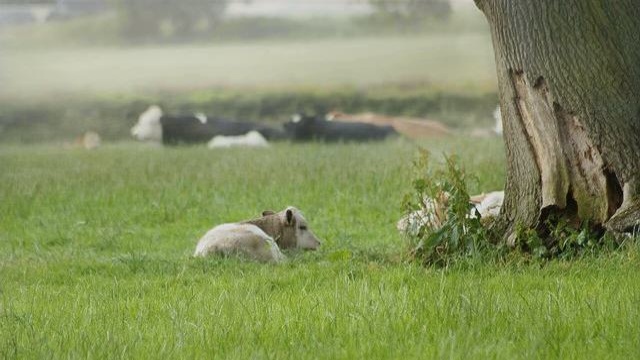}
    \end{subfigure}
    \begin{subfigure}{0.137\textwidth}
        \includegraphics[width=\textwidth]{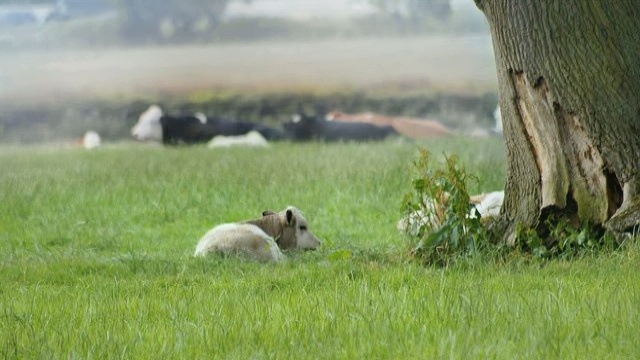}
    \end{subfigure}
    \begin{subfigure}{0.137\textwidth}
        \includegraphics[width=\textwidth]{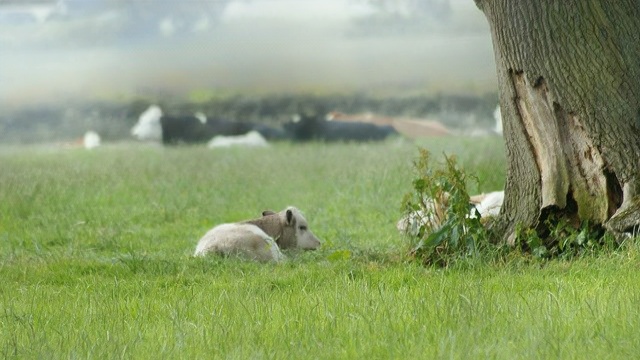}
    \end{subfigure}
    \begin{subfigure}{0.137\textwidth}
        \includegraphics[width=\textwidth]{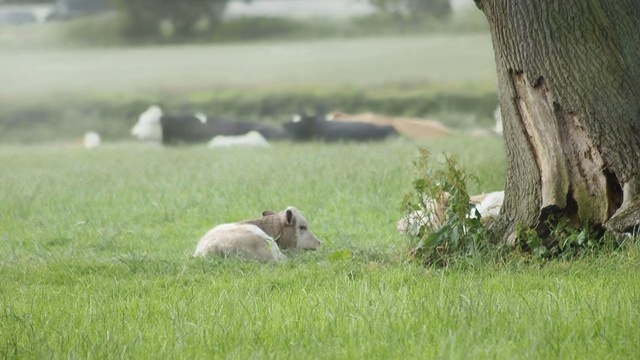}
    \end{subfigure}
    \begin{subfigure}{0.137\textwidth}
        \includegraphics[width=\textwidth]{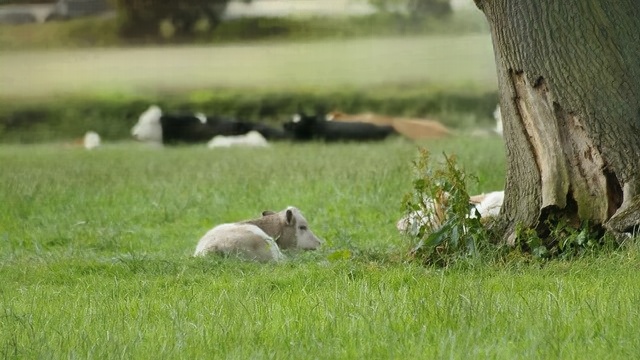}
    \end{subfigure}
    \\ \vspace{2pt}
    \begin{subfigure}{0.137\textwidth}
        \includegraphics[width=\textwidth]{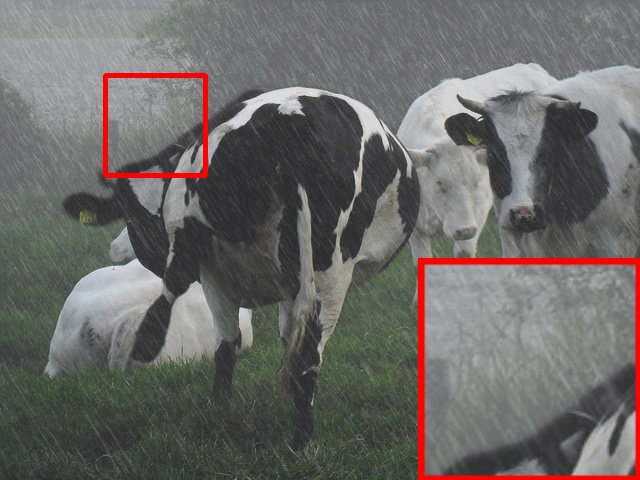}
    \end{subfigure}
    \begin{subfigure}{0.137\textwidth}
        \includegraphics[width=\textwidth]{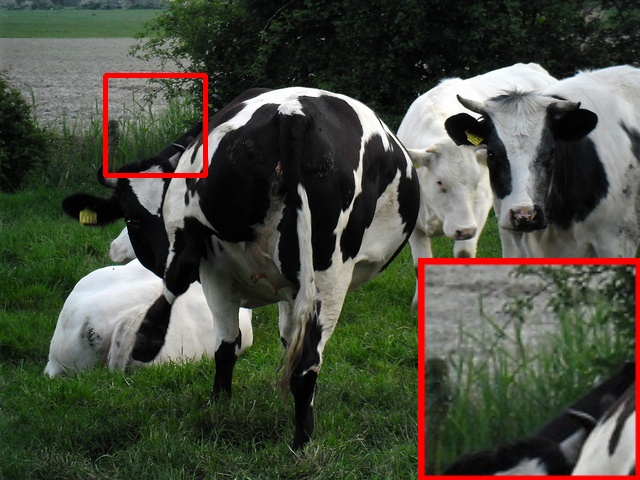}
    \end{subfigure}
    \begin{subfigure}{0.137\textwidth}
        \includegraphics[width=\textwidth]{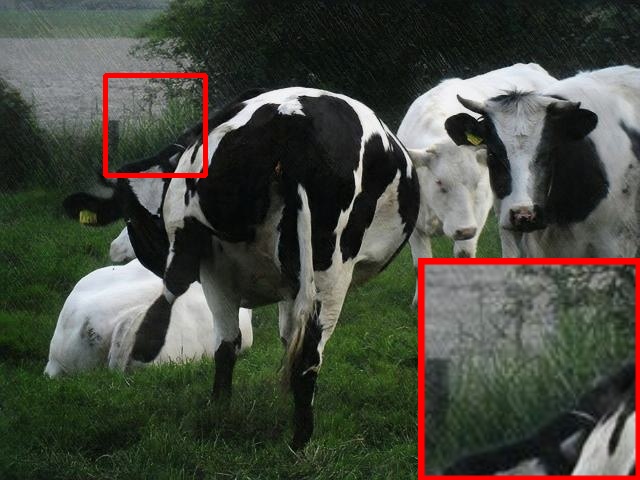}
    \end{subfigure}
    \begin{subfigure}{0.137\textwidth}
        \includegraphics[width=\textwidth]{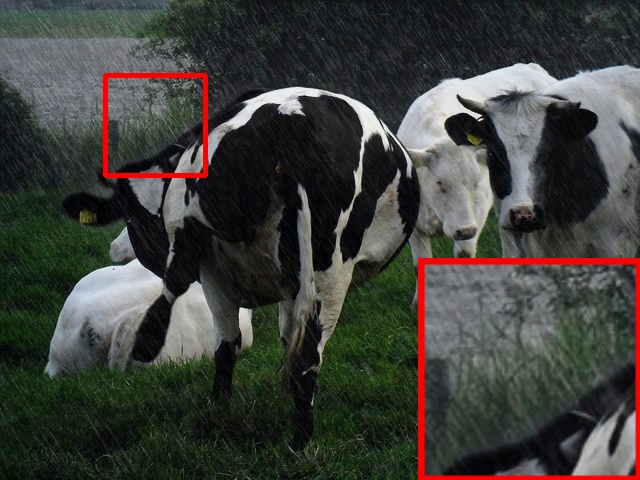}
    \end{subfigure}
    \begin{subfigure}{0.137\textwidth}
        \includegraphics[width=\textwidth]{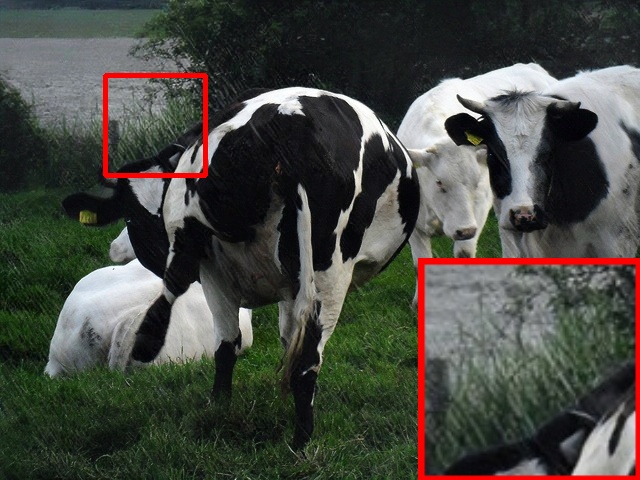}
    \end{subfigure}
    \begin{subfigure}{0.137\textwidth}
        \includegraphics[width=\textwidth]{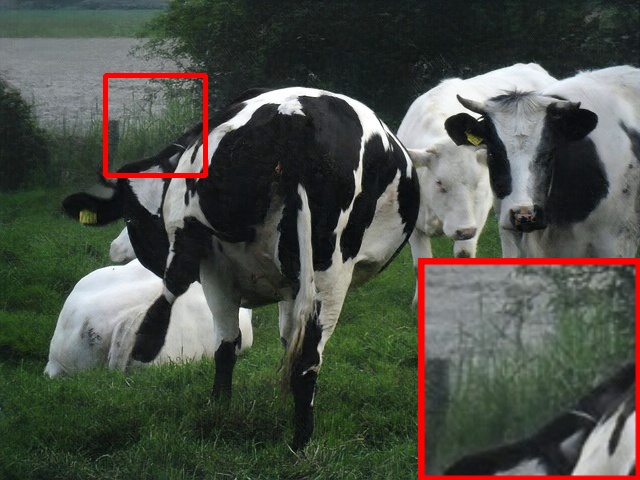}
    \end{subfigure}
    \begin{subfigure}{0.137\textwidth}
        \includegraphics[width=\textwidth]{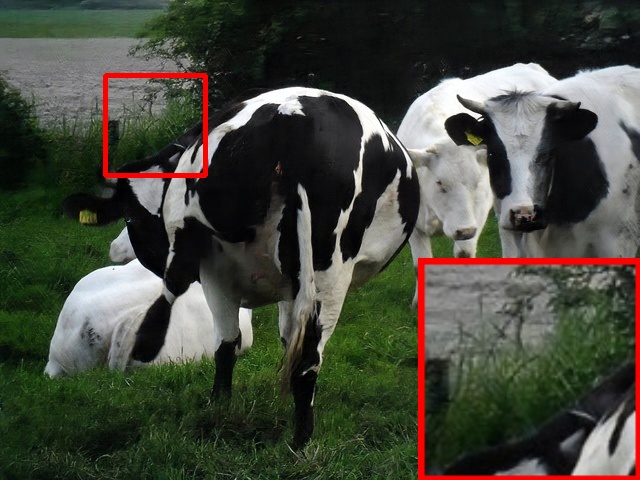}
    \end{subfigure}
    \\ \vspace{2pt}
    \begin{subfigure}{0.137\textwidth}
        \includegraphics[width=\textwidth]{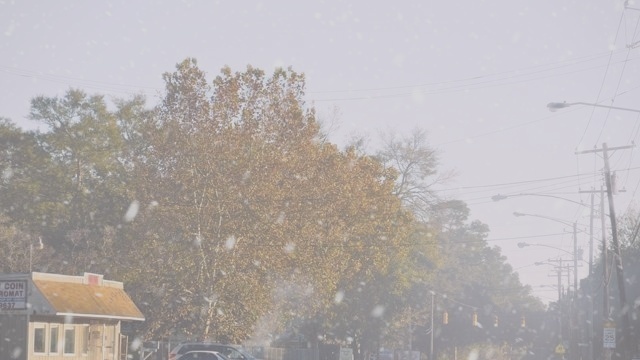}
    \end{subfigure}
    \begin{subfigure}{0.137\textwidth}
        \includegraphics[width=\textwidth]{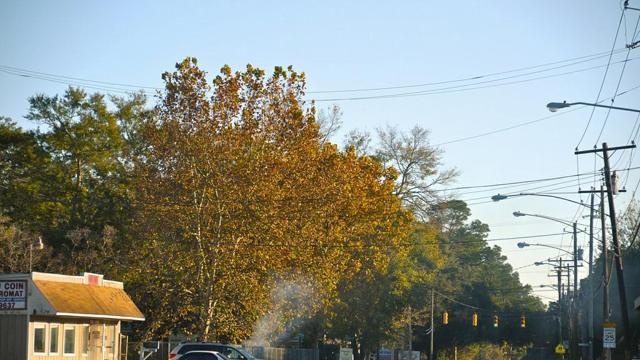}
    \end{subfigure}
    \begin{subfigure}{0.137\textwidth}
        \includegraphics[width=\textwidth]{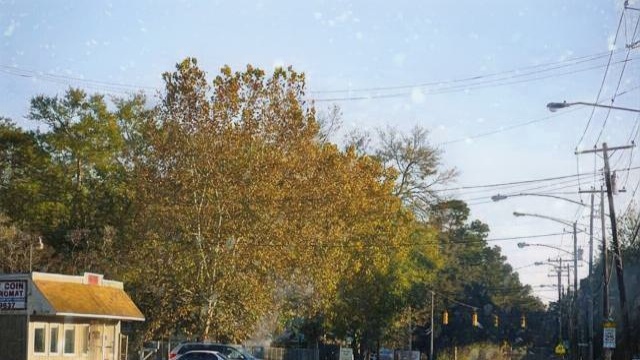}
    \end{subfigure}
    \begin{subfigure}{0.137\textwidth}
        \includegraphics[width=\textwidth]{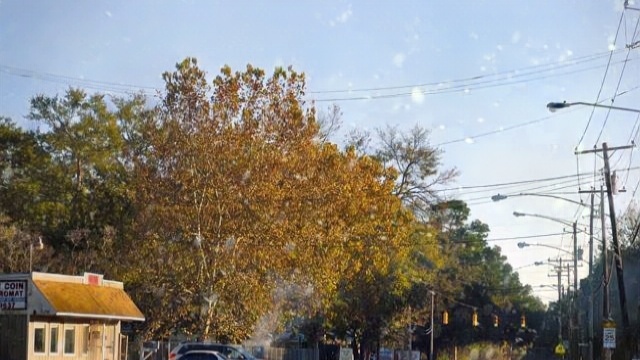}
    \end{subfigure}
    \begin{subfigure}{0.137\textwidth}
        \includegraphics[width=\textwidth]{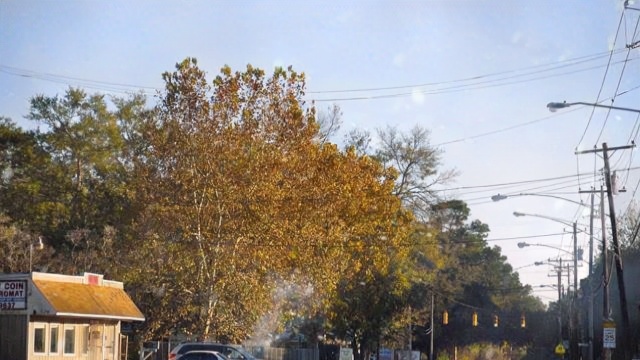}
    \end{subfigure}
    \begin{subfigure}{0.137\textwidth}
        \includegraphics[width=\textwidth]{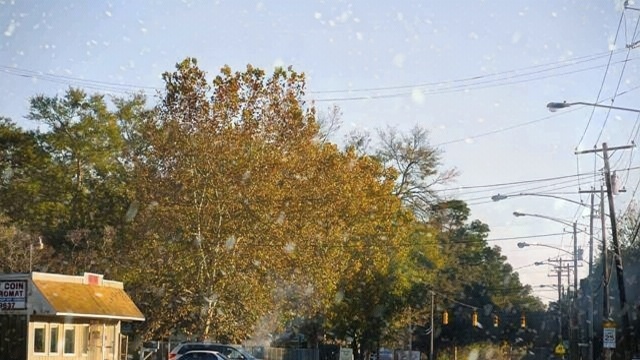}
    \end{subfigure}
    \begin{subfigure}{0.137\textwidth}
        \includegraphics[width=\textwidth]{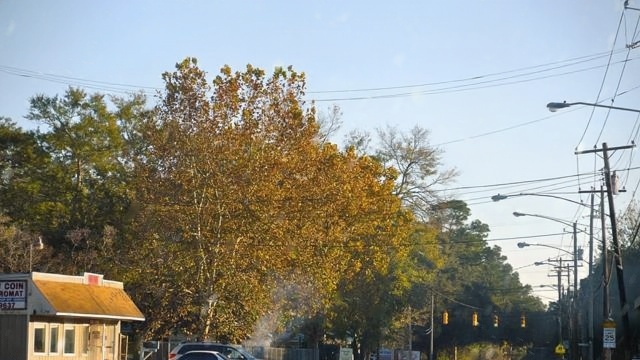}
    \end{subfigure}
    \\ \vspace{2pt}
    \begin{subfigure}{0.137\textwidth}
        \includegraphics[width=\textwidth]{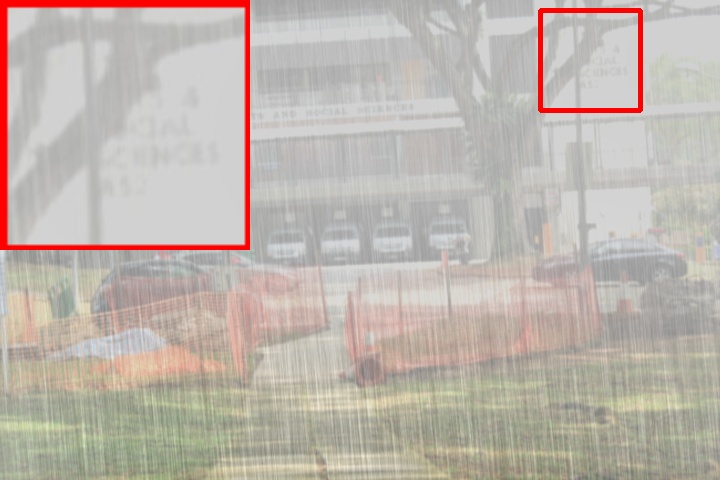}
        \caption{Input}
    \end{subfigure}
    \begin{subfigure}{0.137\textwidth}
        \includegraphics[width=\textwidth]{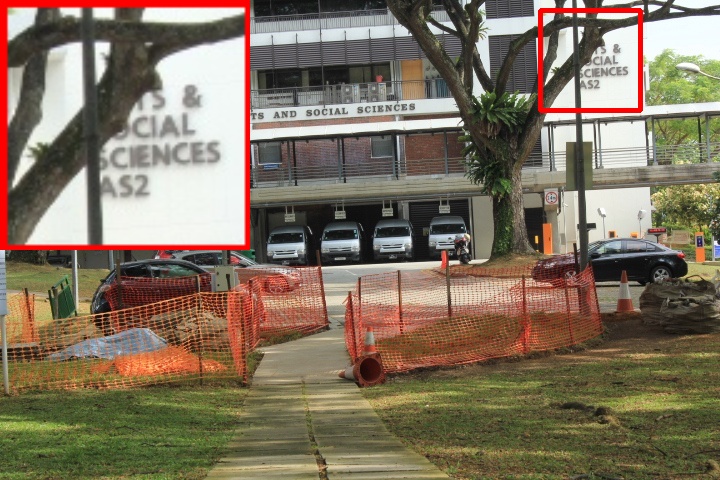}
        \caption{Ground truth}
    \end{subfigure}
    \begin{subfigure}{0.137\textwidth}
        \includegraphics[width=\textwidth]{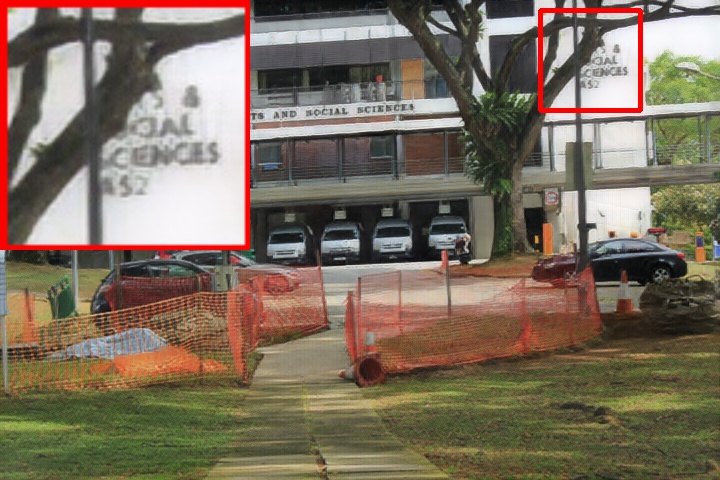}
        \caption{TransWeather~\cite{valanarasu2022transweather}}
    \end{subfigure}
    \begin{subfigure}{0.137\textwidth}
        \includegraphics[width=\textwidth]{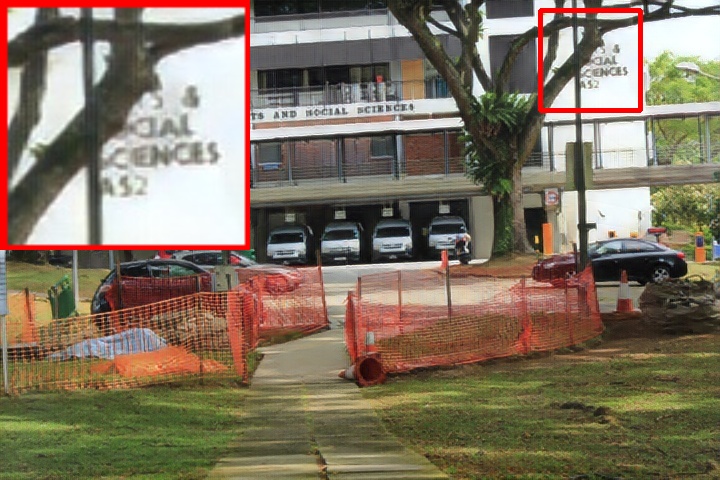}
        \caption{Two-Stage~\cite{chen2022learning}}
    \end{subfigure}
    \begin{subfigure}{0.137\textwidth}
        \includegraphics[width=\textwidth]{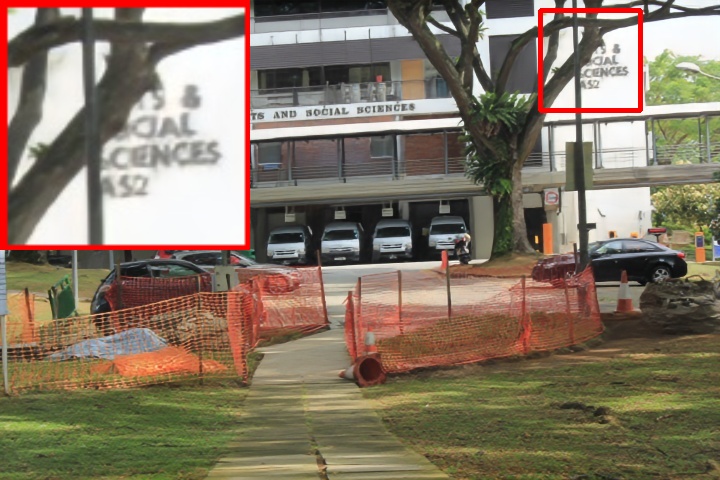}
        \caption{WGWS-Net~\cite{zhu2023learning}}
    \end{subfigure}
    \begin{subfigure}{0.137\textwidth}
        \includegraphics[width=\textwidth]{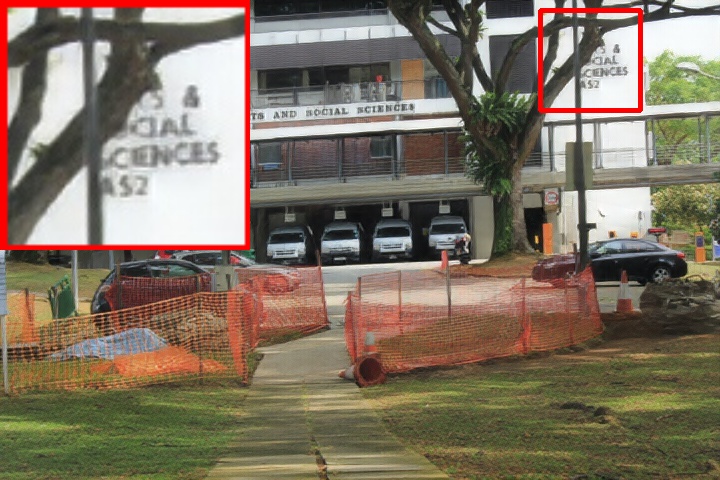}
        \caption{MWFormer \cite{zhu2024mwformer}}
    \end{subfigure}
    \begin{subfigure}{0.137\textwidth}
        \includegraphics[width=\textwidth]{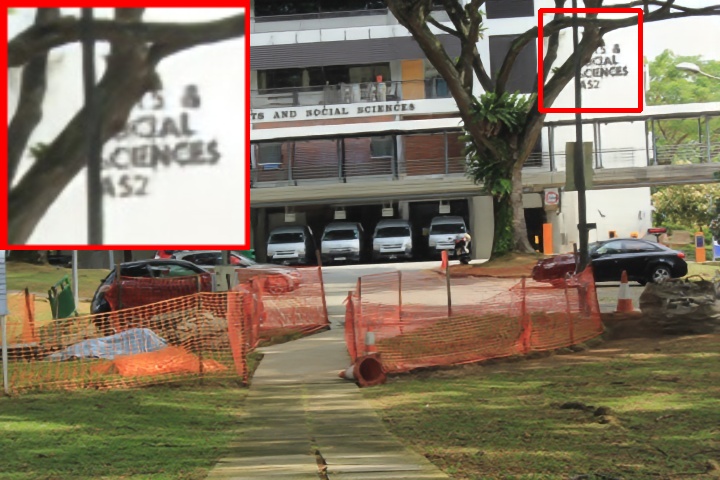}
        \caption{Our method}
    \end{subfigure}
    \\
    \caption{\textbf{Visual comparisons on the synthetic datasets.}
    The first three rows are from our Weather30K dataset, while the last row is from the Rain-Haze dataset~\cite{li2019heavy}.
    Please zoom in for a better view.
    }
    \label{fig:vis_synthetic}
\end{figure*}

\subsection{Training Strategies}
We optimize our network by minimizing the following loss $\mathcal{L}$, with an output loss $\mathcal{L}_{out}$ and a weather prior-based loss $\mathcal{L}_{prior}$:
\begin{equation}
    \label{eq:loss}
    \mathcal{L} = \mathcal{L}_{out} + \lambda_{prior} \mathcal{L}_{prior} \ ,
\end{equation}
where $\lambda_{prior}$ is a weighting parameter.

The $\mathcal{L}_1$ loss is used to supervise the final output $\hat{J}_{out}$ with the ground truth $J$ as $\mathcal{L}_{out} = \mathcal{L}_1 (\hat{J}_{out}, J)$.
Further, to ensure the estimation stage learns the weather priors, \textit{i.e.}, transmission $\hat{t}$, occlusion $\hat{\alpha}$, light $\hat{A}$, and the initially restored scene $\hat{J}$ from Eq.~(\ref{eq:ours4}), the prior loss is computed as
$\mathcal{L}_{prior} = \mathcal{L}_1(\hat{t}, t) + \mathcal{L}_1(\hat{\alpha}, \alpha) + \mathcal{L}_1(\hat{A}, A) + \mathcal{L}_1(\hat{J}, J)$
with the corresponding ground truths $t$, $\alpha$, $A$, and $J$, respectively.
Moreover, since $t$, $\alpha$, and $A$ are not always available in the existing datasets, we only calculate the corresponding loss in $\mathcal{L}_{prior}$ when the ground truths are available using a simple masking operation.

\section{Experimental Results}
\label{sec:experiment}

In this section, we first describe the experimental settings, including the introduction of our proposed Weather30K dataset. We then compare WeatherNet with state-of-the-art approaches. Next, we present ablation studies to analyze the observed imaging priors and the design choices of the proposed method. Additionally, we demonstrate the practical applications of the multi-weather image restoration approach. Finally, we conclude with a summary and discussion.

\subsection{Datasets and Implementation Details}

\begin{table*}[tp]
\centering
\caption{Quantitative comparisons on Rain-Haze~\cite{li2019heavy}, RainDrop~\cite{qian2018attentive}, and Snow100K-L~\cite{liu2018desnownet} of the second experimental setting.}
\label{tab:setting2}
\begin{adjustbox}{width=\hsize}
\begin{tabular}{c|c|cccc|cccc|cc}
\cmidrule[\heavyrulewidth]{1-4} \cmidrule[\heavyrulewidth]{6-8} \cmidrule[\heavyrulewidth]{10-12}
\multirow{2}{*}{Type}                                                    & \multirow{2}{*}{Method}                        & \multicolumn{2}{c}{Rain~\cite{li2019heavy}} &  & \multirow{2}{*}{Method}                        & \multicolumn{2}{c}{RainDrop~\cite{qian2018attentive}} &  & \multirow{2}{*}{Method}                        & \multicolumn{2}{c}{Snow~\cite{liu2018desnownet}} \\ \cline{3-4} \cline{7-8} \cline{11-12} 
                                                                         &                                                & PSNR $\uparrow$      & SSIM $\uparrow$      &  &                                                & PSNR $\uparrow$           & SSIM $\uparrow$           &  &                                                & PSNR $\uparrow$         & SSIM $\uparrow$        \\ \cmidrule[\arrayrulewidth]{1-4} \cmidrule[\arrayrulewidth]{6-8} \cmidrule[\arrayrulewidth]{10-12} 

\multirow{2}{*}{\begin{tabular}[c]{@{}c@{}}Task\\ Specific\end{tabular}} & HRGAN~\cite{li2019heavy}                       & 21.56                & 0.8550               &  & Attn. GAN~\cite{qian2018attentive}             & 30.55                     & 0.9023                    &  & DesnowNet~\cite{liu2018desnownet}              & 27.17                   & 0.8983                 \\
                                                                         & MPRNet~\cite{zamir2021multi}                   & 21.90                & 0.8456               &  & IDT~\cite{xiao2022image}                       & {\ul 31.87}               & {\ul 0.9313}              &  & DDMSNET~\cite{zhang2021deep}                   & 28.85                   & 0.8772                 \\ \cline{1-4} \cline{6-8} \cline{10-12} 
\multirow{5}{*}{\begin{tabular}[c]{@{}c@{}}Multi-\\ Task\end{tabular}}   & All-in-One~\cite{li2020all}                    & 24.71                & 0.8980               &  & All-in-One~\cite{li2020all}                    & 31.12                     & 0.9268                    &  & All-in-One~\cite{li2020all}                    & 28.33                   & 0.8820                 \\
                                                                         & TransWeather~\cite{valanarasu2022transweather} & 28.83                & 0.9000               &  & TransWeather~\cite{valanarasu2022transweather} & 30.17                     & 0.9157                    &  & TransWeather~\cite{valanarasu2022transweather} & 29.31                   & 0.8879                 \\
                                                                         & Two-Stage~\cite{chen2022learning}              & 29.98                & 0.8931               &  & Two-Stage~\cite{chen2022learning}              & {\ul 31.82}               & 0.9295                    &  & Two-Stage~\cite{chen2022learning}              & 28.93                   & 0.8899                 \\
                                                                         & WeatherDiff~\cite{ozdenizci2023restoring}      & 29.64                & \textbf{0.9312}      &  & WeatherDiff~\cite{ozdenizci2023restoring}      & 30.71                     & {\ul 0.9312}              &  & WeatherDiff~\cite{ozdenizci2023restoring}      & 30.09                   & 0.9041                 \\
                                                                         & WGWS-Net~\cite{zhu2023learning}                & 29.06                & 0.9126               &  & WGWS-Net~\cite{zhu2023learning}                & 31.35                     & 0.9228                    &  & WGWS-Net~\cite{zhu2023learning}                & 29.57                   & 0.8931                 \\
                                                                         & MWFormer~\cite{zhu2024mwformer}                & {\ul 30.24}          & 0.9111               &  & MWFormer~\cite{zhu2024mwformer}                & 31.73                     & 0.9254                    &  & MWFormer~\cite{zhu2024mwformer}                & {\ul 30.70}             & {\ul 0.9060}           \\
                                                                         & Our Method                                     & \textbf{31.32}       & {\ul 0.9309}         &  & Our Method                                     & \textbf{32.67}            & \textbf{0.9420}           &  & Our Method                                     & \textbf{32.14}          & \textbf{0.9266}        \\ \cmidrule[\heavyrulewidth]{1-4} \cmidrule[\heavyrulewidth]{6-8} \cmidrule[\heavyrulewidth]{10-12} 
\end{tabular}
\end{adjustbox}
\end{table*}

\paragraph{Datasets}
Existing outdoor evaluation datasets~\cite{li2020all,valanarasu2022transweather,chen2022learning,ozdenizci2023restoring} for single network-based methods fail to consider the common weather properties in different weather conditions.
In detail, some works~\cite{li2020all,valanarasu2022transweather,ozdenizci2023restoring} do not consider the fog effects in snow images, while Chen~\textit{et al.}~\cite{chen2022learning} neglect these effects in rain situations.
Therefore, we create a synthetic dataset by utilizing our imaging formulation with Eqs.~(\ref{eq:ours1})~-~(\ref{eq:ours3}), which comprehensively considers the scattering effects in both rain and snow.
Specifically, the clean scene images for haze, rain, and snow are selected from OTS~\cite{li2018benchmarking}, Rain13K~\cite{jiang2020multi}, and Snow100K~\cite{liu2018desnownet}, respectively, where the duplicated images are removed to avoid redundancy.
We also collect scene images from Cityscapes~\cite{cordts2016cityscapes} and COCO~\cite{lin2014microsoft} to build our final dataset.
MiDaS~\cite{ranftl2020towards} is used to estimate depths when the information is unavailable.
Then, following~\cite{sakaridis2018semantic}, we choose the scattering coefficient $\beta \in [0.005, 0.03]$ and atmospheric light $A \in [0.75, 1]$ in Eq.~(\ref{eq:ours1}) to synthesize the (background) haze image.
The occlusion transparency maps $\alpha$, \textit{i.e.}, rain streaks and snowflakes, are generated based on Eq.~(\ref{eq:ours3}).
We synthesize the scattering effects in half of the rain and snow images and keep the light rain and snow situations with only the occlusion artifacts in the other half.
Besides, the low-light effects~\cite{lv2021attention} commonly observed in adverse weather conditions are also considered for adjusting the scene images.
Finally, our dataset (Weather30K) contains 30,000 images, with each 9,000 training and 1,000 testing images for haze, rain, and snow, respectively, with transmission, light, and occlusion information available; see samples in Fig.~\ref{fig:vis_weather30k}.
The Weather30K dataset is publicly available\footnote{\url{https://github.com/jiaqixuac/WeatherNet}}.

We use our Weather30K for training and testing to serve as the first experimental setting.
Furthermore, we include two additional benchmark datasets for evaluation.
Specifically, following~\cite{li2020all,valanarasu2022transweather,ozdenizci2023restoring},
the second setting consists of Rain-Haze~\cite{li2019heavy}, RainDrop~\cite{qian2018attentive}, and Snow100K-L~\cite{liu2018desnownet} for training and testing.
Additionally, following~\cite{chen2022learning}, the third setting includes RESIDE~\cite{li2018benchmarking}, Rain1400~\cite{fu2017removing}, and CSD~\cite{chen2021all}.
To ensure a fair comparison, we use the same training sets as the compared methods, \textit{i.e.}, the 18,069 training images\footnote{\url{https://github.com/jeya-maria-jose/TransWeather}} used by TransWeather \cite{valanarasu2022transweather} for the second setting and images\footnote{\href{https://github.com/fingerk28/Two-stage-Knowledge-For-Multiple-Adverse-Weather-Removal}{https://github.com/fingerk28/Two-stage-Knowledge-For-Multiple...}} from Two-Stage \cite{chen2022learning} for the third setting.
For the second and third settings, we retrieve or compute the corresponding transmission and occlusion components, if possible, from their original datasets and calculate the prior loss only when the ground truths for these components are available.

\paragraph{Implementation Details}
We utilize the AdamW optimizer and the cosine annealing schedule.
The initial learning rate is set to 2e-4.
The number of training iterations is 160K for our Weather30K.
The batch size is 32, and the patch size is 256$\times$256.
The hyper-parameter $\lambda_{prior}$ in Eq.~(\ref{eq:loss}) is empirically set to 0.1.
We implement our models based on BasicSR\footnote{\url{https://github.com/XPixelGroup/BasicSR}} and train the models on two RTX 3090 GPUs.

\subsection{Comparisons with the State-of-the-Art Methods}

\begin{table}[tp]
\centering
\caption{Quantitative comparisons on RESIDE~\cite{li2018benchmarking}, Rain1400~\cite{fu2017removing}, and CSD~\cite{chen2021all} of the third experimental setting.}
\label{tab:setting3}
\begin{adjustbox}{width=\hsize}
\begin{tabular}{c|c|ccc}
\toprule
\multirow{2}{*}{Type}                                                     & \multirow{2}{*}{Method}                        & \multicolumn{3}{c}{PSNR $\uparrow$ / SSIM $\uparrow$}                                      \\ \cline{3-5} 
                                                                          &                                                & Haze~\cite{li2018benchmarking} & Rain~\cite{fu2017removing}   & Snow~\cite{chen2021all}      \\ \midrule
\multirow{4}{*}{\begin{tabular}[c]{@{}c@{}}Mixed\\ Training\end{tabular}} & AECR-Net~\cite{wu2021contrastive}              & 32.26/0.97                     & 30.43/0.91                   & 27.07/0.92                   \\
                                                                          & JRJG~\cite{ye2021closing}                      & 30.51/0.91                     & 28.92/0.89                   & 28.48/0.86                   \\
                                                                          & HDCW-Net~\cite{chen2021all}                    & 30.07/0.93                     & 27.20/0.85                   & 28.85/0.89                   \\
                                                                          & MPRNet~\cite{zamir2021multi}                   & 29.38/0.95                     & 31.36/0.91                   & 29.68/0.94                   \\ \hline
\multirow{4}{*}{\begin{tabular}[c]{@{}c@{}}Multi-\\ Task\end{tabular}}    & All-in-One~\cite{li2020all}                    & 30.49/0.95                     & 30.82/0.90                   & 28.65/0.92                   \\
                                                                          & TransWeather~\cite{valanarasu2022transweather} & 31.15/0.97                     & 32.20/0.92                   & 31.19/0.94                   \\
                                                                          & Two-Stage~\cite{chen2022learning}              & {\ul 33.95}/{\ul 0.98}         & {\ul 33.13}/{\ul 0.93}       & 31.35/0.95                   \\
                                                                          & WSWG-Net~\cite{zhu2023learning}                & 31.62/0.98                     & 32.93/0.92                   & {\ul 32.70}/{\ul 0.96}       \\
                                                                          & MWFormer~\cite{zhu2024mwformer}                & 30.73/0.97                     & 32.61/0.92                   & 32.52/0.95       \\
                                                                          & Our Method                                     & \textbf{36.26}/\textbf{0.99}   & \textbf{33.41}/\textbf{0.94} & \textbf{36.07}/\textbf{0.97} \\ \bottomrule
\end{tabular}
\end{adjustbox}
\end{table}

\begin{figure*}
    \centering
    \captionsetup[subfigure]{labelformat=empty,justification=centering}
    \begin{subfigure}{0.137\textwidth}
        \includegraphics[width=\hsize]{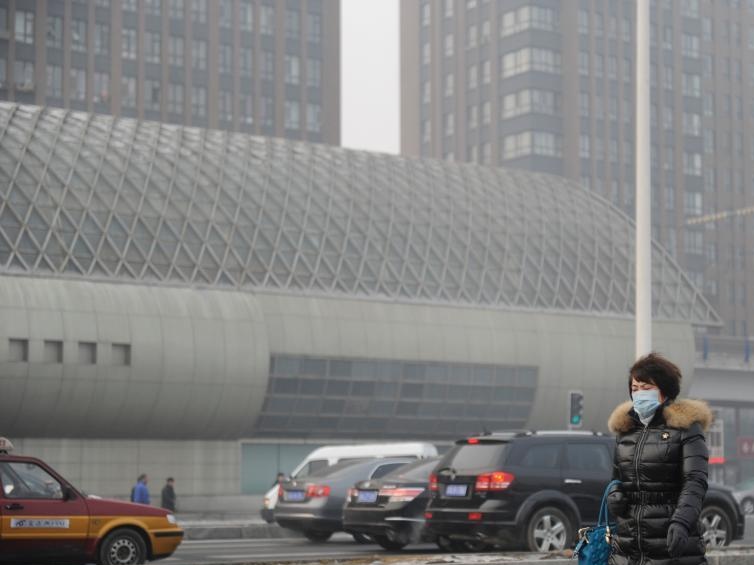}
    \end{subfigure}
    \begin{subfigure}{0.137\textwidth}
        \includegraphics[width=\hsize]{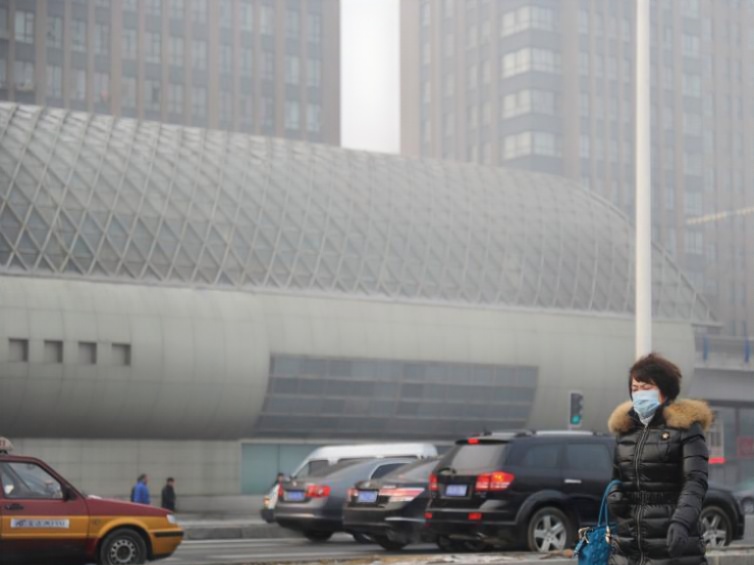}
    \end{subfigure}
    \begin{subfigure}{0.137\textwidth}
        \includegraphics[width=\hsize]{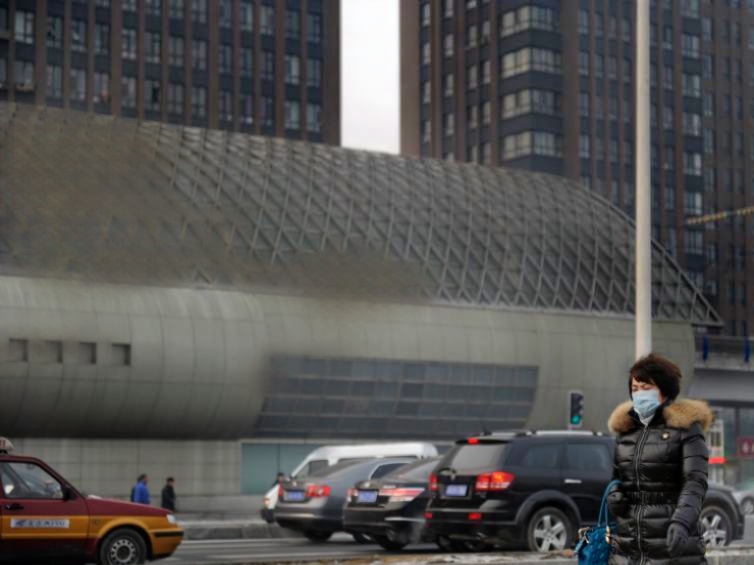}
    \end{subfigure}
    \begin{subfigure}{0.137\textwidth}
        \includegraphics[width=\hsize]{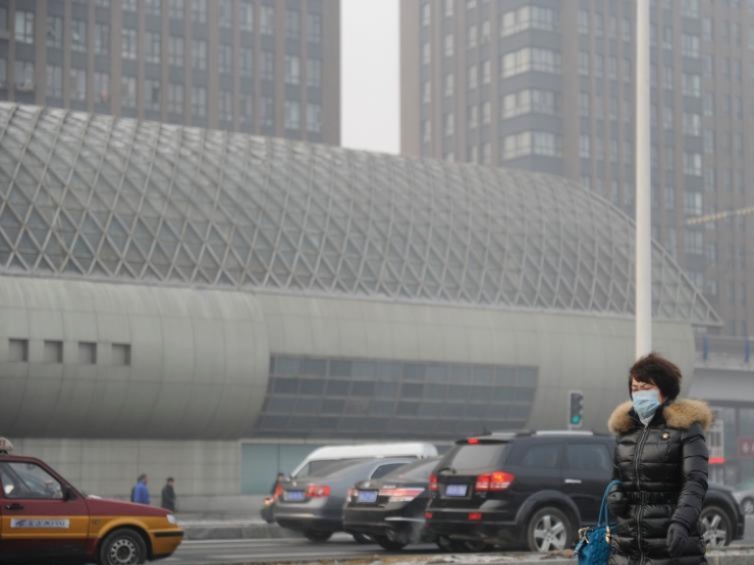}
    \end{subfigure}
    \begin{subfigure}{0.137\textwidth}
        \includegraphics[width=\hsize]{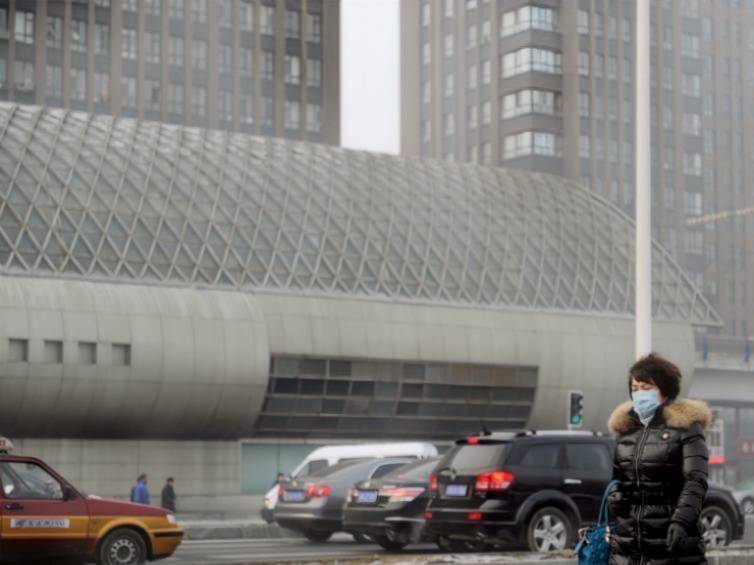}
    \end{subfigure}
    \begin{subfigure}{0.137\textwidth}
        \includegraphics[width=\hsize]{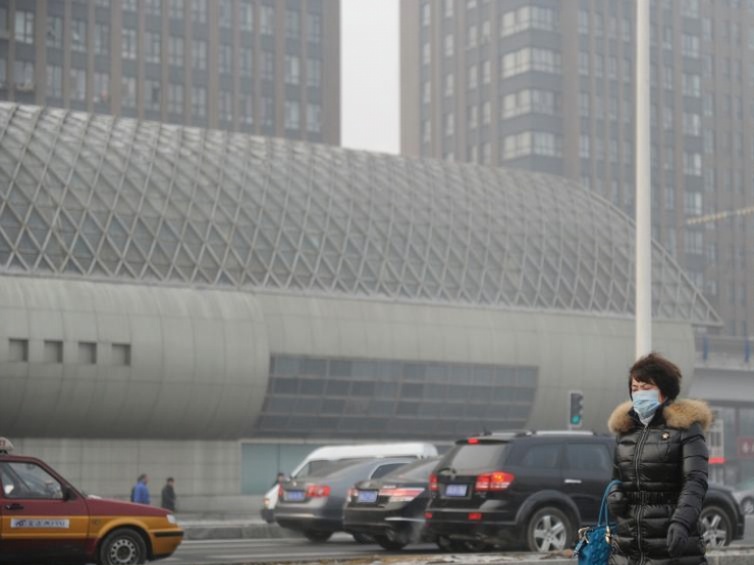}
    \end{subfigure}
    \begin{subfigure}{0.137\textwidth}
        \includegraphics[width=\hsize]{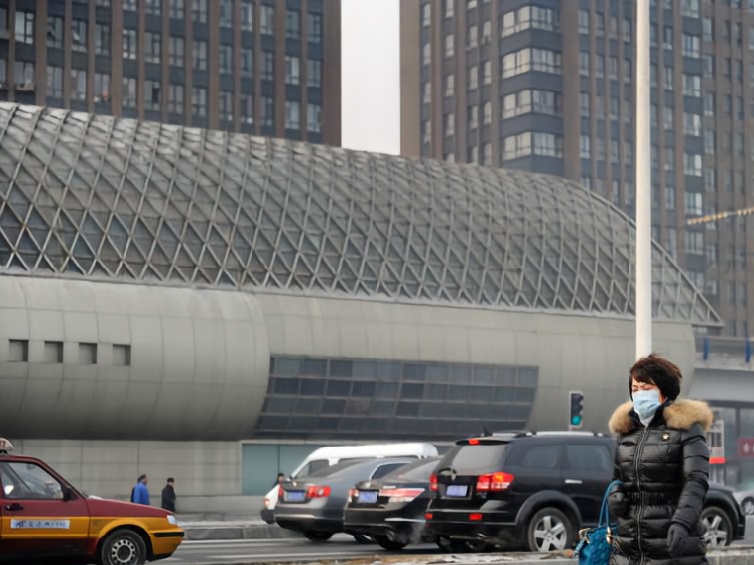}
    \end{subfigure}
    \\ \vspace{2pt}
    \begin{subfigure}{0.137\textwidth}
        \includegraphics[width=\hsize]{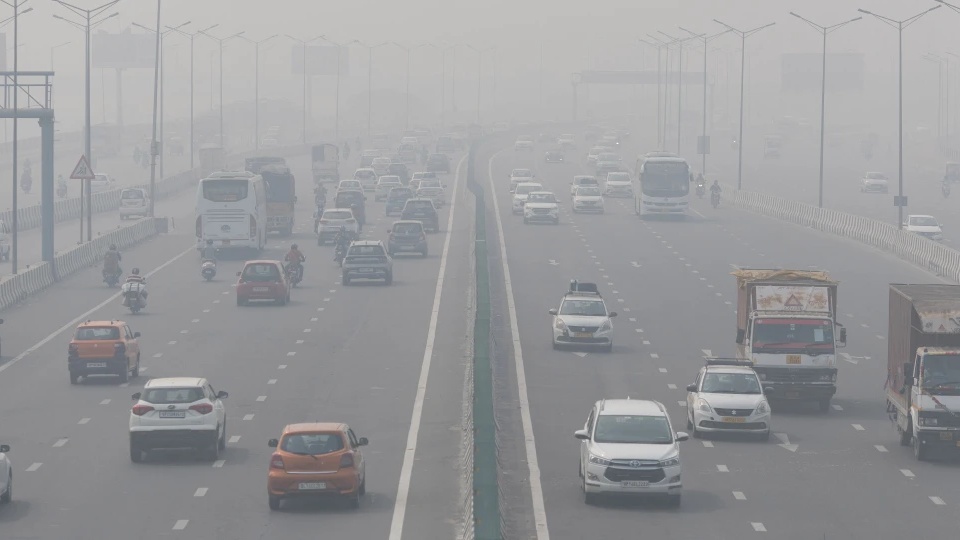}
    \end{subfigure}
    \begin{subfigure}{0.137\textwidth}
        \includegraphics[width=\hsize]{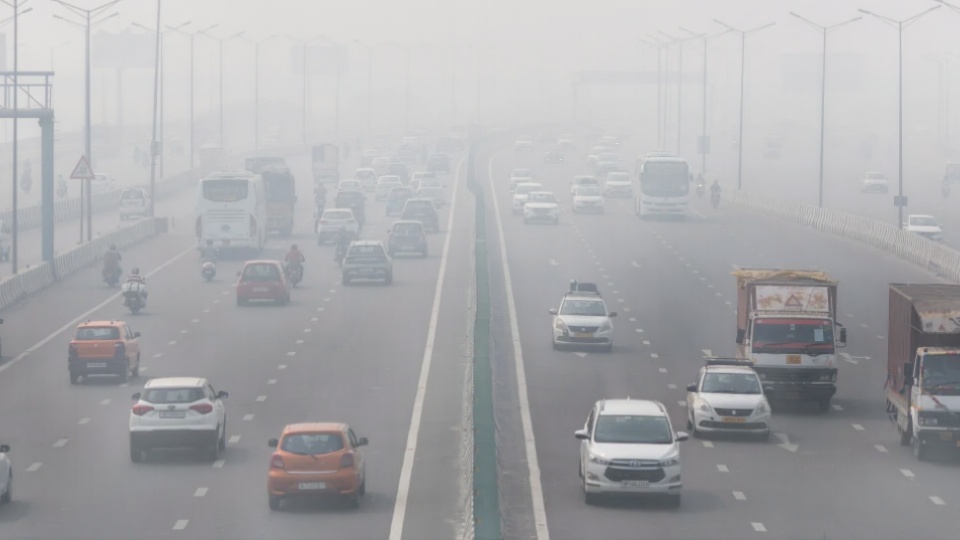}
    \end{subfigure}
    \begin{subfigure}{0.137\textwidth}
        \includegraphics[width=\hsize]{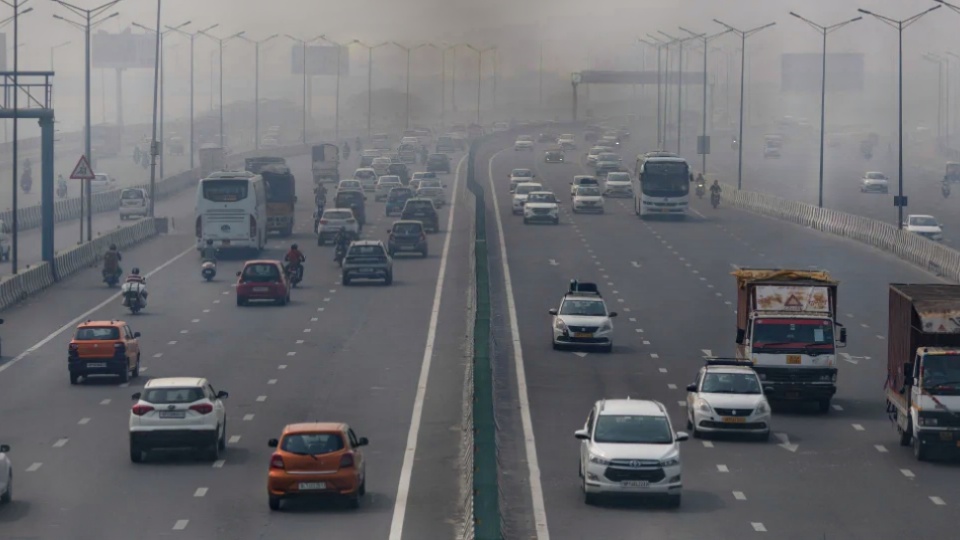}
    \end{subfigure}
    \begin{subfigure}{0.137\textwidth}
        \includegraphics[width=\hsize]{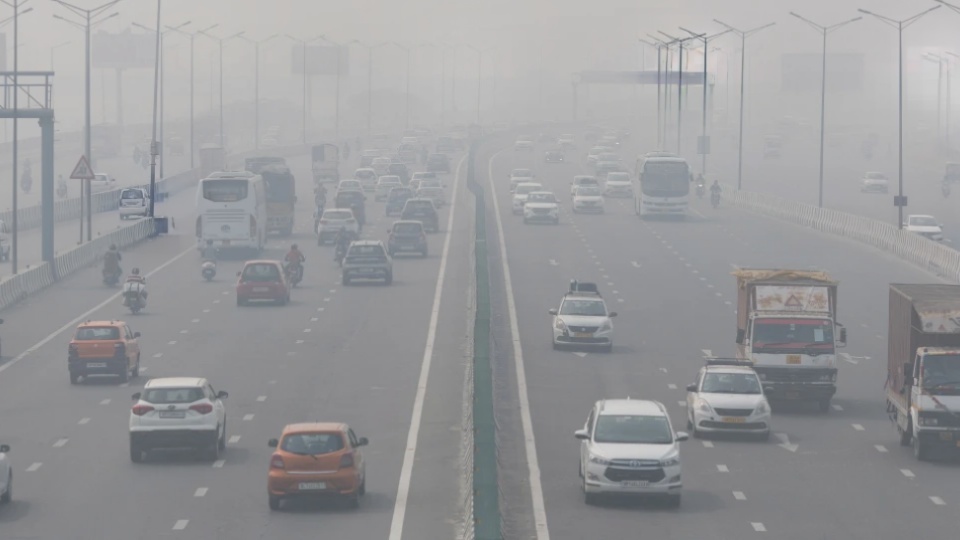}
    \end{subfigure}
    \begin{subfigure}{0.137\textwidth}
        \includegraphics[width=\hsize]{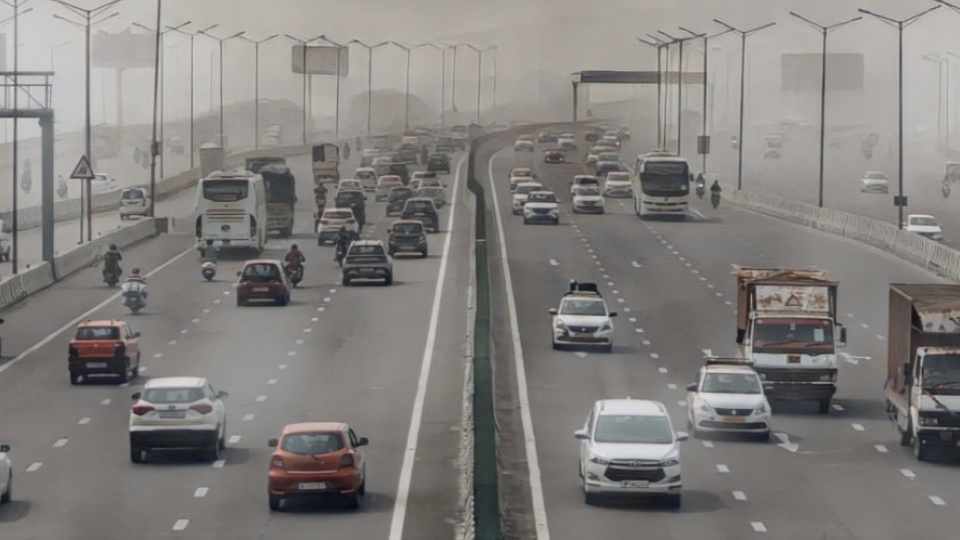}
    \end{subfigure}
    \begin{subfigure}{0.137\textwidth}
        \includegraphics[width=\hsize]{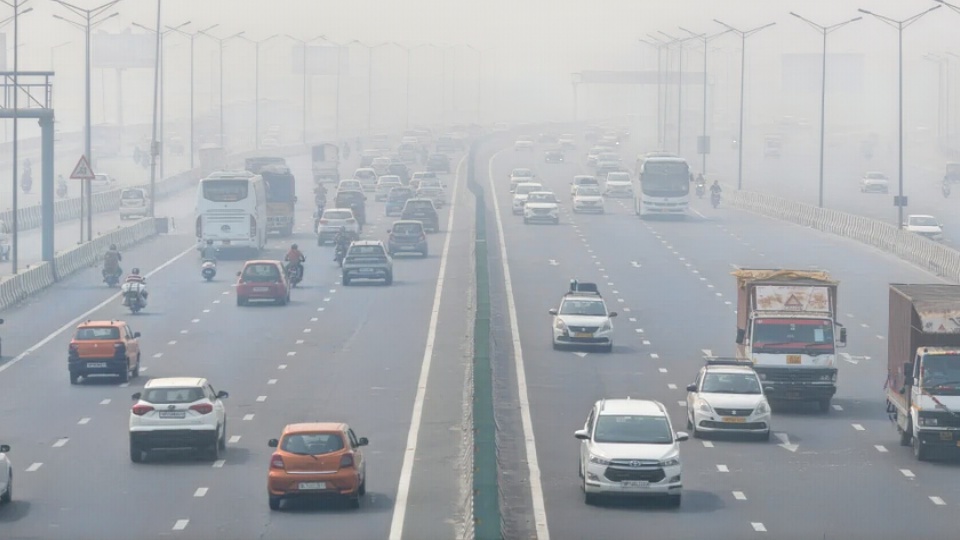}
    \end{subfigure}
    \begin{subfigure}{0.137\textwidth}
        \includegraphics[width=\hsize]{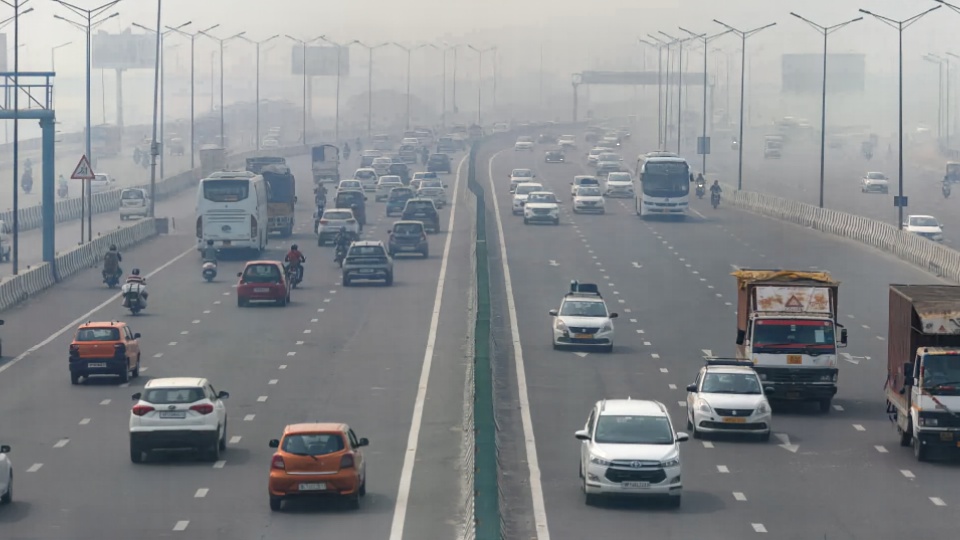}
    \end{subfigure}
    \\ \vspace{2pt}
    \begin{subfigure}{0.137\textwidth}
        \includegraphics[width=\textwidth]{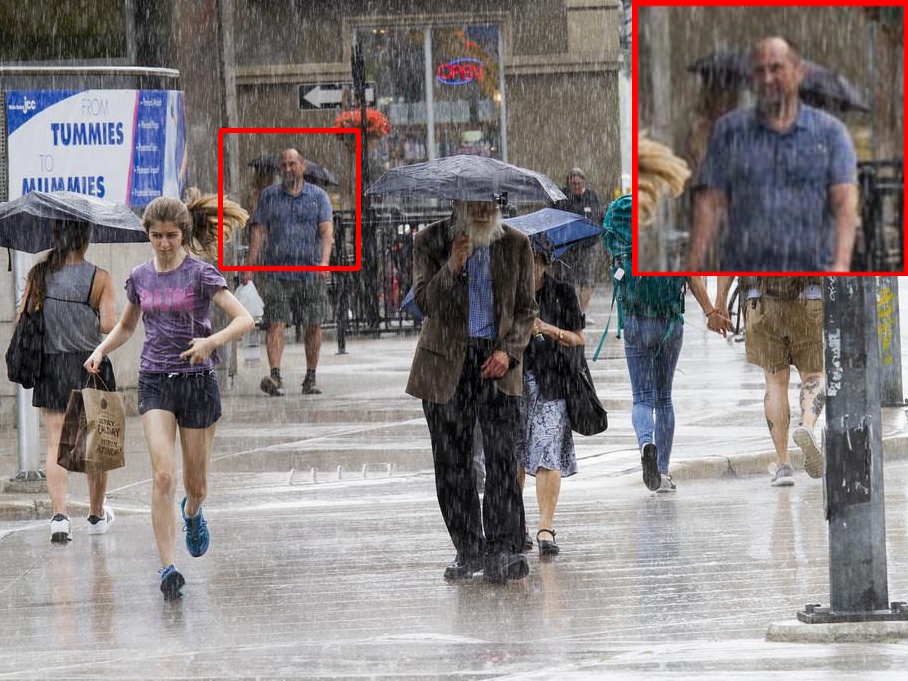}
    \end{subfigure}
    \begin{subfigure}{0.137\textwidth}
        \includegraphics[width=\textwidth]{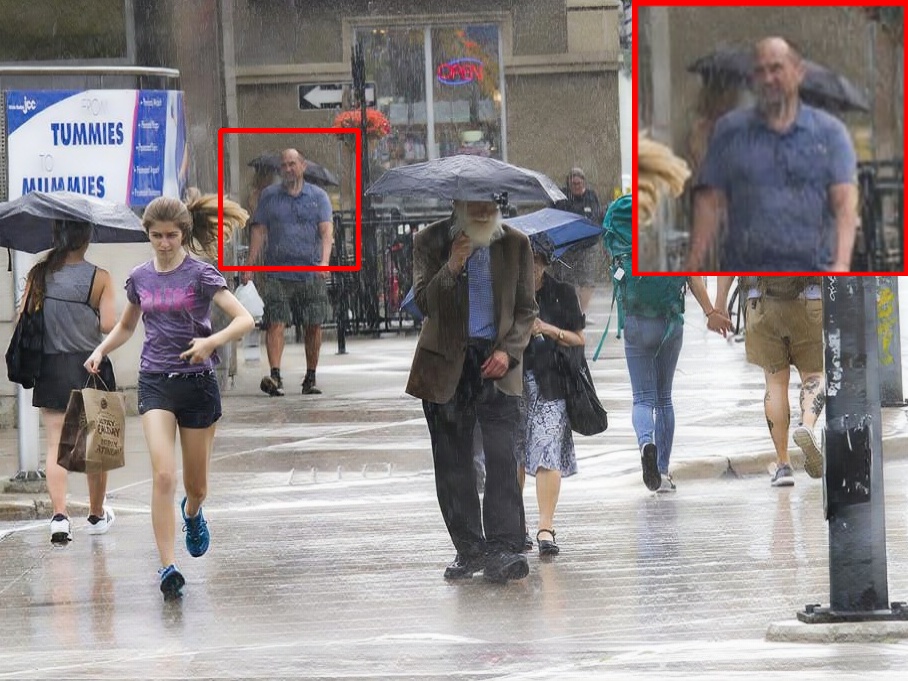}
    \end{subfigure}
    \begin{subfigure}{0.137\textwidth}
        \includegraphics[width=\textwidth]{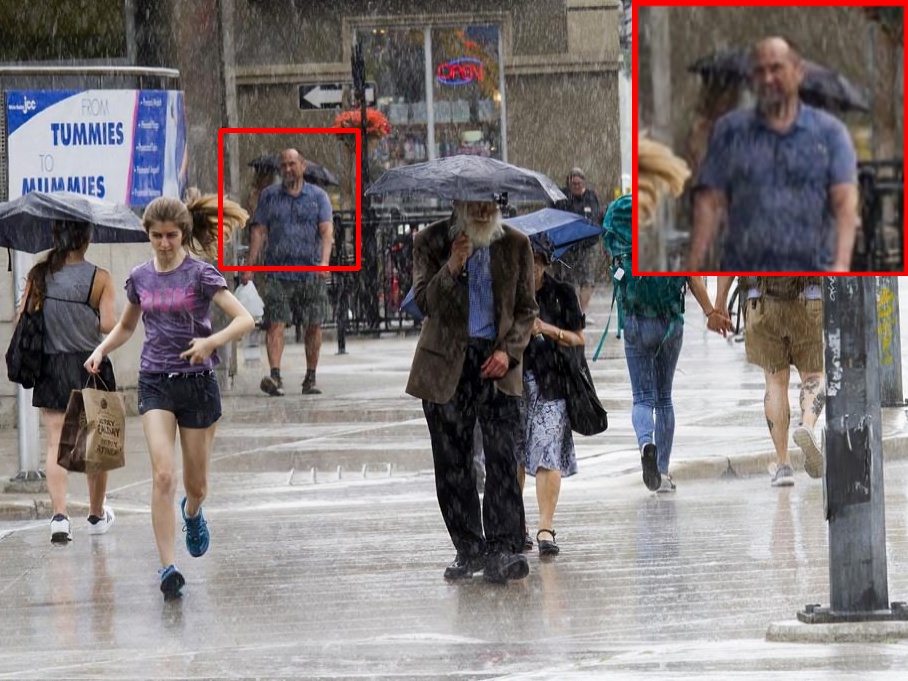}
    \end{subfigure}
    \begin{subfigure}{0.137\textwidth}
        \includegraphics[width=\textwidth]{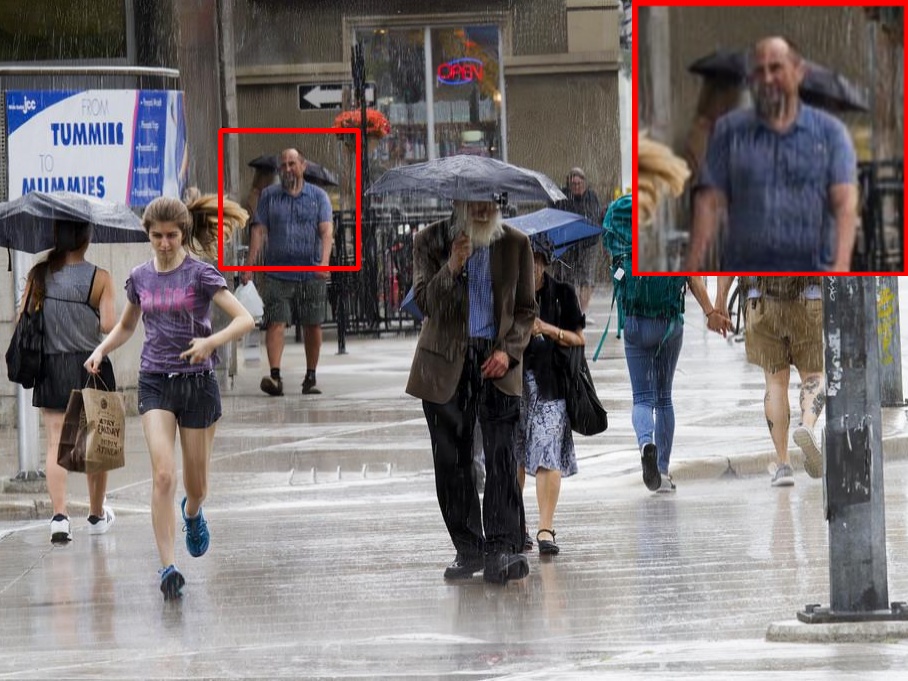}
    \end{subfigure}
    \begin{subfigure}{0.137\textwidth}
        \includegraphics[width=\textwidth]{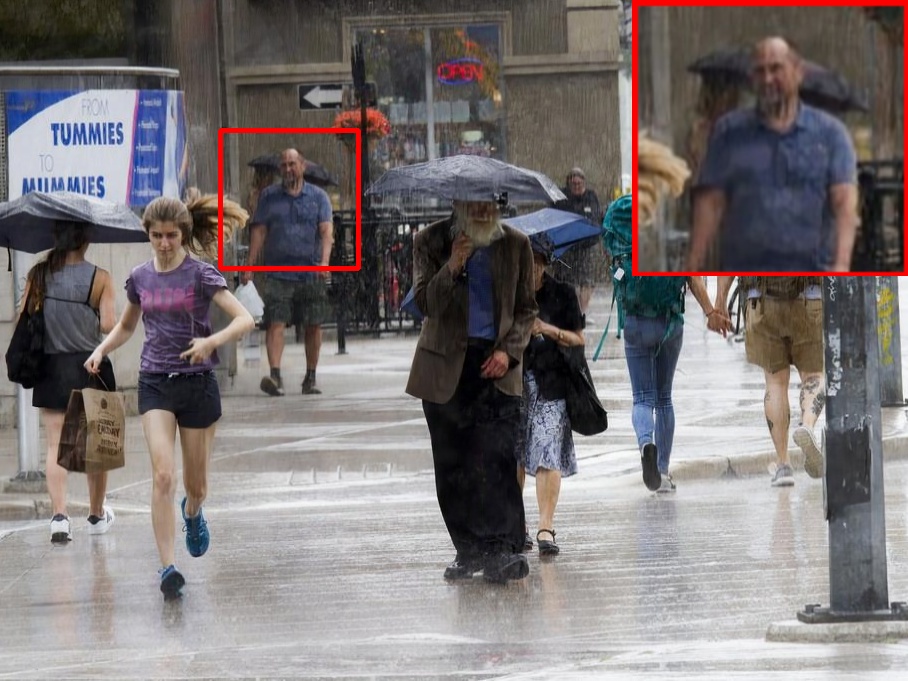}
    \end{subfigure}
    \begin{subfigure}{0.137\textwidth}
        \includegraphics[width=\textwidth]{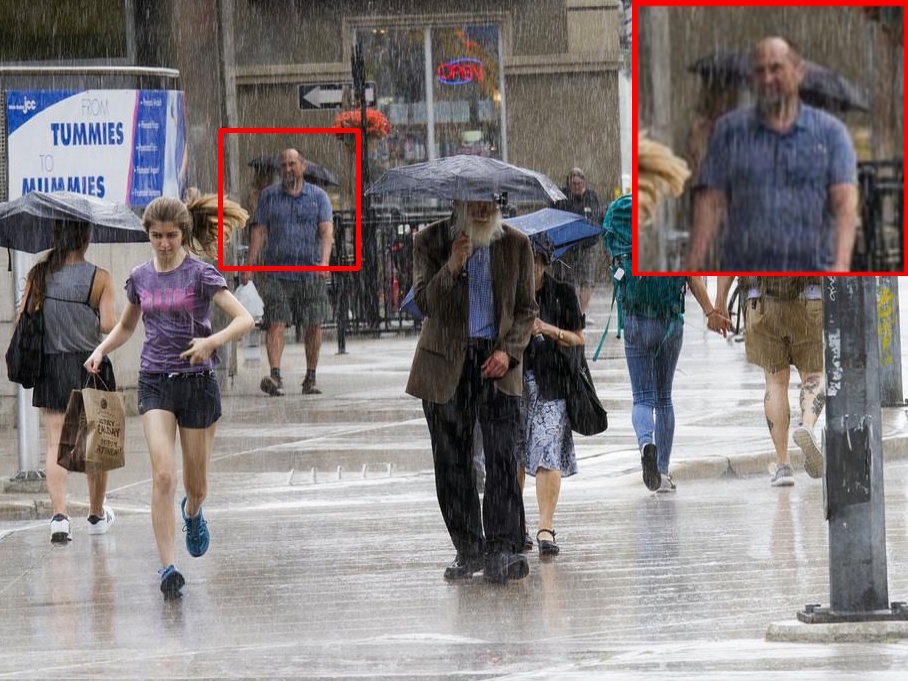}
    \end{subfigure}
    \begin{subfigure}{0.137\textwidth}
        \includegraphics[width=\textwidth]{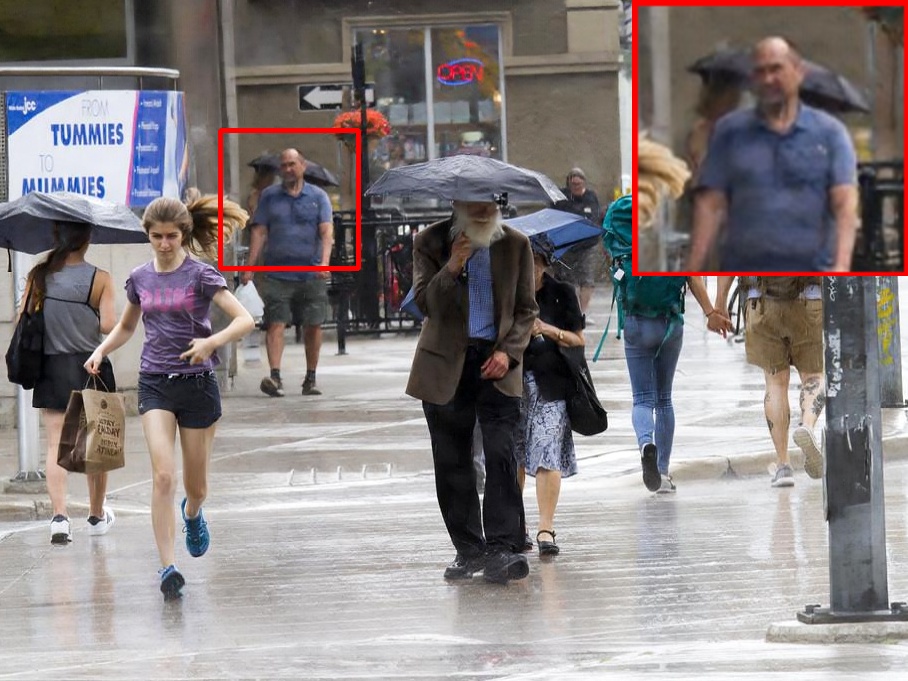}
    \end{subfigure}
    \\ \vspace{2pt}
    \begin{subfigure}{0.137\textwidth}
        \includegraphics[width=\textwidth]{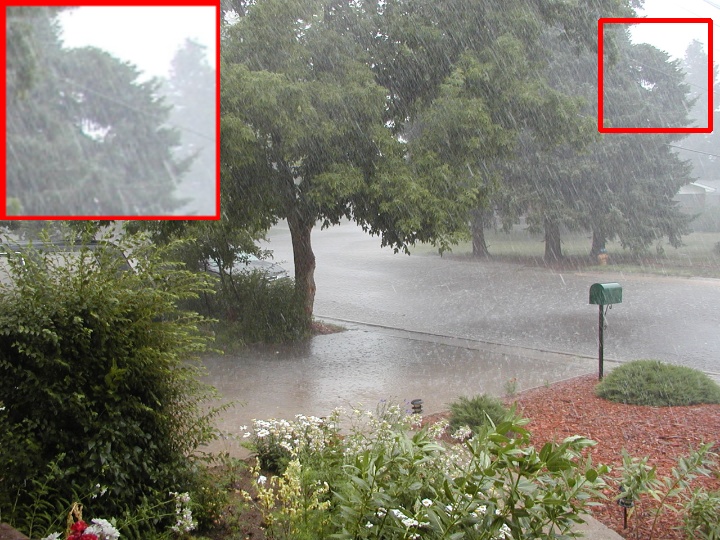}
    \end{subfigure}
    \begin{subfigure}{0.137\textwidth}
        \includegraphics[width=\textwidth]{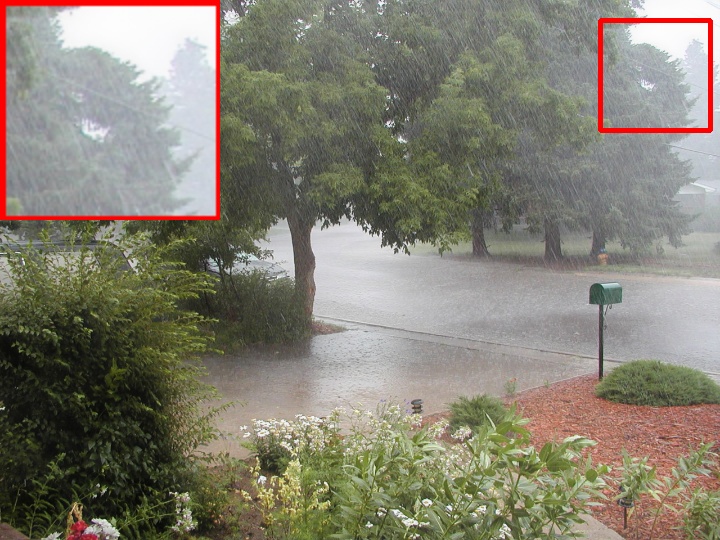}
    \end{subfigure}
    \begin{subfigure}{0.137\textwidth}
        \includegraphics[width=\textwidth]{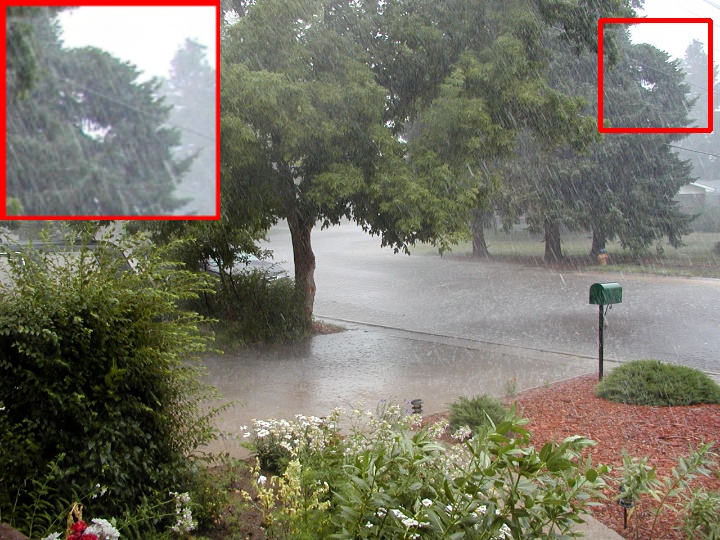}
    \end{subfigure}
    \begin{subfigure}{0.137\textwidth}
        \includegraphics[width=\textwidth]{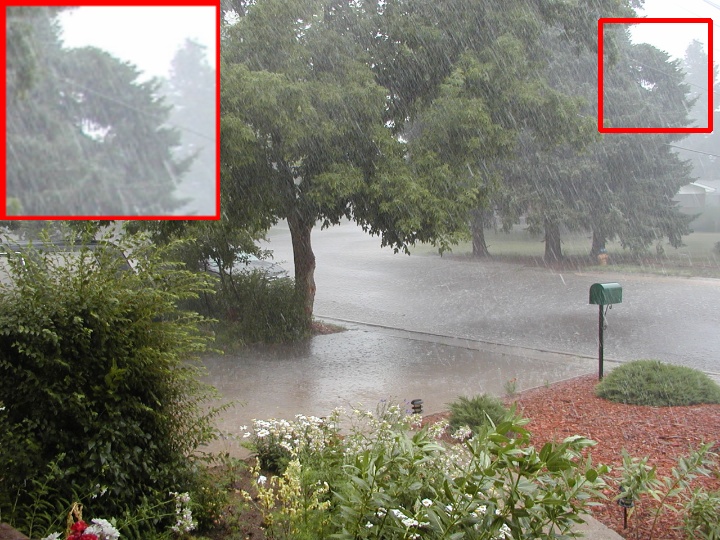}
    \end{subfigure}
    \begin{subfigure}{0.137\textwidth}
        \includegraphics[width=\textwidth]{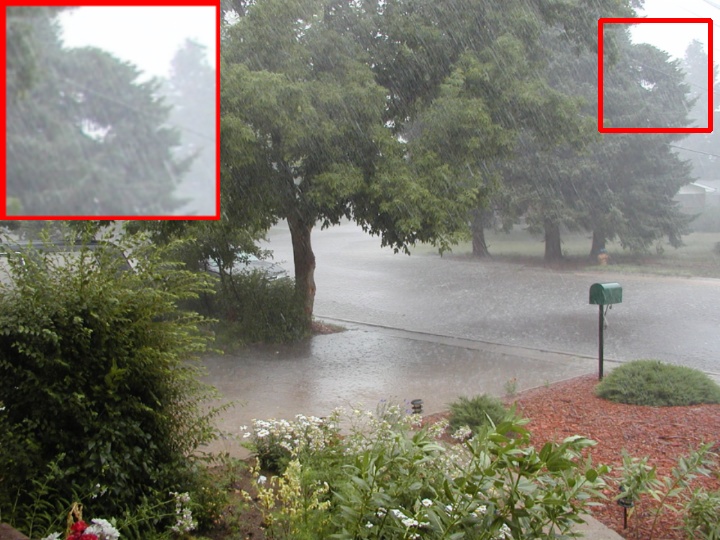}
    \end{subfigure}
    \begin{subfigure}{0.137\textwidth}
        \includegraphics[width=\textwidth]{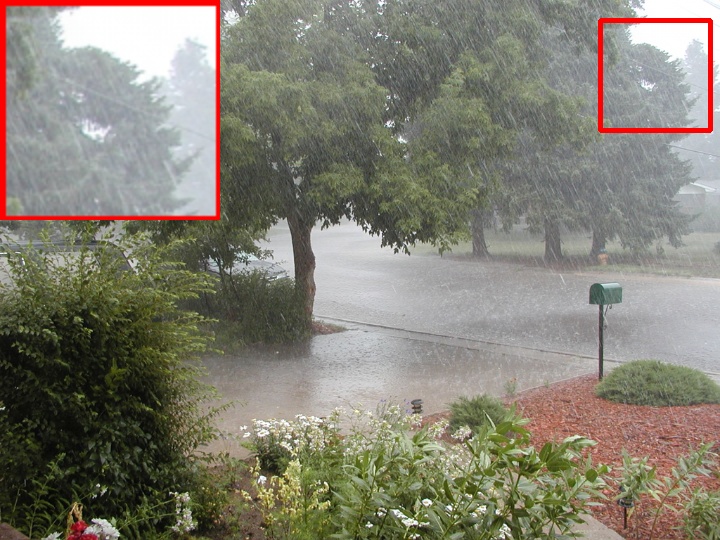}
    \end{subfigure}
    \begin{subfigure}{0.137\textwidth}
        \includegraphics[width=\textwidth]{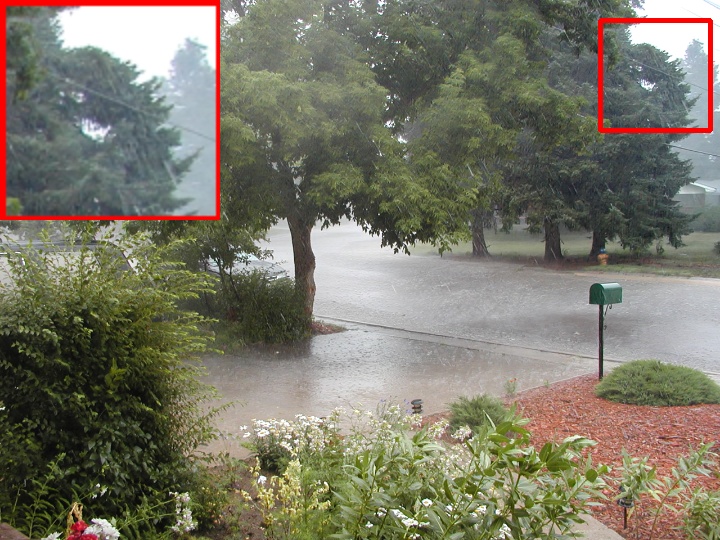}
    \end{subfigure}
    \\ \vspace{2pt}
    \begin{subfigure}{0.137\textwidth}
        \includegraphics[width=\textwidth]{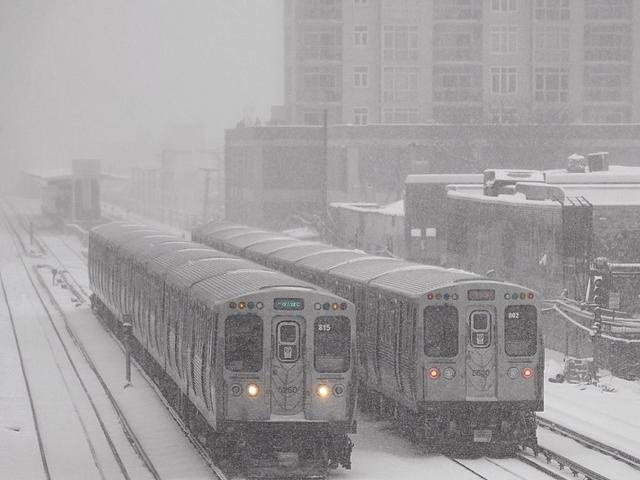}
    \end{subfigure}
    \begin{subfigure}{0.137\textwidth}
        \includegraphics[width=\textwidth]{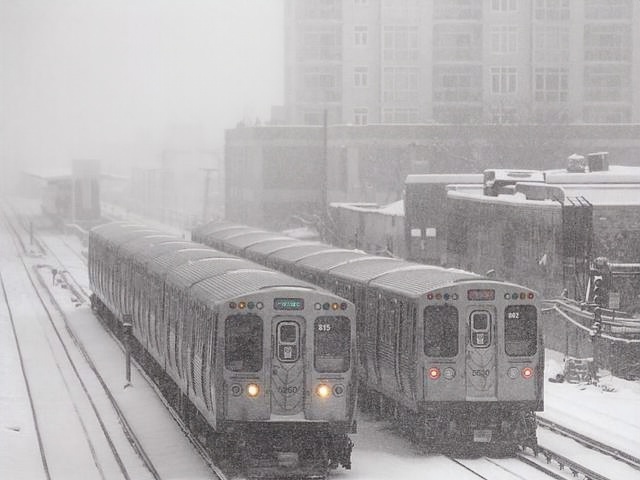}
    \end{subfigure}
    \begin{subfigure}{0.137\textwidth}
        \includegraphics[width=\textwidth]{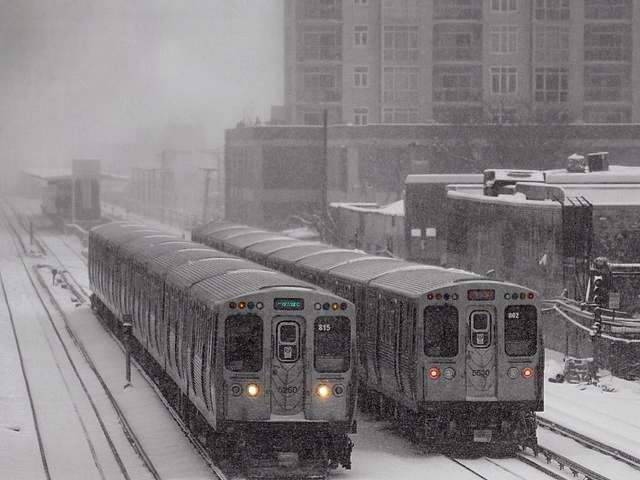}
    \end{subfigure}
    \begin{subfigure}{0.137\textwidth}
        \includegraphics[width=\textwidth]{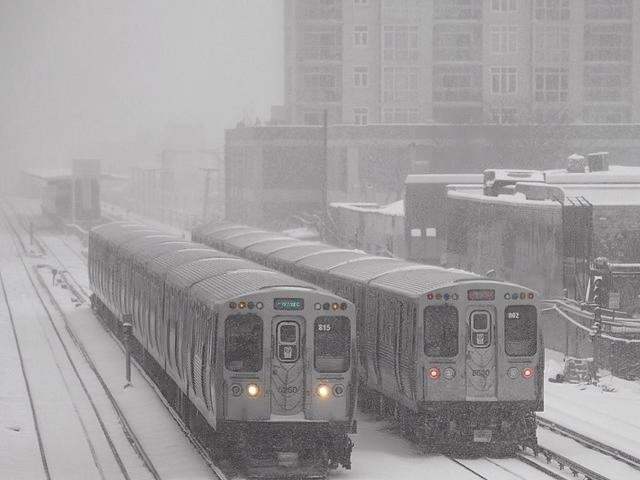}
    \end{subfigure}
    \begin{subfigure}{0.137\textwidth}
        \includegraphics[width=\textwidth]{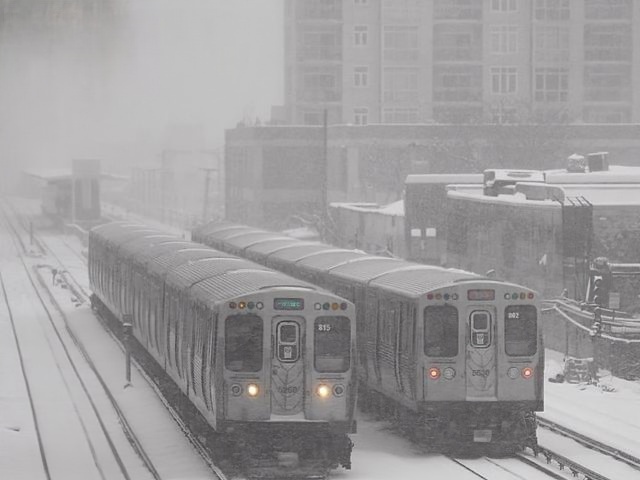}
    \end{subfigure}
    \begin{subfigure}{0.137\textwidth}
        \includegraphics[width=\textwidth]{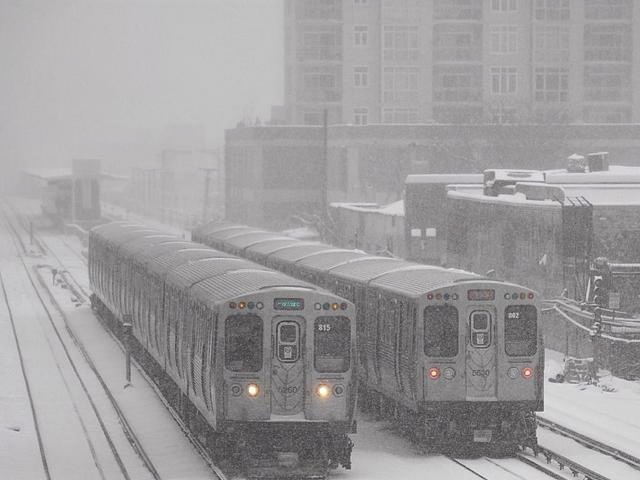}
    \end{subfigure}
    \begin{subfigure}{0.137\textwidth}
        \includegraphics[width=\textwidth]{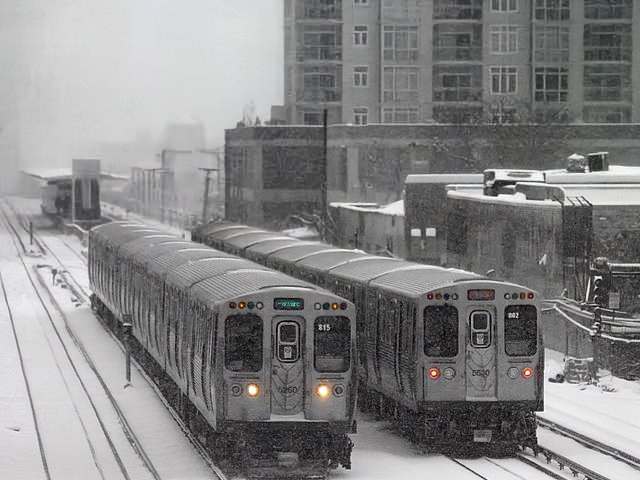}
    \end{subfigure}
    \\ \vspace{2pt}
    \begin{subfigure}{0.137\textwidth}
        \includegraphics[width=\textwidth]{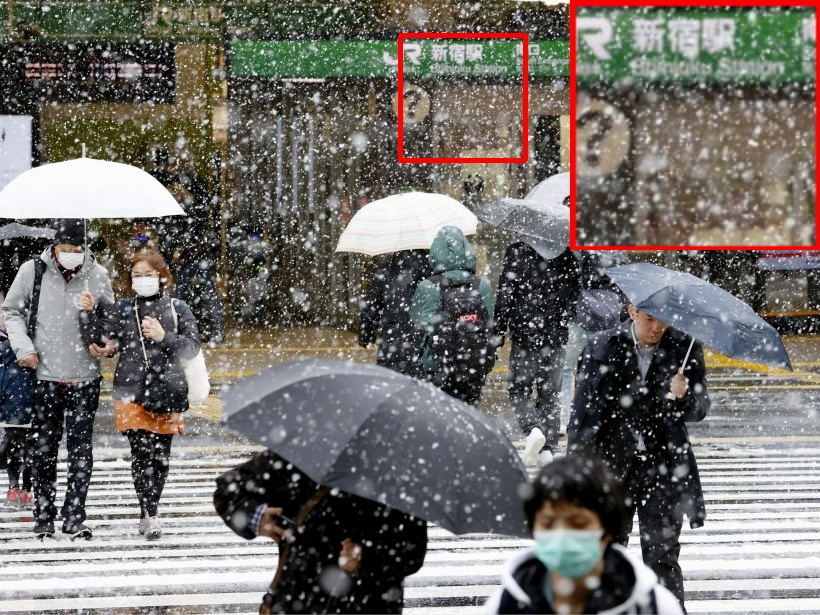}
        \caption{Input}
    \end{subfigure}
    \begin{subfigure}{0.137\textwidth}
        \includegraphics[width=\textwidth]{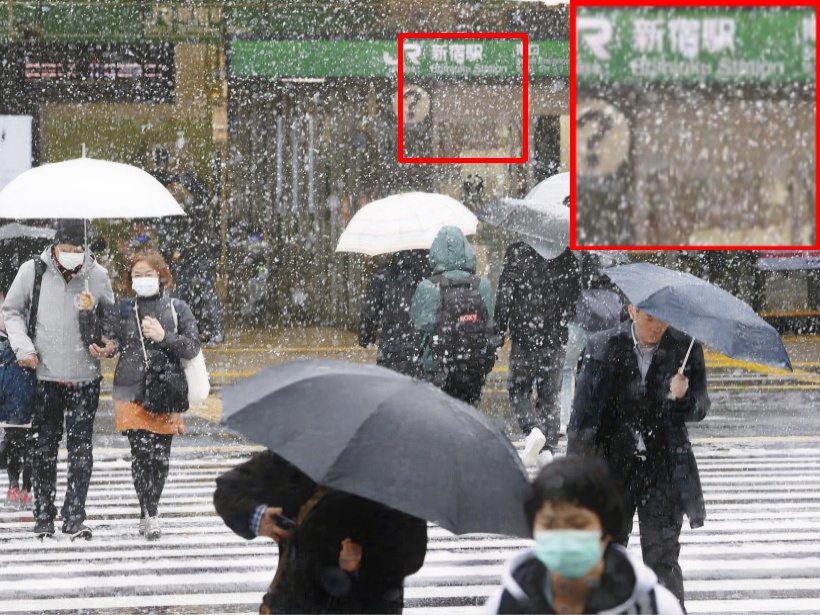}
        \caption{TransWeather~\cite{valanarasu2022transweather}}
    \end{subfigure}
    \begin{subfigure}{0.137\textwidth}
        \includegraphics[width=\textwidth]{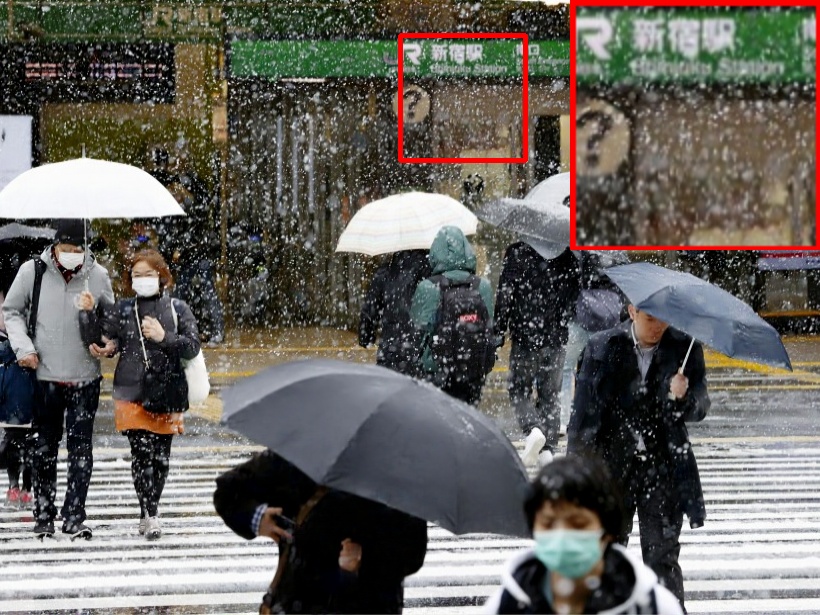}
        \caption{Two-Stage~\cite{chen2022learning}}
    \end{subfigure}
    \begin{subfigure}{0.137\textwidth}
        \includegraphics[width=\textwidth]{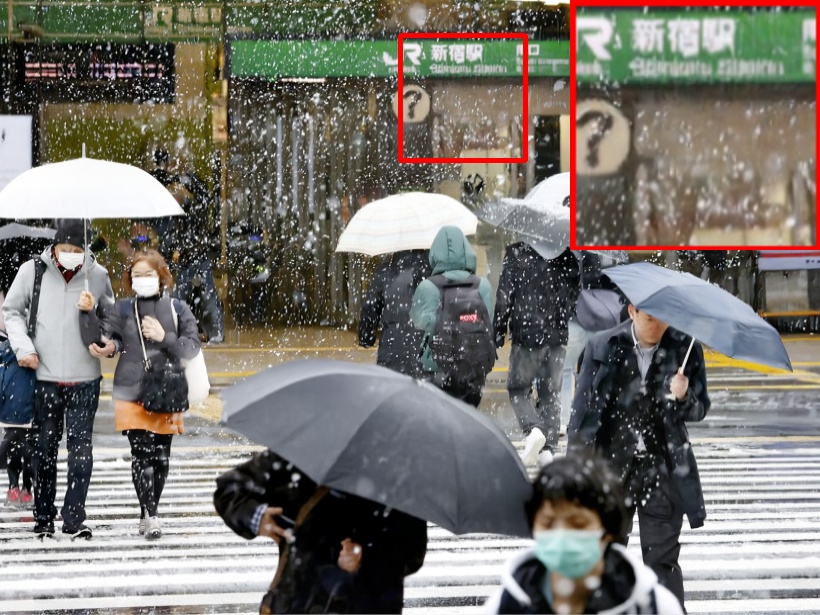}
        \caption{WeatherDiff~\cite{ozdenizci2023restoring}}
    \end{subfigure}
    \begin{subfigure}{0.137\textwidth}
        \includegraphics[width=\textwidth]{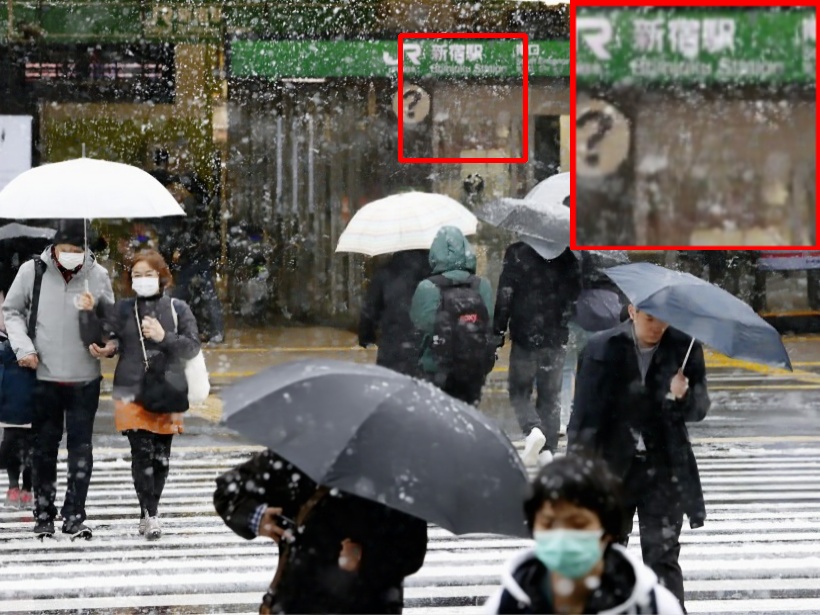}
        \caption{WGWS-Net~\cite{zhu2023learning}}
    \end{subfigure}
    \begin{subfigure}{0.137\textwidth}
        \includegraphics[width=\textwidth]{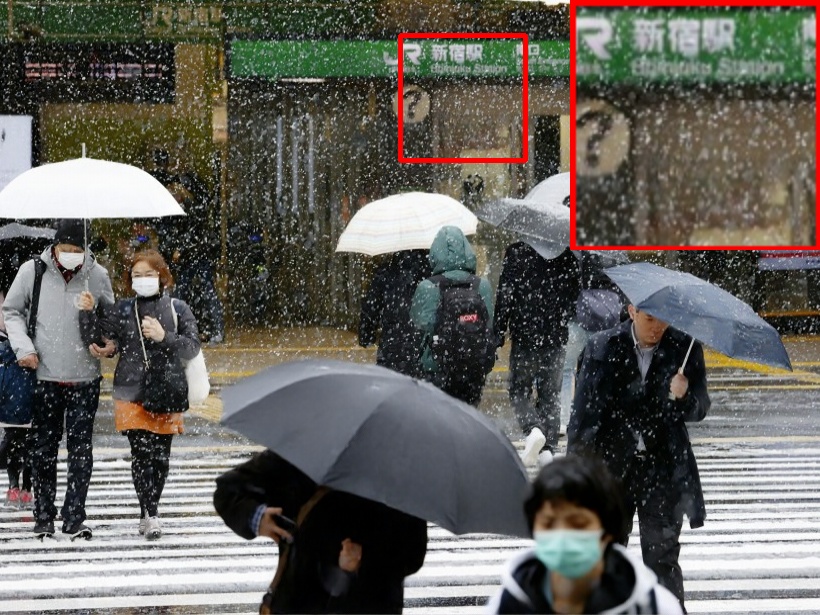}
        \caption{MWFormer~\cite{zhu2024mwformer}}
    \end{subfigure}
    \begin{subfigure}{0.137\textwidth}
        \includegraphics[width=\textwidth]{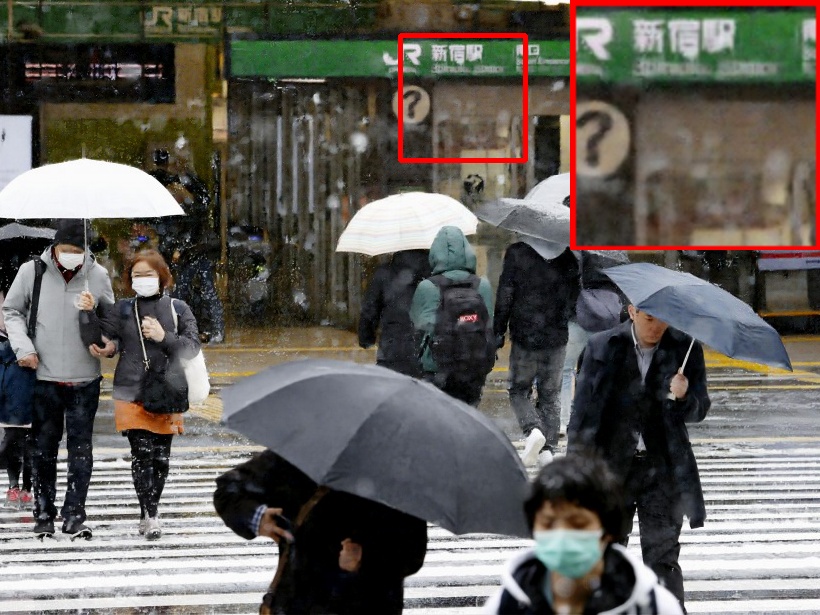}
        \caption{Our method}
    \end{subfigure}
    \\
    \caption{\textbf{Visual comparisons of real-world photos under adverse weather conditions.}
    Our method clearly removes haze, rain, and snow artifacts and generates more visually appealing results.
    Please zoom in for a better view.
    }
    \label{fig:vis_real}
\end{figure*}

\paragraph{Evaluation Setup}
(i) the first experimental setting: we compare our method with the state-of-the-art image restoration methods~\cite{valanarasu2022transweather,chen2022learning,ozdenizci2023restoring,zhu2023learning,zhu2024mwformer} for multiple weather conditions (multi-task).
Besides, advanced image restoration methods~\cite{zamir2022restormer,chen2022simple}, trained with mixed data (mixed training), are also compared.
We re-train these models based on their public implementations.
(ii) the second experimental setting: we compare with the task-specific and multi-task image restoration methods, following the evaluation in \cite{valanarasu2022transweather,ozdenizci2023restoring}.
(iii) the third experimental setting: we compare our method with the task-specific image restoration methods using mixed training data and multi-task methods, as introduced in \cite{chen2022learning}.

We adopt two widely-used metrics (\textit{i.e.}, PSNR and SSIM) to quantitatively measure image restoration performance in the main study, following prior works \cite{valanarasu2022transweather,chen2022learning,zamir2022restormer,zhu2023learning}.
For the first and third settings, we follow common practices from the original implementations \cite{li2018benchmarking,li2019heavy,liu2018desnownet}, calculating PSNR and SSIM in the RGB space for dehazing and desnowing, and on the Y channel for deraining.
For the second setting, we calculate PSNR and SSIM on the Y channel, following TransWeather \cite{valanarasu2022transweather} and WeatherDiff \cite{ozdenizci2023restoring}.

\paragraph{Quantitative Comparisons}
Table~\ref{tab:setting1}, Table~\ref{tab:setting2}, and Table~\ref{tab:setting3} report the PSNR and SSIM scores of our network and the compared methods for three experimental settings, respectively.
From these quantitative results, we find that our method outperforms the state-of-the-art weather-related restoration methods in terms of almost all the metrics.
Specifically, regarding the first experimental setting, WSWG-Net~\cite{zhu2023learning} has high PSNR (26.09~dB), while Two-Stage~\cite{chen2022learning} achieves the high mean SSIM (0.8966), among all compared methods.
More importantly, compared to WSWG-Net and Two-stage, our network further improves the PNSR and SSIM scores on the haze, rain, and snow subsets, with PSNR improvements of 1.89~dB, 1.24~dB, and 1.10~dB over WSWG-Net, respectively, and has an SSIM gain of 0.012 over Two-Stage on average.

For the second experimental setting, our network achieves the best PSNR and SSIM for the raindrop and snow subsets, and the best PSNR for the rain subset.
Our SSIM score (0.9309) on the rain subset is slightly smaller than the best one (0.9312).
Our method outperforms the state-of-the-art method MWFormer~\cite{zhu2024mwformer} by 1.15~dB in PSNR on average; see Table~\ref{tab:setting2}.
In terms of the third experimental setting, our method consistently achieves the highest PSNR and SSIM across all three subsets (namely haze, rain, and snow), and has a PSNR gain of 2.44~dB on average over Two-Stage, which stands out as the best among the compared methods; see Table~\ref{tab:setting3}.

\paragraph{Qualitative Comparisons}
The qualitative comparisons on the synthetic datasets are shown in Fig.~\ref{fig:vis_synthetic}.
From the visual observations, it is evident that the compared methods either retain some noticeable artifacts (such as residual haze in the background and remaining rain and snow effects) or compromise the structural details.
In comparison, our method effectively removes the haze, rain streaks, and snowflakes, and recovers the underlying scenes.
It is clear that restored images generated by our method have more natural and appealing appearances, aligning closely with the ground truth images.

\begin{table}[tp]
\centering
\caption{No-reference evaluation results on the real-world RTTS~\cite{li2018benchmarking}, DDN-SIRR~\cite{wei2019semi}, and Snow100K~\cite{liu2018desnownet} datasets.}
\label{tab:no-reference}
\begin{tabular}{c|ccc}
\toprule
\multirow{2}{*}{Method}                        & \multicolumn{3}{c}{NIQE $\downarrow$ / MUSIQ $\uparrow$ }                               \\ \cline{2-4} 
                                               & Haze                         & Rain                      & Snow                         \\
\midrule
TransWeather~\cite{valanarasu2022transweather} & 6.29/47.17                   & {\ul 3.88}/55.24          & 3.56/54.46                   \\
Two-Stage~\cite{chen2022learning}              & {\ul 5.20}/48.22             & 4.03/52.79                & 3.22/59.25                   \\
WeatherDiff~\cite{ozdenizci2023restoring}      & 5.59/47.68                   & 4.21/53.51                & 3.29/60.52                   \\
WGWS-Net~\cite{zhu2023learning}                & 5.29/45.77                   & 4.05/51.97                & 3.54/57.96                   \\
MWFormer~\cite{zhu2024mwformer}                & 5.38/{\ul 54.48}             & 4.14/\textbf{57.28}       & {\ul 2.98}/{\ul 61.30}       \\
Our Method                                     & \textbf{4.96}/\textbf{55.19} & \textbf{3.79}/{\ul 55.51} & \textbf{2.91}/\textbf{61.93} \\ \bottomrule
\end{tabular}
\end{table}

\subsection{Image Restoration for Real-World Images}
To assess the generalization capabilities, we evaluate the performance of the trained models on real-world images.
For quantitative comparisons, we employ widely used no-reference image quality assessment metrics, namely NIQE~\cite{mittal2012making} and MUSIQ~\cite{ke2021musiq}, since there are no ground truth images for the real images captured in bad weather.
The realistic haze, rain, and snow images from RESIDE RTTS~\cite{li2018benchmarking}, DDN-SIRR~\cite{wei2019semi}, and Snow100K~\cite{liu2018desnownet} are selected for the evaluation.
As shown in Table~\ref{tab:no-reference}, our method consistently surpasses the performance of other compared methods across most metrics.

Furthermore, we present visual examples of real-world image samples sourced from these datasets and the internet.
These images under adverse weather conditions are paired with their corresponding restoration results, as shown in Fig.~\ref{fig:vis_real}.
Our method demonstrates superior performance in removing haze, rain, and snow-related artifacts, enhancing visibility, and producing visually appealing results compared to other methods.
This enhanced capability is especially evident in images affected by rain and snow, where visible occlusions are present and scattering effects are noticeable.

\subsection{Ablation Studies}

We first analyze the proposed imaging model, particularly the combination of near and far regions with multiple layers in $\alpha$ to synthesize rain and snow, while accounting for their volumetric effects.
Then, we evaluate the effectiveness of our network and its key components through ablation study experiments on our dataset.
We train models for 120K iterations with batch size 16 and report the average PSNR and SSIM results over three weather conditions in Table~\ref{tab:ablation1}.

\begin{figure}[t]
    \centering
    \captionsetup[subfigure]{font=small,labelformat=empty,justification=centering}
    \begin{subfigure}{0.24\hsize}
        \includegraphics[width=\hsize]{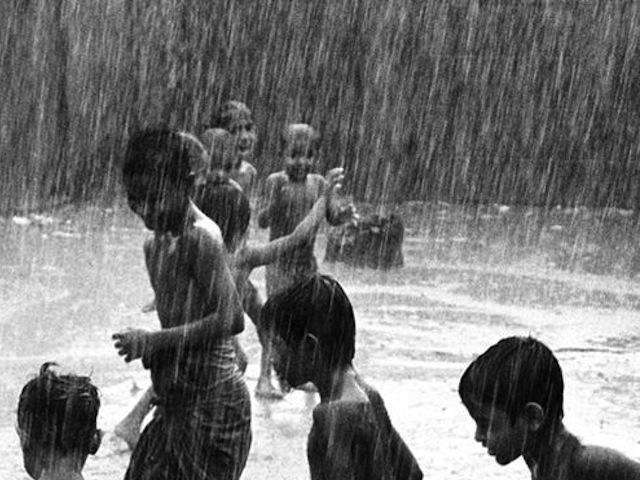}
    \end{subfigure}
    \begin{subfigure}{0.24\hsize}
        \includegraphics[width=\hsize]{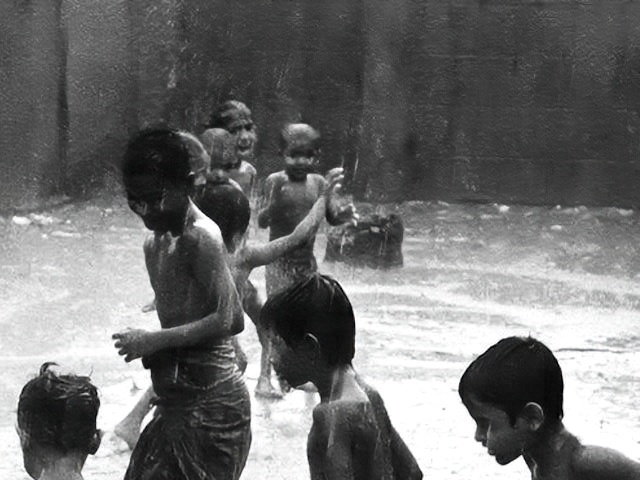}
    \end{subfigure}
    \begin{subfigure}{0.24\hsize}
        \includegraphics[width=\hsize]{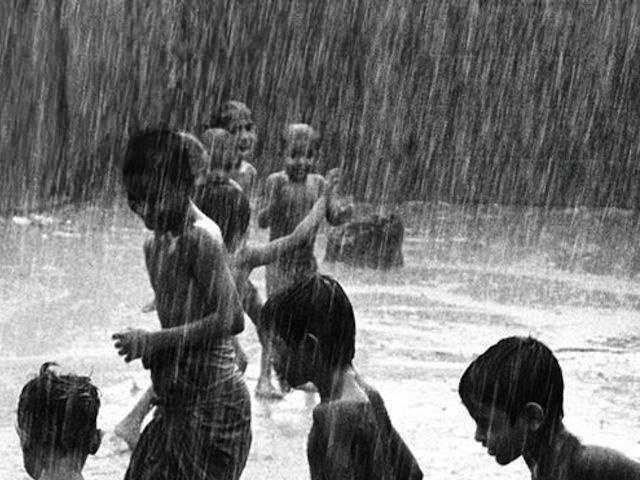}
    \end{subfigure}
    \begin{subfigure}{0.24\hsize}
        \includegraphics[width=\hsize]{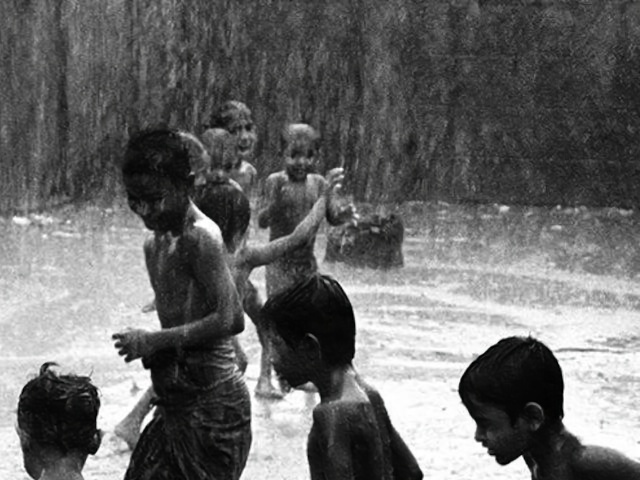}
    \end{subfigure}
    \\ \vspace{1pt}
    \begin{subfigure}{0.24\hsize}
        \includegraphics[width=\hsize]{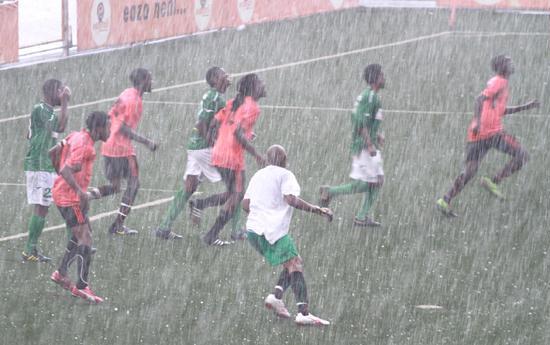}
    \end{subfigure}
    \begin{subfigure}{0.24\hsize}
        \includegraphics[width=\hsize]{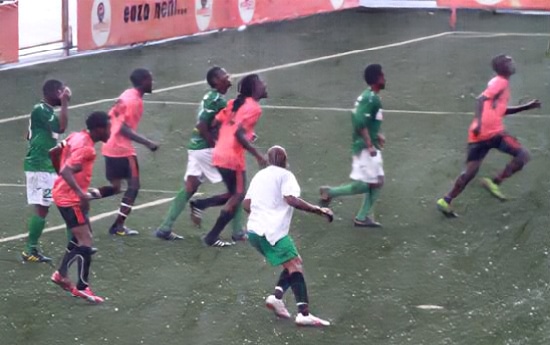}
    \end{subfigure}
    \begin{subfigure}{0.24\hsize}
        \includegraphics[width=\hsize]{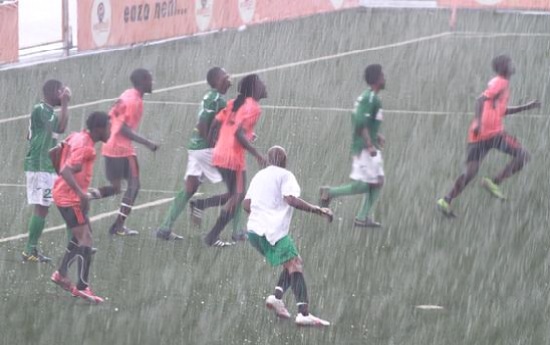}
    \end{subfigure}
    \begin{subfigure}{0.24\hsize}
        \includegraphics[width=\hsize]{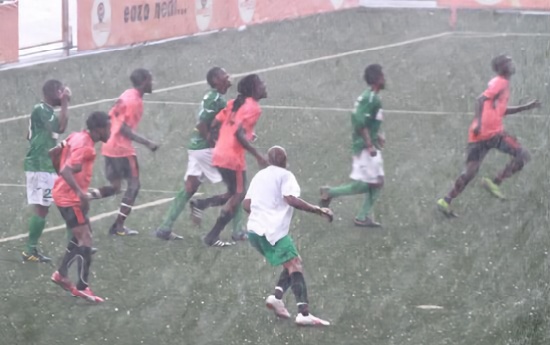}
    \end{subfigure}
    \\ \vspace{1pt}
    \begin{subfigure}{0.24\hsize}
        \includegraphics[width=\hsize]{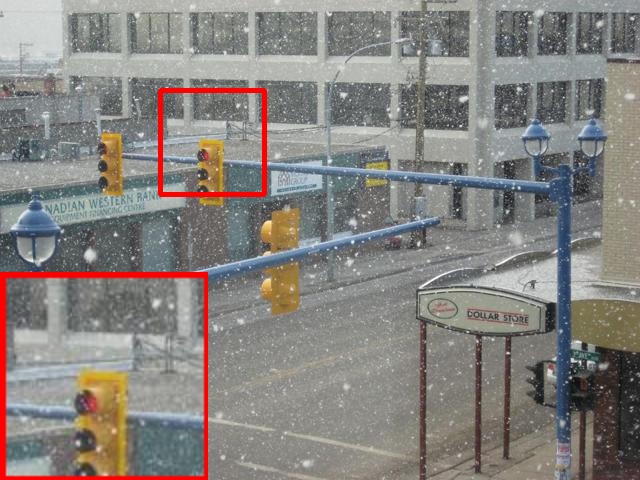}
        \caption{Input}
    \end{subfigure}
    \begin{subfigure}{0.24\hsize}
        \includegraphics[width=\hsize]{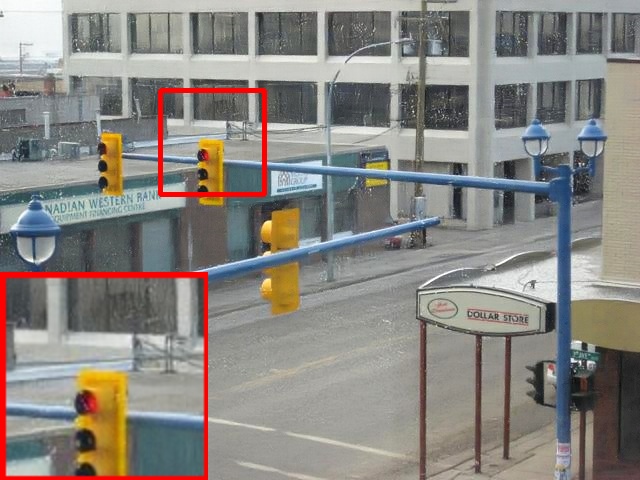}
        \caption{\textbf{Weather30K}}
    \end{subfigure}
    \begin{subfigure}{0.24\hsize}
        \includegraphics[width=\hsize]{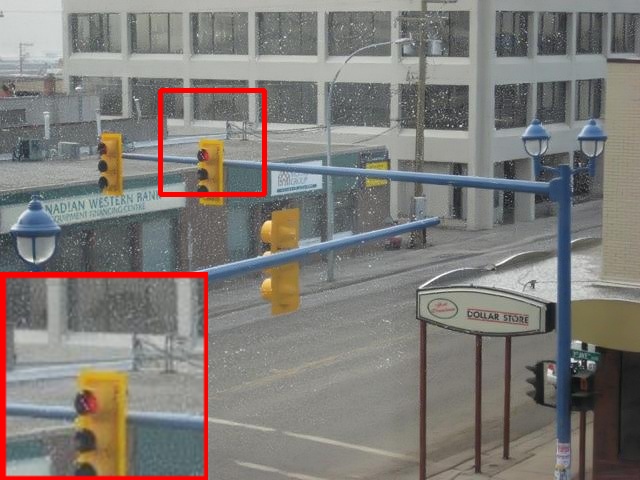}
        \caption{2nd Setting}
    \end{subfigure}
    \begin{subfigure}{0.24\hsize}
        \includegraphics[width=\hsize]{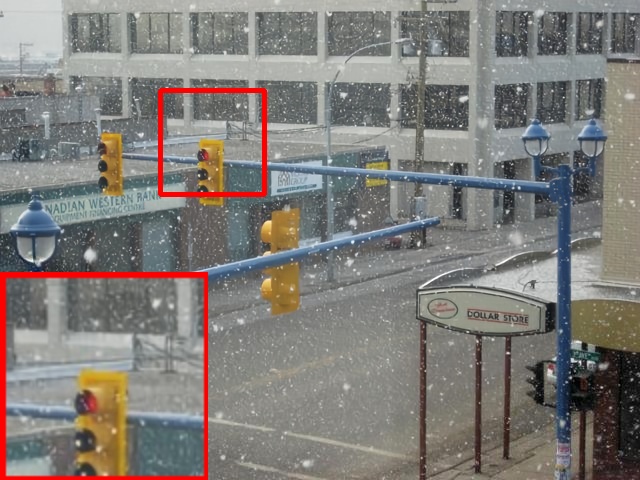}
        \caption{3rd Setting}
    \end{subfigure}
    \\
    \caption{\textbf{Ablation study results of the training data.}
    The model trained with Weather30K demonstrates a superior ability to remove foreground and background weather effects across light to heavy rain and snow conditions.
    }
    \label{fig:vis_ablation_training_data}
\end{figure}

\begin{table}[tp]
\centering
\caption{B-FEN \cite{wu2020subjective} results on the DDN-SIRR dataset for models trained with different data.
}
\label{tab:training_data}
\begin{tabular}{c|ccc}
\toprule
Training Data    & Weather30K     & Second Setting & Third Setting \\ \midrule
B-FEN $\uparrow$ & \textbf{0.341} & 0.293          & 0.311         \\ \bottomrule
\end{tabular}
\end{table}

\begin{figure}[t]
    \centering
    \captionsetup[subfigure]{font=small,labelformat=empty,justification=centering}
    \begin{subfigure}{0.32\hsize}
        \includegraphics[width=\hsize]{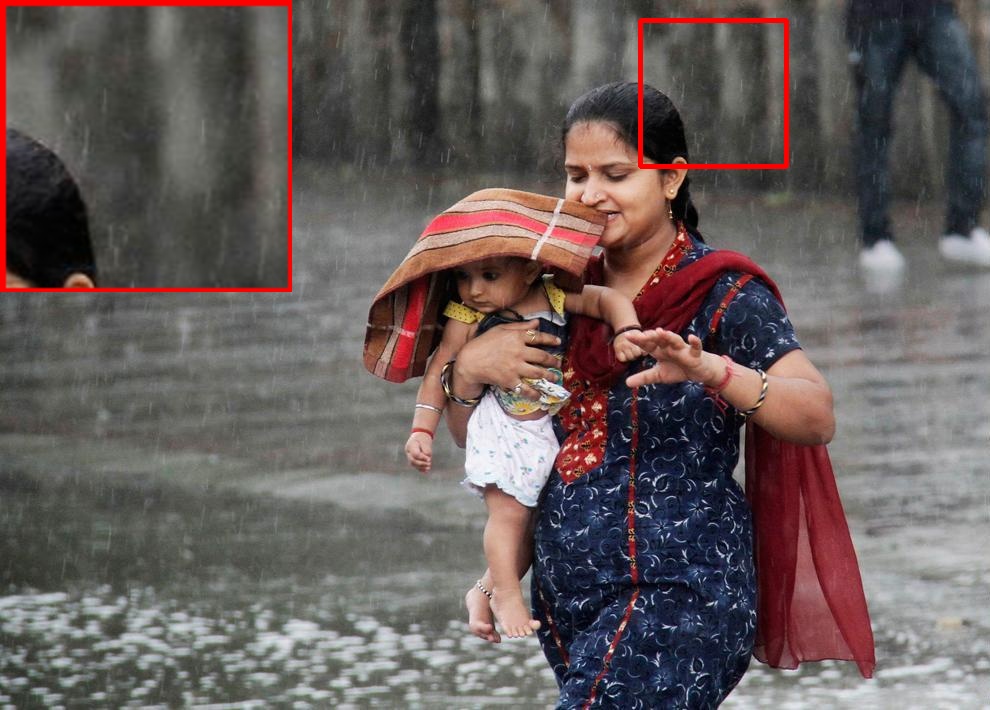}
    \end{subfigure}
    \begin{subfigure}{0.32\hsize}
        \includegraphics[width=\hsize]{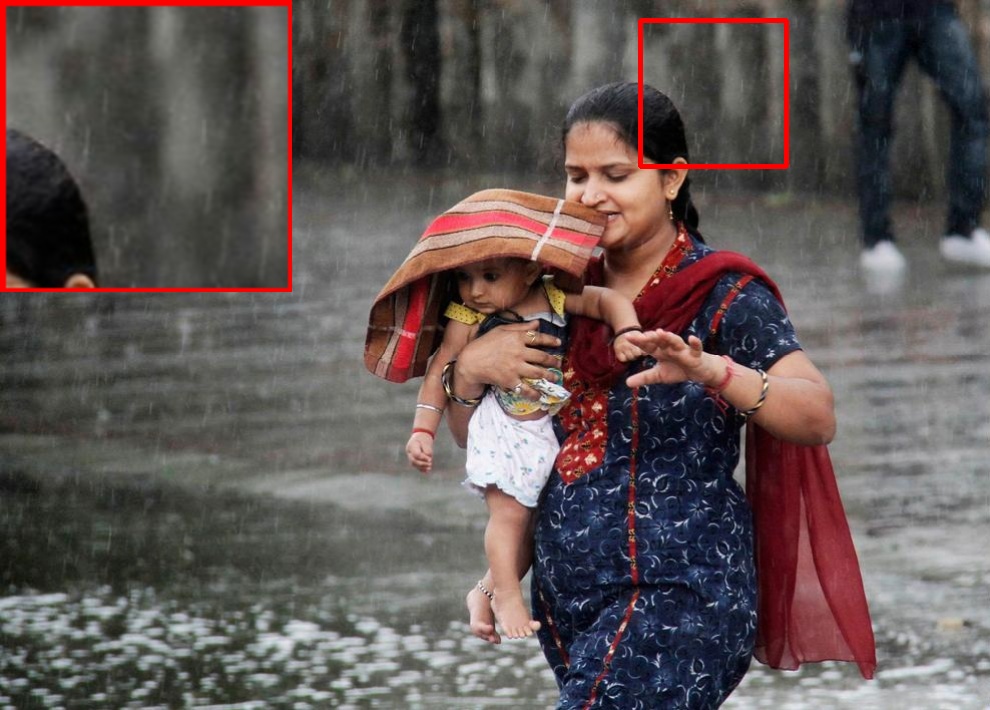}
    \end{subfigure}
    \begin{subfigure}{0.32\hsize}
        \includegraphics[width=\hsize]{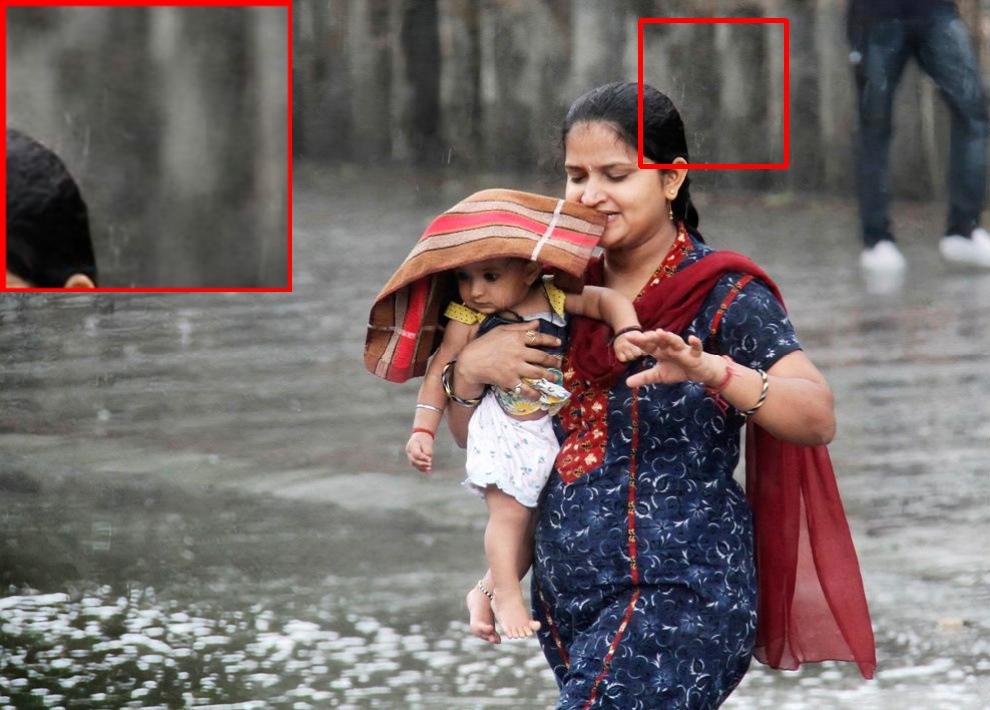}
    \end{subfigure}
    \\ \vspace{1pt}
    \begin{subfigure}{0.32\hsize}
        \includegraphics[width=\hsize]{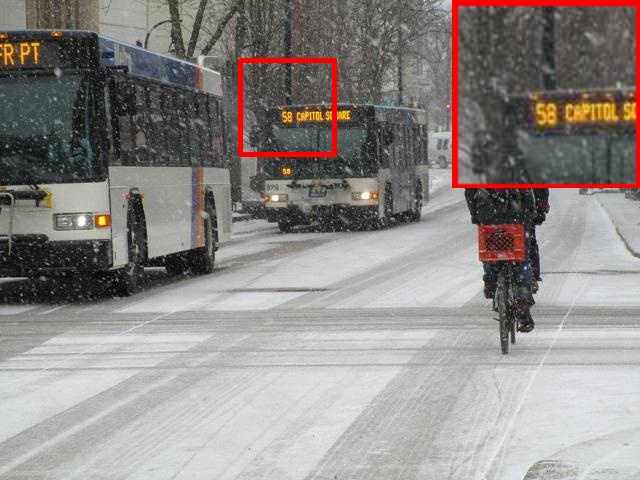}
        \caption{Input}
    \end{subfigure}
    \begin{subfigure}{0.32\hsize}
        \includegraphics[width=\hsize]{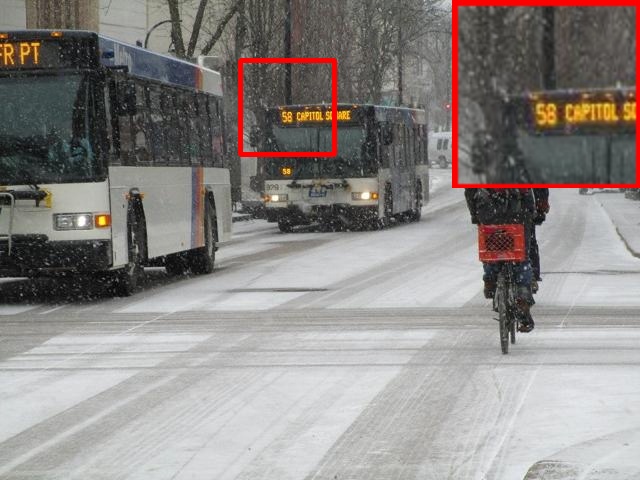}
        \caption{w/o multiple layers}
    \end{subfigure}
    \begin{subfigure}{0.32\hsize}
        \includegraphics[width=\hsize]{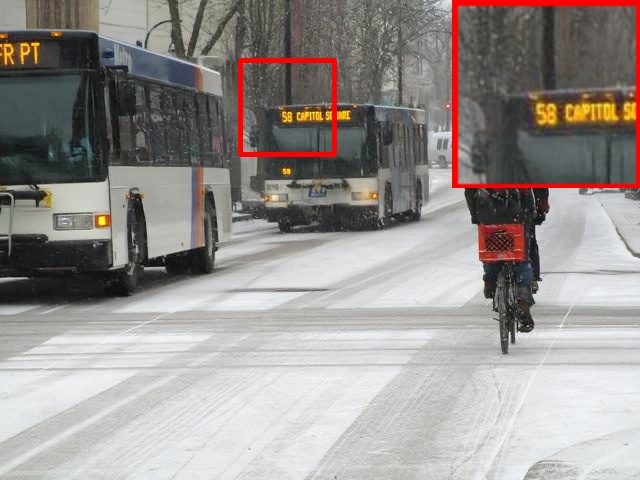}
        \caption{Our modeling}
    \end{subfigure}
    \\
    \caption{\textbf{Ablation study results of the imaging model.}
    The model trained with data synthesized by our imaging model, considering the rain and snow volumetric effects with multiple layers, removes more visible particles in real-world images.
    }
    \label{fig:vis_ablation_formula}
\end{figure}

\begin{figure}
    \centering
    \captionsetup[subfigure]{font=small,labelformat=empty,justification=centering}
    \begin{subfigure}{0.24\hsize}
        \includegraphics[width=\hsize]{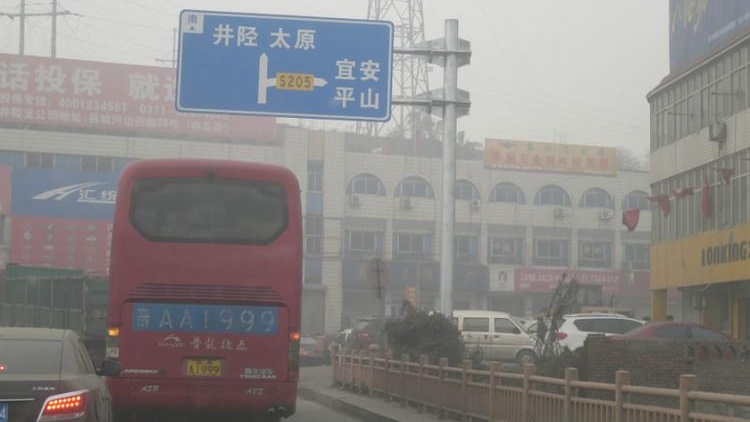}
    \end{subfigure}
    \begin{subfigure}{0.24\hsize}
        \includegraphics[width=\hsize]{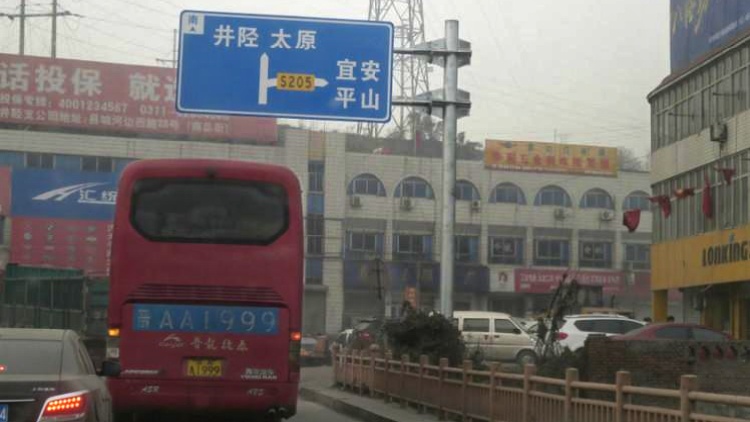}
    \end{subfigure}
    \begin{subfigure}{0.24\hsize}
        \includegraphics[width=\hsize]{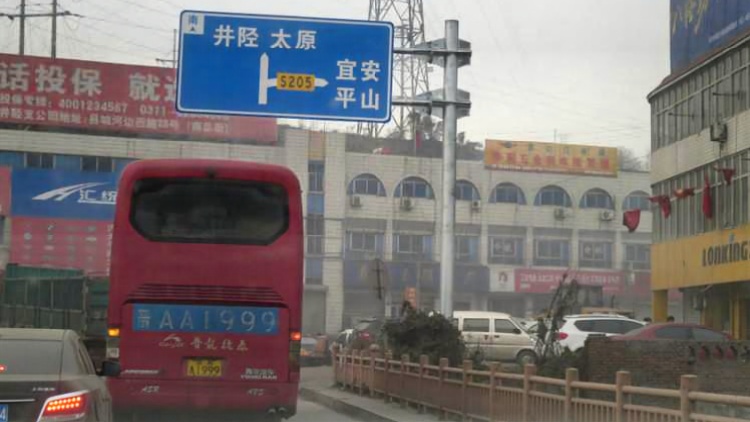}
    \end{subfigure}
    \begin{subfigure}{0.24\hsize}
        \includegraphics[width=\hsize]{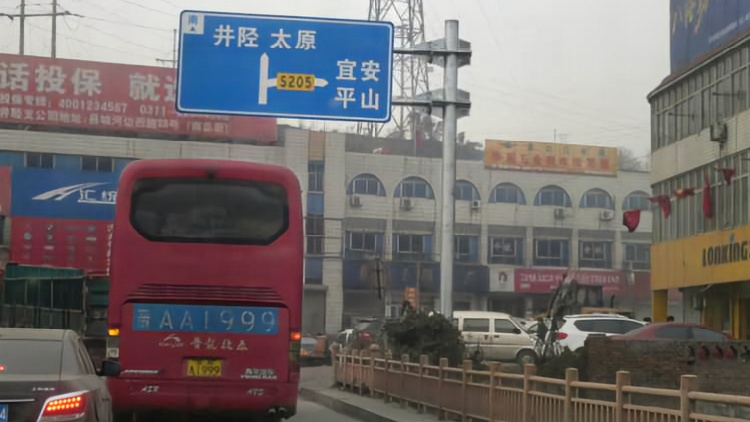}
    \end{subfigure}
    \\ \vspace{1pt}
    \begin{subfigure}{0.24\hsize}
        \includegraphics[width=\hsize]{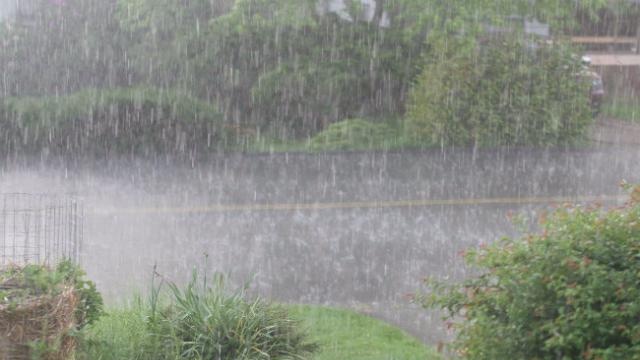}
    \end{subfigure}
    \begin{subfigure}{0.24\hsize}
        \includegraphics[width=\hsize]{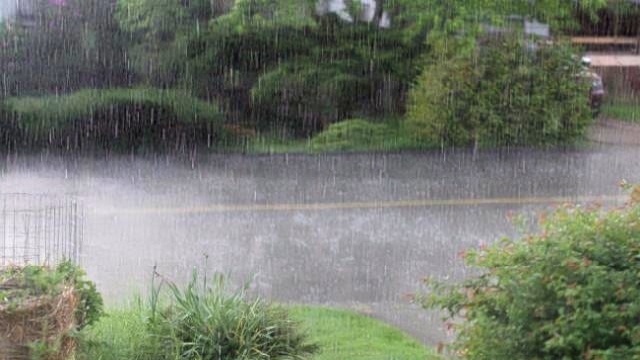}
    \end{subfigure}
    \begin{subfigure}{0.24\hsize}
        \includegraphics[width=\hsize]{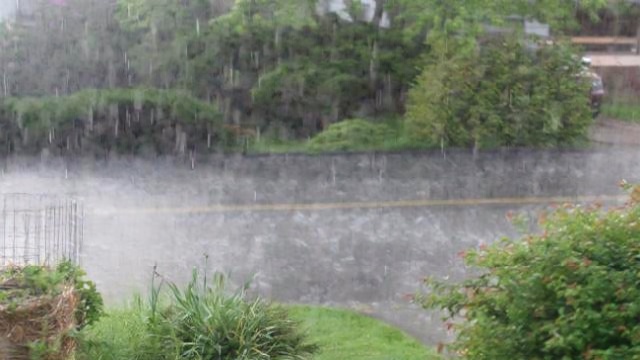}
    \end{subfigure}
    \begin{subfigure}{0.24\hsize}
        \includegraphics[width=\hsize]{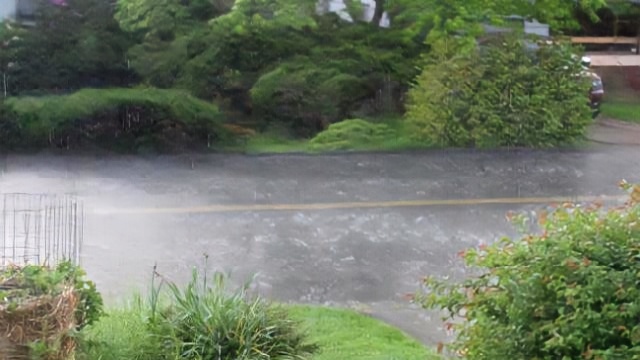}
    \end{subfigure}
    \\ \vspace{1pt}
    \begin{subfigure}{0.24\hsize}
        \includegraphics[width=\hsize]{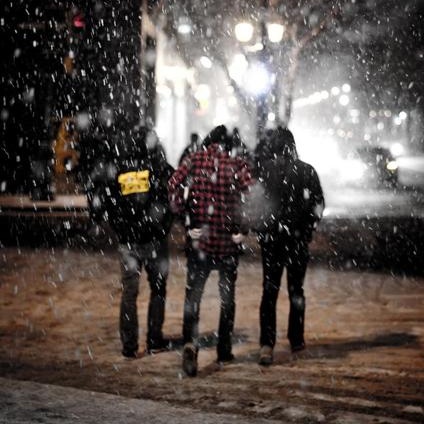}
        \caption{Input}
    \end{subfigure}
    \begin{subfigure}{0.24\hsize}
        \includegraphics[width=\hsize]{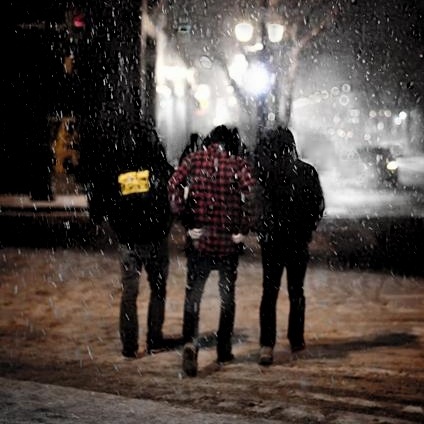}
        \caption{Initially restored}
    \end{subfigure}
    \begin{subfigure}{0.24\hsize}
        \includegraphics[width=\hsize]{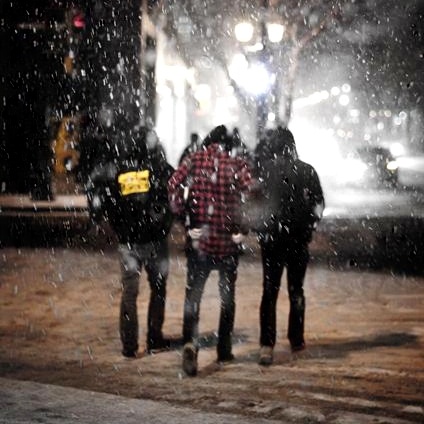}
        \caption{w/o initial}
    \end{subfigure}
    \begin{subfigure}{0.24\hsize}
        \includegraphics[width=\hsize]{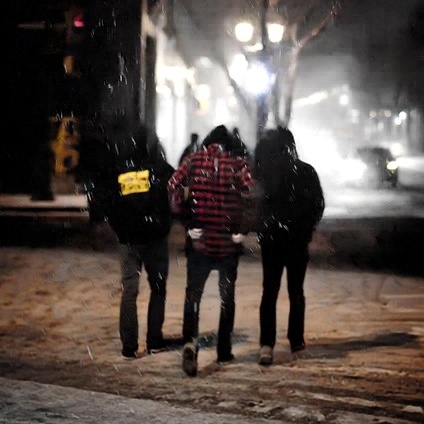}
        \caption{Our method}
    \end{subfigure}
    \\
    \caption{\textbf{Ablation study results of the proposed framework.}
    The weather-related artifacts of inputs are greatly reduced in the initially restored images, and our two-stage network further refines the clear scenes.
    }
    \label{fig:vis_ablation_framework}
\end{figure}

\begin{table}[t]
\centering
\caption{Ablation study on the overall framework and the major components design.}
\label{tab:ablation1}
\begin{adjustbox}{width=\hsize}
\begin{tabular}{c|ccc|ccc}
\toprule
                 & (a)        & (b)        & (c)        & (d)        & (e)        & Our method      \\ \midrule
U-Net $\theta_e$ & \checkmark &            & \checkmark & \checkmark & \checkmark & \checkmark      \\
U-Net $\theta_r$ &            & \checkmark & \checkmark & \checkmark & \checkmark & \checkmark      \\
TGGA             &            &            &            & \checkmark &            & \checkmark      \\
OGLA             &            &            &            &            & \checkmark & \checkmark      \\ \midrule
PSNR             & 24.86      & 23.97      & 25.63      & 26.70      & 26.47      & \textbf{27.22}  \\
SSIM             & 0.8098     & 0.8172     & 0.8488     & 0.8946     & 0.8963     & \textbf{0.9038} \\ \bottomrule
\end{tabular}
\end{adjustbox}
\end{table}

\begin{figure}[t]
    \centering
    \captionsetup[subfigure]{font=small,labelformat=empty,justification=centering}
    \begin{subfigure}{0.19\hsize}
        \includegraphics[width=\hsize]{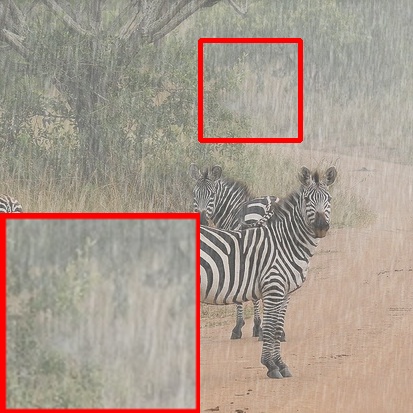}
    \end{subfigure}
    \begin{subfigure}{0.19\hsize}
        \includegraphics[width=\hsize]{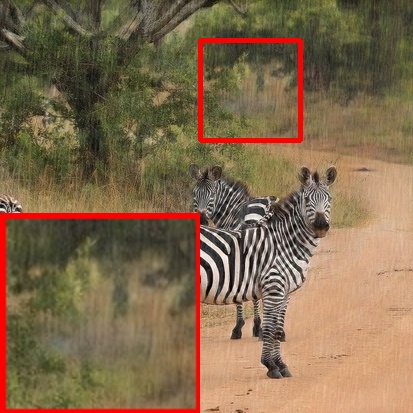}
    \end{subfigure}
    \begin{subfigure}{0.19\hsize}
        \includegraphics[width=\hsize]{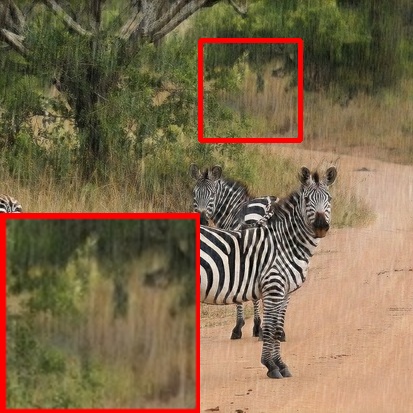}
    \end{subfigure}
    \begin{subfigure}{0.19\hsize}
        \includegraphics[width=\hsize]{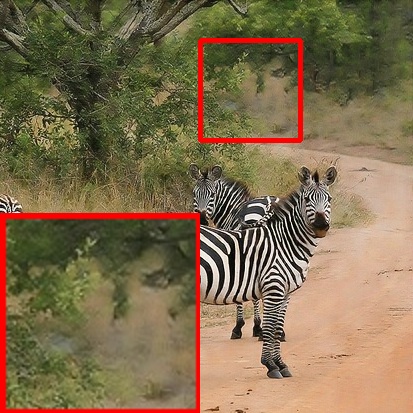}
    \end{subfigure}
    \begin{subfigure}{0.19\hsize}
        \includegraphics[width=\hsize]{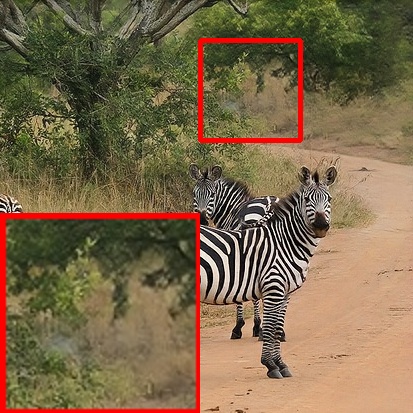}
    \end{subfigure}
    \\ \vspace{1pt}
    \begin{subfigure}{0.19\hsize}
        \includegraphics[width=\hsize]{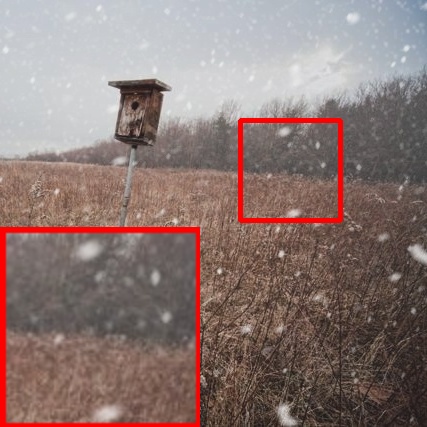}
        \caption{Input}
    \end{subfigure}
    \begin{subfigure}{0.19\hsize}
        \includegraphics[width=\hsize]{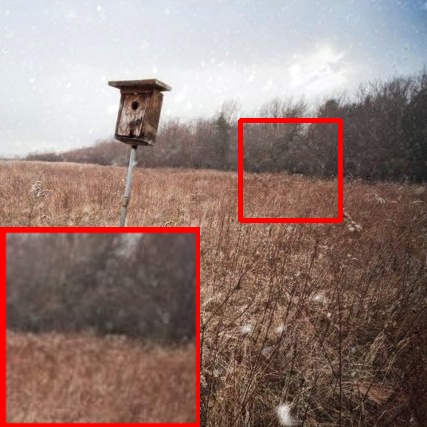}
        \caption{Basic}
    \end{subfigure}
    \begin{subfigure}{0.19\hsize}
        \includegraphics[width=\hsize]{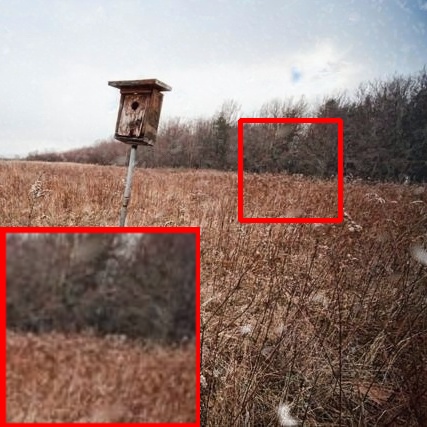}
        \caption{Basic + $t$}
    \end{subfigure}
    \begin{subfigure}{0.19\hsize}
        \includegraphics[width=\hsize]{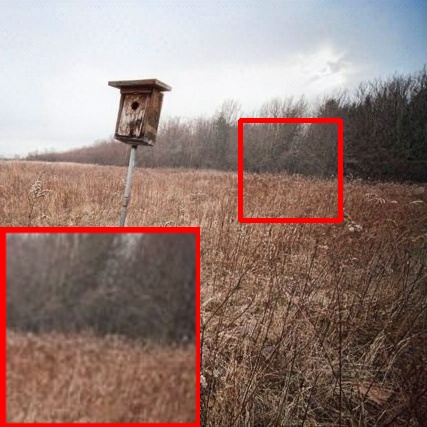}
        \caption{Basic + $\alpha$}
    \end{subfigure}
    \begin{subfigure}{0.19\hsize}
        \includegraphics[width=\hsize]{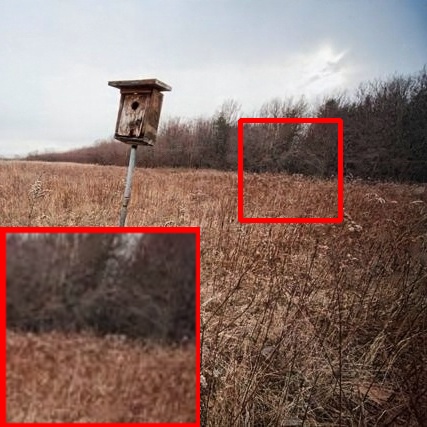}
        \caption{Our method}
    \end{subfigure}
    \\
    \caption{\textbf{Ablation on the guidance information.}
    Our method with transmission and occlusion guidance improves the visual quality of restoration results for ambiguous regions.
    }
    \label{fig:vis_ablation_guidance}
\end{figure}

\begin{table}[t]
\centering
\caption{Ablation study on the guidance information in our WACA module.}
\label{tab:ablation2}
\begin{tabular}{c|ccc}
\toprule
\multirow{2}{*}{Method} & \multicolumn{3}{c}{PSNR}                         \\ \cline{2-4} 
                        & Haze           & Rain           & Snow           \\ \midrule
Basic                   & 26.65          & 28.02          & 25.59          \\
Basic+Transmission      & 26.92          & 28.33          & 25.86          \\
Basic+Occlusion         & 26.79          & 28.41          & 25.93          \\
Our Method              & \textbf{27.01} & \textbf{28.56} & \textbf{26.09} \\ \bottomrule
\end{tabular}
\end{table}

\paragraph{Analysis of imaging model}
We evaluate the effectiveness of our imaging model, which accounts for common visual factors like occlusions and scattering effects in haze, rain, and snow, while incorporating the volumetric effects of rain and snow particles.
First, as shown in Fig. \ref{fig:vis_ablation_training_data} and Table \ref{tab:training_data}, our model trained with Weather30K demonstrates superior restoration performance, particularly in rain and snow conditions, compared to two other settings.
Second, an ablation study on these volumetric effects, modeled by multiple layers based on scene depth (Eq. (\ref{eq:ours3})), shows that the model trained on data synthesized with these effects removes more visible particles than the single-layer model \cite{hu2019depth}, as shown in Fig.~\ref{fig:vis_ablation_formula}.

\paragraph{Effectiveness of our two-stage framework} 
We evaluate the effectiveness of our two-stage framework by testing U-Net~$\theta_e$ (weather-related prior estimation) using the first stage's restoration output ($\hat{J}$ in Eq.~\eqref{eq:ours4}), and U-Net~$\theta_r$ (weather prior-based scene refinement) independently. As shown in Table~\ref{tab:ablation1} (a) and (b), both models perform poorly due to estimation errors and missing weather prior information. 
In contrast, incorporating $\hat{J}$ as input in the second stage improves performance, particularly in SSIM, as shown in Table~\ref{tab:ablation1} (c). An ablation study replacing U-Net~$\theta_e$ with a second scene refinement U-Net yields inferior results (24.31~dB in PSNR), highlighting the importance of weather prior knowledge in weather-related image restoration.

\paragraph{Effectiveness of the major components in our network}
We further verify the effectiveness of the transmission-guided global attention (TGGA) and occlusion-guided local attention (OGLA). Starting with the basic two-stage framework in Table~\ref{tab:ablation1}~(c), we progressively add these components, which enhance the refinement-stage features using prior information, and estimated transmission and occlusion from the estimation stage, within the weather-aware cross-attention (WACA) module (Fig.~\ref{fig:weathernet}~(c)). The weather-aware fuser (WAF) is also included to integrate the enhanced features when both TGGA and OGLA are applied. As shown in Table~\ref{tab:ablation1}~(d) and (e), TGGA and OGLA improve PSNR by around 1~dB compared to the basic framework, demonstrating their effectiveness in propagating weather prior-related features.
Moreover, combining TGGA and OGLA with WAF leads to further performance gains, highlighting the complementary benefits of the scattering effect and visible occlusion perspectives.

\paragraph{Effectiveness of guidance information in our WACA module}

In the WACA module, estimated transmission and occlusion are key for guiding attention in TGGA and OGLA, as well as feature integration in WAF (see Fig. \ref{fig:tgaola} and Fig. \ref{fig:weathernet}~(d)).
For the ablation study, we start with a baseline model, ``Basic,'' which includes global and local attention but lacks weather prior guidance. We then progressively add estimated transmission and occlusion as guiding signals.
As shown in Table~\ref{tab:ablation2}, the ``Basic'' model performs reasonably well by incorporating weather prior features during scene refinement with a cross-attention mechanism. 
Notably, using either transmission or occlusion as guidance improves metric scores, with transmission aiding haze restoration and occlusion benefiting rain and snow restoration.
When both transmission and occlusion are used together, our method achieves the best result, as seen in Table~\ref{tab:ablation2}.

\begin{table}[b]
\centering
\caption{Efficiency comparisons of model size (\textit{i.e.}, Params (M)) and runtime (s).}
\label{tab:efficiency}
\begin{adjustbox}{width=\hsize}
\begin{tabular}{c|cccccc}
\toprule
Method      & \begin{tabular}[c]{@{}c@{}}Trans-\\ Weather~\cite{valanarasu2022transweather}\end{tabular} & \begin{tabular}[c]{@{}c@{}}Two-\\ Stage~\cite{chen2022learning}\end{tabular} & \begin{tabular}[c]{@{}c@{}}Weather-\\ Diff~\cite{ozdenizci2023restoring}\end{tabular} & \begin{tabular}[c]{@{}c@{}}WGWS-\\ Net~\cite{zhu2023learning}\end{tabular} & \begin{tabular}[c]{@{}c@{}}MWFormer\\ \cite{zhu2024mwformer}\end{tabular} & \begin{tabular}[c]{@{}c@{}}Our \\ Method\end{tabular} \\ \midrule
Params   & 38.1                                                                   & 28.7                                                                                       & 83.0                                                                         & 7.6                                                                                  & 182.8                                                                        & 32.7                                                  \\
Runtime  & 0.013                                                                      & 0.023                                                                                      & 17.881                                                                        & 0.073                                                                                & 0.092                                                                      & 0.031                                                 \\ \bottomrule
\end{tabular}
\end{adjustbox}
\end{table}

\paragraph{Additional qualitative analysis}
The quantitative results in Tables~\ref{tab:ablation1} and~\ref{tab:ablation2} validate the effectiveness of our network and the WACA module. To further illustrate the framework’s efficacy, we provide qualitative analysis focusing on the impact of transmission and occlusion guidance.
Fig.~\ref{fig:vis_ablation_framework} shows the initially restored images, $\hat{J}$, from our two-stage framework. While the model may not capture all physical complexities, it effectively removes weather artifacts like rain streaks, snowflakes, and fog scattering. The weather priors also help address ambiguous occlusions, enhancing scene clarity.
As shown in Fig.~\ref{fig:vis_ablation_guidance}, transmission and occlusion guidance improve restoration in complementary ways. Transmission guidance (``Basic'' + Transmission) better removes global haze effects, especially in background areas, while occlusion guidance (``Basic'' + Occlusion) excels at restoring rain streaks and snowflakes in local regions. Together, these guiding signals significantly enhance both qualitative and quantitative restoration performance.

\begin{figure}
    \centering
    \captionsetup[subfigure]{font=small,labelformat=empty,justification=centering}
    \begin{subfigure}{0.32\hsize}
        \includegraphics[width=\hsize]{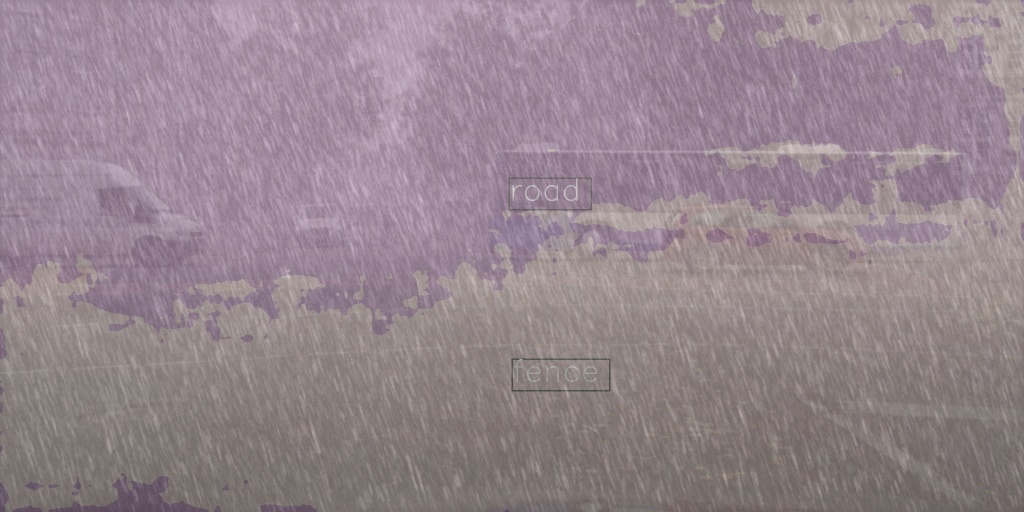}
        \caption{Input}
    \end{subfigure}
    \begin{subfigure}{0.32\hsize}
        \includegraphics[width=\hsize]{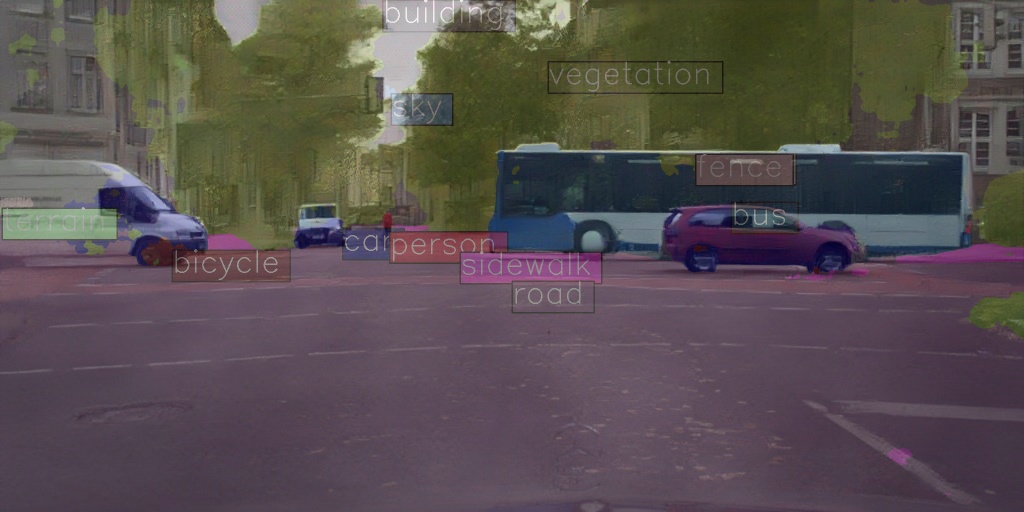}
        \caption{WGWS-Net~\cite{zhu2023learning}}
    \end{subfigure}
    \begin{subfigure}{0.32\hsize}
        \includegraphics[width=\hsize]{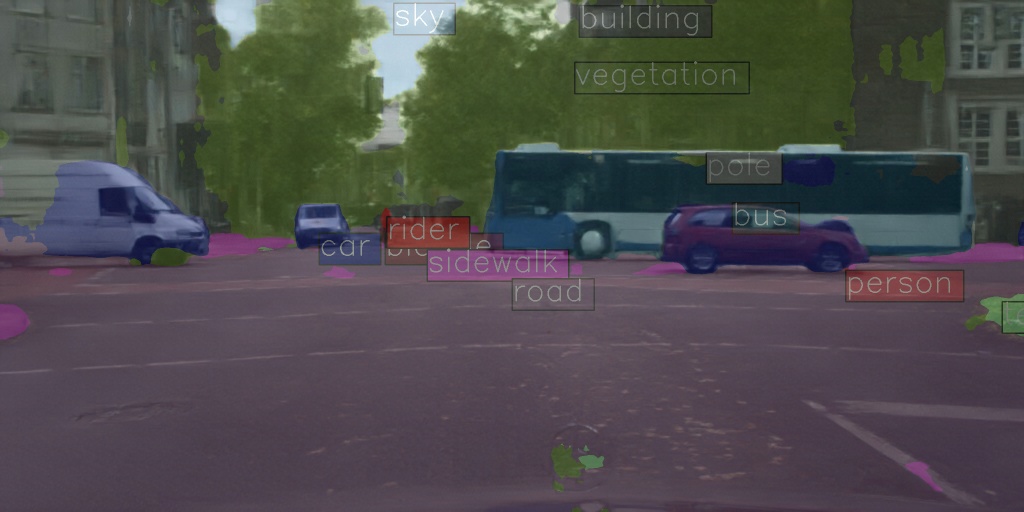}
        \caption{Our method}
    \end{subfigure}
    \\ \vspace{1pt}
    \begin{subfigure}{0.32\hsize}
        \includegraphics[width=\hsize]{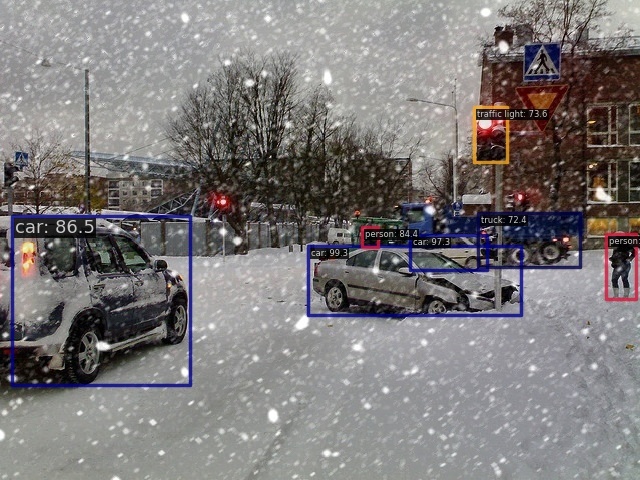}
        \caption{Input}
    \end{subfigure}
    \begin{subfigure}{0.32\hsize}
        \includegraphics[width=\hsize]{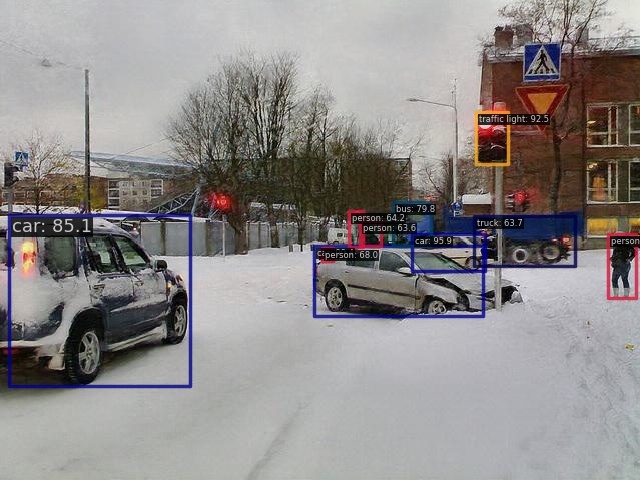}
        \caption{TransWeather~\cite{valanarasu2022transweather}}
    \end{subfigure}
    \begin{subfigure}{0.32\hsize}
        \includegraphics[width=\hsize]{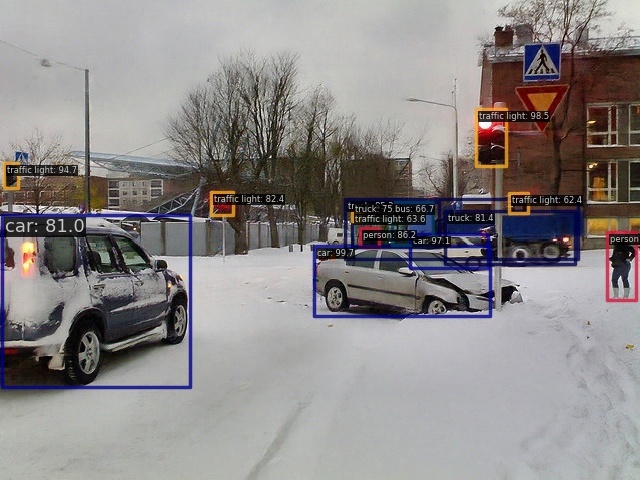}
        \caption{Our method}
    \end{subfigure}
    \\ \vspace{1pt}
    \begin{subfigure}{0.32\hsize}
        \includegraphics[width=\hsize]{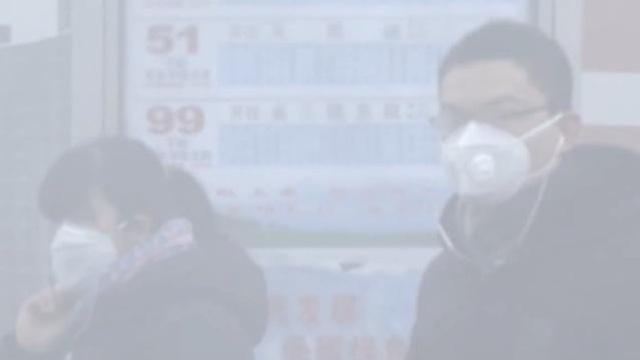}
        \caption{Input}
    \end{subfigure}
    \begin{subfigure}{0.32\hsize}
        \includegraphics[width=\hsize]{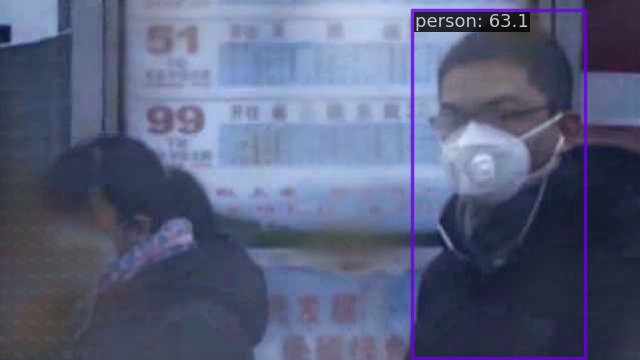}
        \caption{Two-Stage~\cite{chen2022learning}}
    \end{subfigure}
    \begin{subfigure}{0.32\hsize}
        \includegraphics[width=\hsize]{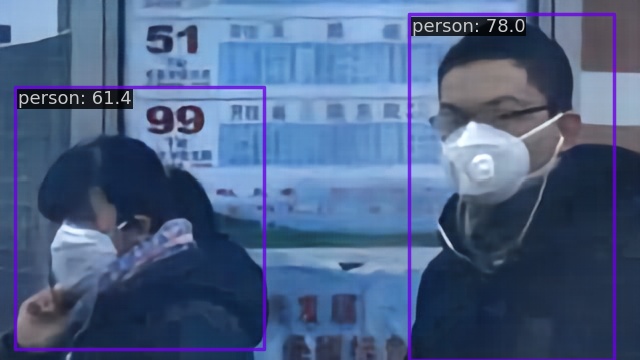}
        \caption{Our method}
    \end{subfigure}
    \\ \vspace{1pt}
    \begin{subfigure}{0.32\hsize}
        \includegraphics[width=\hsize]{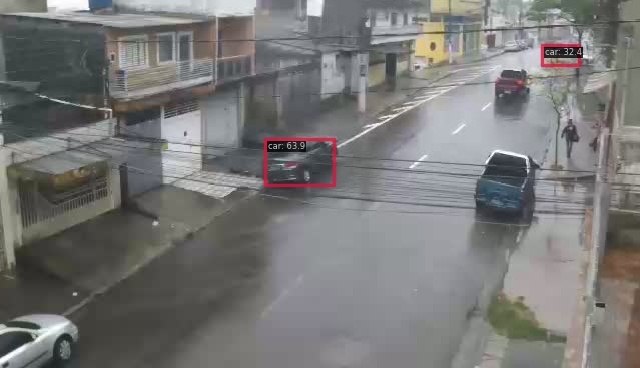}
        \caption{Input}
    \end{subfigure}
    \begin{subfigure}{0.32\hsize}
        \includegraphics[width=\hsize]{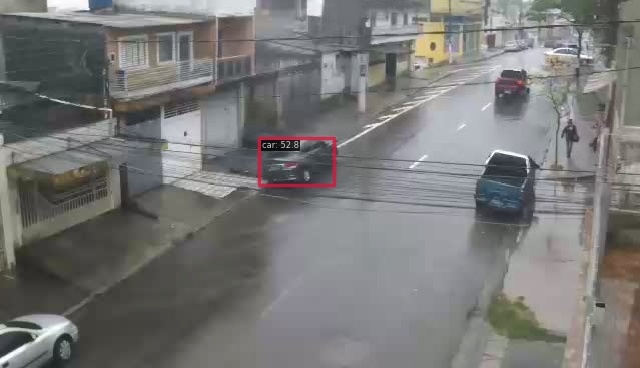}
        \caption{WeatherDiff~\cite{ozdenizci2023restoring}}
    \end{subfigure}
    \begin{subfigure}{0.32\hsize}
        \includegraphics[width=\hsize]{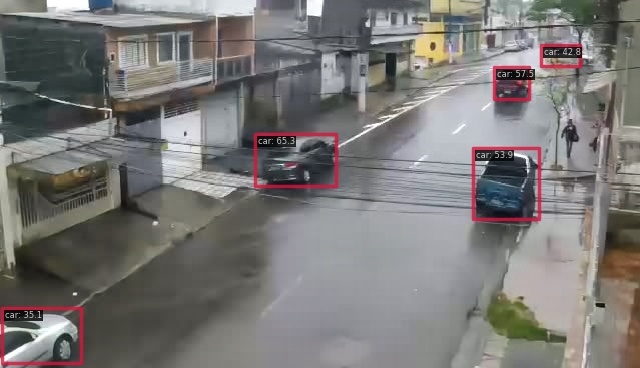}
        \caption{Our method}
    \end{subfigure}
    \\
    \caption{Visualization of downstream application results before and after applying different weather removal approaches.
    Example images are from Weather30K (first two rows), RTTS~\cite{li2018benchmarking}, and RIS~\cite{li2019single} (last two rows) datasets.
    }
    \label{fig:vis_application}
\end{figure}

\subsection{Computational Efficiency Analysis}

\begin{table}[t]
\centering
\caption{Comparisons of downstream application results. mIoU and mAP are metrics for image segmentation and object detection. Our method helps the downstream applications best.}
\label{tab:application}
\begin{tabular}{c|ccc}
\toprule
\multirow{2}{*}{Method}                        & \multicolumn{3}{c}{mIoU $\uparrow$ / mAP $\uparrow$ / (mAP $\uparrow$)}                                             \\ \cline{2-4} 
                                               & Haze                                      & Rain                                      & Snow                        \\ \midrule
Input                                          & 51.4/33.0/50.1                            & \ 5.0/34.4/21.7                           & 19.7/33.2                   \\
Ground Truth                                   & 72.7/38.1/\ \ - \ \                       & 70.5/44.9/\ \ - \ \                       & 72.9/52.6                   \\ \hline
TransWeather~\cite{valanarasu2022transweather} & 41.8/35.4/37.7                            & 25.5/38.6/16.7                            & 23.9/40.2                   \\
Two-Stage~\cite{chen2022learning}              & 57.2/36.4/{\ul 50.6}                      & 41.2/40.0/20.4                            & 40.9/40.6                   \\
WeatherDiff~\cite{ozdenizci2023restoring}      & {\ul 65.5}/36.0/49.0                      & {\ul 51.1}/41.1/21.3                      & {\ul 52.7}/{\ul 43.7}       \\
WGWS-Net~\cite{zhu2023learning}                & 63.7/{\ul 36.8}/49.7                      & 50.9/{\ul 41.7}/17.4                      & 51.1/43.5                   \\
MWFormer~\cite{zhu2024mwformer}                & 57.0/35.7/44.2                            & 38.8/40.6/{\ul 22.0}                      & 41.6/47.0                   \\
Our Method                                     & \textbf{66.3}/\textbf{37.0}/\textbf{52.7} & \textbf{51.7}/\textbf{41.8}/\textbf{22.4} & \textbf{53.2}/\textbf{44.2} \\ \bottomrule

\end{tabular}
\end{table}

We compare the model efficiency of our network with state-of-the-art methods, including model size and runtime.
The runtime is computed based on input sizes of (256, 256) on a TITAN RTX GPU.
For the diffusion model-based WeatherDiff~\cite{ozdenizci2023restoring} with a sliding-window inference strategy, we adopt the default settings during evaluation, \textit{i.e.}, sampling timesteps of 25 and grid cell width of 16.
As summarized in Table~\ref{tab:efficiency}, the model efficiency of our network is comparable to strong competitors~\cite{valanarasu2022transweather,chen2022learning,zhu2024mwformer}, \textit{e.g.}, fewer parameters than TransWeather (32.7M vs. 38.1M) and MWFormer, and slightly larger timing than Two-Stage.
Our method has less inference time cost than WGWS-Net~\cite{zhu2023learning} composed of complicated path routings.
Meanwhile, our model size and runtime are much smaller than the diffusion-based model WeatherDiff.
Moreover, our network notably outperforms these state-of-the-art methods quantitatively and qualitatively in various adverse weather scenarios.

\subsection{Applications on Downstream Vision Tasks}

\begin{figure*}
    \centering
    \includegraphics[width=\hsize]{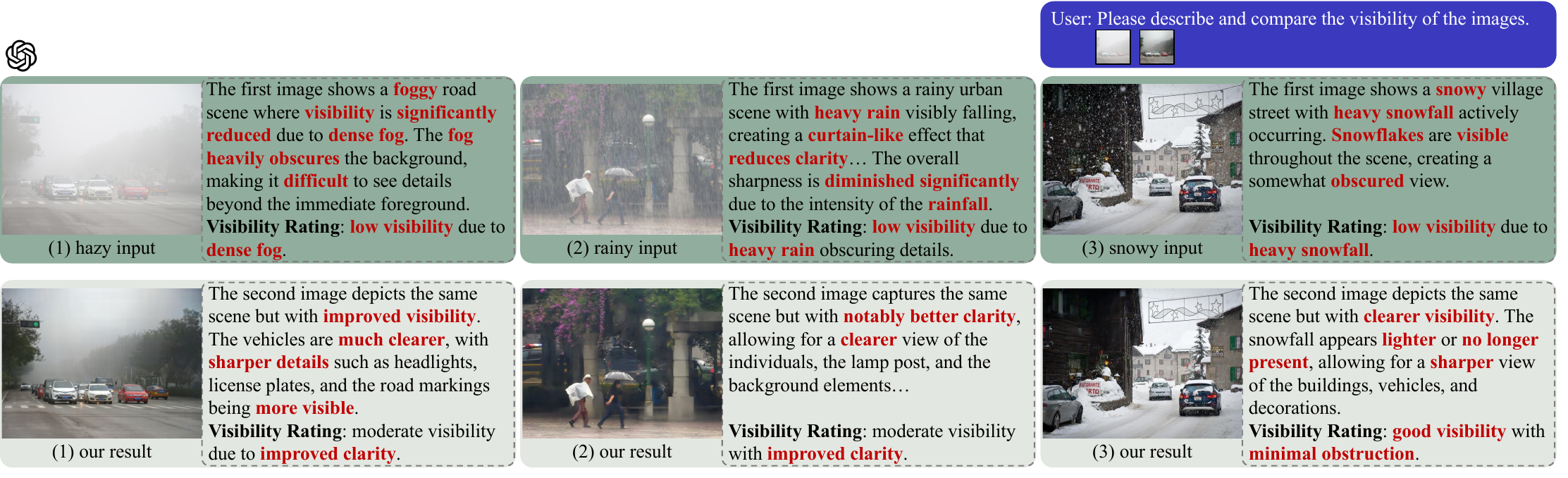}
    \caption{
    GPT-4o is used to describe and compare the input real-world images (first row) and our restoration results (second row).
    The results clearly demonstrate that our method significantly improves the visibility of the input images.}
    \label{fig:vlm}
\end{figure*}

Adverse weather conditions, such as haze, rain, and snow, challenge high-level computer vision tasks. In this section, we evaluate the impact of these conditions on image segmentation and object detection.
We test the PSPNet~\cite{zhao2017pyramid} image segmentation model and the Faster-RCNN~\cite{ren2015faster} object detection model on the Cityscapes (500 images) and COCO (600 images) subsets of our Weather30K testing set. The models are initially trained on clear images using MMSegmentation\footnote{\url{https://github.com/open-mmlab/mmsegmentation}} and MMDetection\footnote{\url{https://github.com/open-mmlab/mmdetection}} toolkits. Evaluation is done on inputs, ground truths, and restored images from various weather removal methods.
Additionally, for object detection, we use RetinaNet~\cite{lin2017focal} on real haze and rain images from the RTTS~\cite{li2018benchmarking} and RIS~\cite{li2019single} datasets, with ground truth bounding boxes available.

Quantitative results, including mean intersection-over-union (mIoU) for segmentation and mean average precision (mAP) for detection, are shown in Table~\ref{tab:application}. 
We observe a significant drop in model performance under adverse weather, due to reduced visibility. However, our method, which restores clear images, enhances model accuracy and outperforms other approaches in both synthetic and real-world scenarios. Fig.~\ref{fig:vis_application} illustrates the effectiveness of our method in improving high-level vision tasks.

We further use the advanced large multimodal model (LMM), GPT-4o \cite{hurst2024gpt}, to describe and compare the visibility of images before and after restoration. As shown in Fig.~\ref{fig:vlm}, our method significantly improves image visibility, allowing LMMs to more accurately describe scene content under adverse weather conditions.

\begin{figure}
    \centering
    \captionsetup[subfigure]{font=small,labelformat=empty,justification=centering}
    \begin{subfigure}{0.32\hsize}
        \includegraphics[width=\hsize]{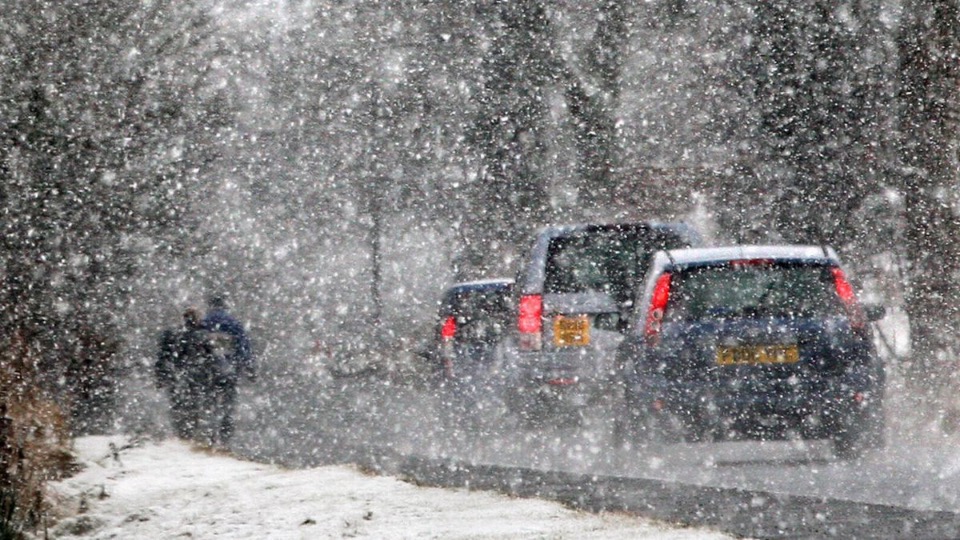}
        \caption{Input}
    \end{subfigure}
    \begin{subfigure}{0.32\hsize}
        \includegraphics[width=\hsize]{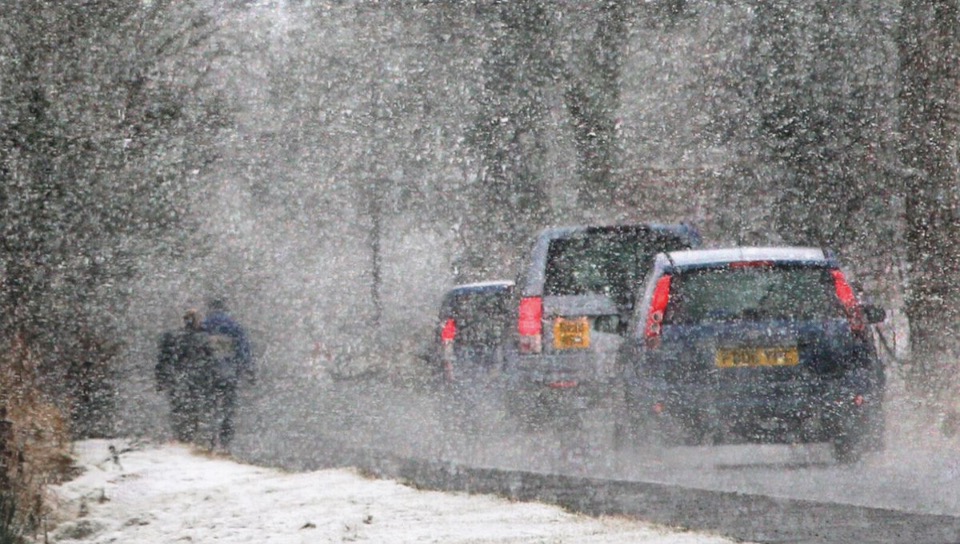}
        \caption{TransWeather~\cite{valanarasu2022transweather}}
    \end{subfigure}
    \begin{subfigure}{0.32\hsize}
        \includegraphics[width=\hsize]{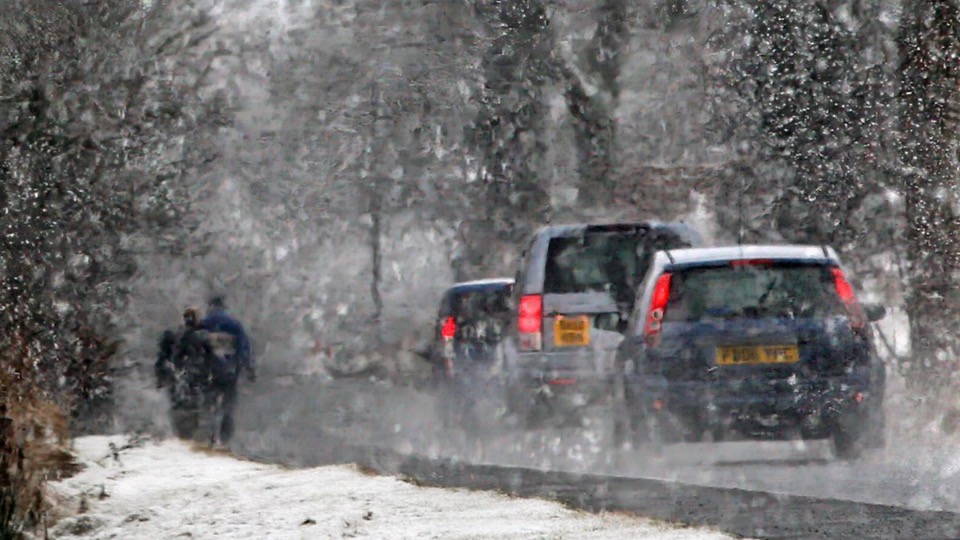}
        \caption{Our method}
    \end{subfigure}
    \caption{A failure case with extreme adverse weather effects.
    }
    \label{fig:vis_limitation}
\end{figure}

\subsection{Discussion}
The experimental results demonstrate that our proposed imaging model effectively captures real-world image characteristics, making significant progress in adverse weather image restoration. By addressing image degradation from an imaging perspective and complementing data-driven approaches, our work contributes to the development of a more robust and comprehensive all-weather image restoration system.
While our approach introduces a unified imaging formulation and accounts for a wide range of natural visual phenomena, there remain certain areas for improvement. For instance, as shown in Fig. \ref{fig:vis_limitation}, under extreme adverse weather conditions, our method can recover much of the underlying scene, though restoring finer details remains challenging. 
However, this is a common limitation across current methods, given the difficulty of modeling the vast diversity of real-world scenarios.

Future directions could include leveraging larger and more diverse datasets, combined with semi-supervised learning techniques, to further enhance generalization across a broader range of conditions \cite{xu2024towards}. Additionally, incorporating generative priors from advanced pre-trained text-to-image diffusion models \cite{wang2024exploiting} may provide further improvements in handling extreme weather effects, driving advancements in all-weather image restoration systems.

\section{Conclusion}
\label{sec:conclusion}
In this work, we analyze the visual effects of common adverse weather conditions and propose a unified imaging model that addresses occlusions and fog-like scattering across various weather types. We develop a novel weather-prior-based network and a weather-aware cross-attention module that enhance scene recovery using estimated occlusion and transmission maps. Various experiments on synthetic and real-world datasets show that our method outperforms current state-of-the-art approaches. This work advances all-in-one adverse weather image restoration by revisiting visual phenomena under natural weather conditions. Future work will explore incorporating real-world data, human feedback, and generative priors to further improve restoration performance.

\section*{Acknowledgments}
This work was supported by the National Key R\&D Program of China (Grant No. 2023YFE0202700), the Research Grants Council of Hong Kong (Grant No. 14200824), and the Hong Kong Innovation and Technology Fund (Grant No. MHP/092/22).

\bibliographystyle{IEEEtran}
\bibliography{ref}

\end{document}